\pdfoutput=1
\newif\ifazpreprint
\azpreprinttrue
\newif\ifazdatapublic
\azdatapublictrue
\newif\ifazcodepublic
\documentclass{article}

\usepackage[T1]{fontenc}
\usepackage{microtype}
\usepackage{graphicx}
\usepackage{subcaption}
\usepackage{booktabs}
\usepackage{multirow}
\usepackage{array}
\usepackage{multicol}
\usepackage{enumitem}
\usepackage{xcolor}
\usepackage{colortbl}
\definecolor{overallbg}{HTML}{F2F5FA}
\usepackage{xurl}
\usepackage{hyperref}

\ifdefined\ifazpreprint\else\expandafter\newif\csname ifazpreprint\endcsname\fi
\ifazpreprint
  \usepackage[preprint]{icml2026}
\else
  \usepackage{icml2026}
\fi

\usepackage{amsmath}
\usepackage{amssymb}
\usepackage{mathtools}
\usepackage{bm}
\usepackage{siunitx}
\usepackage{needspace}
\usepackage{placeins}

\usepackage{listings}
\usepackage[listings,breakable]{tcolorbox}
\definecolor{codebg}{HTML}{F6F8FA}
\definecolor{codeframe}{HTML}{C8D1DC}
\definecolor{codekw}{HTML}{1D4E9E}
\definecolor{codestr}{HTML}{A33A1D}
\definecolor{codecom}{HTML}{6E7781}
\definecolor{codetype}{HTML}{0B7A75}
\newcommand{\codefont}{\fontfamily{lmtt}\selectfont}
\lstdefinestyle{azbase}{
  basicstyle=\codefont\footnotesize, columns=fullflexible, keepspaces=true,
  upquote=true, showstringspaces=false, tabsize=4, breaklines=false,
  keywordstyle=\color{codekw}, stringstyle=\color{codestr},
  commentstyle=\color{codecom}\itshape,
}
\lstdefinestyle{azpython}{
  style=azbase, language=Python, morekeywords={None,True,False},
  emph={torch,cKDTree,DomainMesh,float,len,range,abs,max}, emphstyle=\color{codetype},
}
\lstdefinestyle{aztree}{
  style=azbase, basicstyle=\codefont\fontsize{6.6}{7.6}\selectfont,
  alsoletter={-`|()}, moredelim=[s][\color{codecom}]{[}{]},
  emph={float32,float64,int64,uint8,bool}, emphstyle=\color{codetype},
  emph={[2]|--,`--,|,(Mesh),(empty),()}, emphstyle={[2]\color{codecom}},
}
\tcbset{azcode/.style={listing only, breakable, colback=codebg, colframe=codeframe,
  boxrule=0.4pt, arc=1.5pt, outer arc=1.5pt, boxsep=1pt, left=4pt, right=4pt,
  top=2pt, bottom=2pt, before skip=6pt, after skip=6pt}}
\newtcbinputlisting{\pyinput}[1]{azcode, listing options={style=azpython}, listing file={#1}}
\newtcblisting{treecode}{azcode, listing options={style=aztree}}
\newtcblisting{pycode}[1][]{azcode, listing options={style=azpython}, #1}
\usepackage[capitalize,noabbrev]{cleveref}

\makeatletter\@secpenalty=0\makeatother
\ifazpreprint
  \makeatletter
  \renewcommand{\ICML@preprint}{\textit{Preprint.}}
  \makeatother
  \long\def\reviewonly#1{}
  \long\def\preprintonly#1{#1}
  \long\def\reviewpreprint#1#2{#2}
\else
  \long\def\reviewonly#1{#1}
  \long\def\preprintonly#1{}
  \long\def\reviewpreprint#1#2{#1}
\fi

\ifdefined\ifazdatapublic\else\expandafter\newif\csname ifazdatapublic\endcsname\fi
\ifdefined\ifazcodepublic\else\expandafter\newif\csname ifazcodepublic\endcsname\fi
\ifazpreprint
  \ifazdatapublic\long\def\azpublic#1#2{#1}\else\long\def\azpublic#1#2{#2}\fi
  \ifazcodepublic\long\def\azcodepublic#1#2{#1}\else\long\def\azcodepublic#1#2{#2}\fi
\else
  \long\def\azpublic#1#2{#1}
  \long\def\azcodepublic#1#2{#1}
\fi

\definecolor{tbdcolor}{RGB}{204,51,17}
\ifazpreprint\ifdefined\azpreprintdraft\else\def\azplaceholdersfail{}\fi\fi
\ifdefined\azplaceholdersfail
  \newcommand{\azplaceholder}[1]{\PackageError{aerozoo}{Placeholder #1 in the preprint}%
    {Resolve it, or wrap the text in \string\reviewonly.}}
  \newcommand{\tbd}[1]{\azplaceholder{\string\tbd}}
  \newcommand{\res}{\azplaceholder{\string\res}}
\else
  \newcommand{\tbd}[1]{{\leavevmode\color{tbdcolor}\textbf{[TBD: #1]}}}
  \newcommand{\res}{\textcolor{tbdcolor}{--}}
\fi

\reviewpreprint{%
  \newcommand{\datasetlink}{\url{https://huggingface.co/datasets/ANONYMIZED/AeroZoo}\ \textcolor{tbdcolor}{(placeholder)}}%
  \newcommand{\codelink}{\url{https://github.com/ANONYMIZED/physicsnemo}\ \textcolor{tbdcolor}{(placeholder)}}%
}{%
  \newcommand{\datasetlink}{\url{https://huggingface.co/datasets/nvidia/AeroZoo}}%
  \newcommand{\codelink}{\url{https://github.com/NVIDIA/physicsnemo}}%
}

\newcommand{\aerozoo}{AeroZoo}
\newcommand{\key}[1]{\texttt{\detokenize{#1}}}

\newcommand{\Lref}{L_\mathrm{ref}}
\newcommand{\Uref}{U_\mathrm{ref}}
\newcommand{\qref}{q_\mathrm{ref}}
\newcommand{\Rey}{\mathit{Re}}
\newcommand{\Tc}{T_c}
\newcommand{\vu}{\mathbf{u}}
\newcommand{\vn}{\mathbf{n}}
\newcommand{\vx}{\mathbf{x}}
\newcommand{\vC}{\mathbf{C}}
\newcommand{\vtau}{\boldsymbol{\tau}}
\newcommand{\hatvu}{\hat{\mathbf{u}}}



\makeatletter
\newcommand{\aznset}[2]{\expandafter\def\csname azn@#1\endcsname{#2}}
\newcommand{\azn}[1]{\ifcsname azn@#1\endcsname\csname azn@#1\endcsname\else\PackageError{aznumbers}{Unknown number key #1}{See generated/numbers.tex}\fi}
\makeatother
\aznset{planned}{4{,}360}
\aznset{complete}{3{,}815}
\aznset{complete-pct}{87.5}
\aznset{failed}{208}
\aznset{timed-out}{337}
\aznset{stable}{2{,}816}
\aznset{unstable}{999}
\aznset{stable-share}{74}
\aznset{frozen-date}{2026-10-02}
\aznset{index-sha}{dc01540167cd2f0e}
\aznset{ducts}{303}
\aznset{body-free-ducts}{161}
\aznset{duct-shapes}{10}
\aznset{stable-pct-crossflow}{30}
\aznset{stable-pct-opposed_hemisphere_symmetry}{34}
\aznset{stable-pct-internal_duct}{67}
\aznset{stable-pct-other-min}{88}
\aznset{stable-pct-other-max}{94}
\aznset{shortfall-confined}{183}
\aznset{shortfall-confined-planned}{617}
\aznset{shortfall-confined-pct}{29.7}
\aznset{shortfall-duct}{135}
\aznset{shortfall-duct-planned}{438}
\aznset{shortfall-duct-pct}{30.8}
\aznset{shortfall-confined-timed-out}{167}
\aznset{shortfall-duct-failed}{121}
\aznset{timed-out-top-decade}{230}
\aznset{nu-decades}{10}
\aznset{nu-air-water-pct}{23}
\aznset{air-nu-change-15-25-pct}{6}
\aznset{air-nu-change-0-35-pct}{24}
\aznset{speed-span}{200}
\aznset{q-span}{40{,}000}
\aznset{cells-total-b}{23.8}
\aznset{cells-median-m}{5.0}
\aznset{cells-min-m}{0.6}
\aznset{cells-max-m}{24}
\aznset{bytes-total-tb}{1.48}
\aznset{core-hours-total-m}{8.1}
\aznset{boundary-faces-median-m}{0.43}
\aznset{unstable-fail-velocity-pct}{83}
\aznset{unstable-fail-pressure-pct}{8}
\aznset{unstable-fail-force-pct}{58}
\aznset{laminar-low-share-pct}{90}
\aznset{laminar-high-median-pct}{0.7}
\aznset{cf-over-laminar-lo}{2.3}
\aznset{cf-over-laminar-hi}{2.5}
\aznset{cf-over-turbulent-lo}{1.1}
\aznset{cf-over-turbulent-hi}{1.4}
\aznset{cf-median-fall}{18}
\aznset{force-small-pct}{21}
\aznset{force-span}{100}
\aznset{yplus-median}{18}
\aznset{yplus-q1}{15}
\aznset{yplus-q3}{25}
\aznset{yplus-5-30-pct}{77}
\aznset{yplus-safety-factor}{2}
\aznset{yplus-plate-target}{15}
\aznset{nut-inlet-coefficient}{0.67}
\aznset{nut-median-low}{2}
\aznset{nut-median-high}{28{,}000}
\aznset{nut-above-100-pct}{56}
\aznset{nut-over-ul-median}{0.0008}
\aznset{nut-inlet-over-ul-median}{0.0016}
\aznset{nut-over-ul-p05}{0.00006}
\aznset{nut-over-ul-p95}{0.011}
\aznset{nut-top-decade-above-90k-pct}{28}
\aznset{nut-volume-over-inlet-median}{0.96}
\aznset{re-eff-median}{1{,}100}
\aznset{vol-enclosure-ratio}{38}
\aznset{vol-med-confined}{5.8}
\aznset{vol-med-hemisphere}{217}
\aznset{study-cases}{four}
\aznset{scheme-u-min}{1.8}
\aznset{scheme-u-max}{8.9}
\aznset{fine-cells-max-m}{4.6}
\aznset{courant-u-pct}{0.46}
\aznset{courant-cf}{0.0011}
\aznset{restart-u-pct}{0.009}
\aznset{restart-resumes}{two}
\aznset{symmetry-samples}{391}
\aznset{symmetry-offset-max-pct}{2.5}
\aznset{vendi-n}{400}
\aznset{vendi-ratio-max}{36}
\aznset{vendi-k-lo}{5}
\aznset{vendi-k-hi}{100}
\aznset{vendi-ratio-min-k-lo}{1.9}
\aznset{vendi-ratio-max-k-lo}{23}
\aznset{vendi-ratio-max-k-hi}{64}
\aznset{vendi-hist-ratio-min-lo}{1.5}
\aznset{vendi-hist-ratio-min-hi}{1.6}
\aznset{vendi-settings}{35}
\aznset{vendi-settings-ordered}{30}
\aznset{vendi-d2-ratio-min-lo}{0.8}
\aznset{vendi-d2-ratio-min-hi}{1.0}
\aznset{vendi-n-lo}{50}
\aznset{vendi-ratio-max-n-lo}{7}
\aznset{vendi-ceiling-pct}{97}
\aznset{vendi-shiftsuv-missing}{three}
\aznset{splits}{22}
\aznset{split-bank_cars-train-canonical}{2{,}173}
\aznset{split-bank_cars-val-canonical}{321}
\aznset{split-bank_cars-test-canonical}{310}
\aznset{split-family_non_box-train-canonical}{1{,}850}
\aznset{split-family_non_box-val-canonical}{274}
\aznset{split-family_non_box-test-canonical}{692}
\aznset{split-full-train-canonical}{2{,}011}
\aznset{split-full-val-canonical}{290}
\aznset{split-full-test-canonical}{515}
\aznset{split-profile_dual_inlet-train-canonical}{2{,}044}
\aznset{split-profile_dual_inlet-val-canonical}{313}
\aznset{split-profile_dual_inlet-test-canonical}{459}
\aznset{split-reynolds-train-canonical}{1{,}577}
\aznset{split-reynolds-val-canonical}{238}
\aznset{split-reynolds-test-canonical}{1{,}001}
\aznset{split-scarce-train-canonical}{321}
\aznset{split-scarce-val-canonical}{290}
\aznset{split-scarce-test-canonical}{515}
\aznset{split-super_scarce-train-canonical}{53}
\aznset{split-super_scarce-val-canonical}{290}
\aznset{split-super_scarce-test-canonical}{515}
\aznset{test-size-ratio-canonical}{7.5}
\aznset{test-size-ratio-released}{4.6}
\aznset{test-size-ratio-planned}{4.5}
\aznset{test-canonical-min}{134}
\aznset{test-canonical-max}{1{,}001}
\aznset{split-full-val-in-held-out-test}{165}
\aznset{split-ho-val-in-full-test}{77}
\aznset{split-ho-val-in-full-test-pct}{15.0}
\aznset{layout-test-bodiless-planned}{66}
\aznset{bank-test-bodiless-min}{9}
\aznset{bank-test-bodiless-max}{45}
\aznset{bank-cars-test-bodiless}{12}
\aznset{layout-test-bodiless}{38}
\aznset{inflow-cases}{three}
\aznset{inflow-A-epsU-near-two-lo}{0.05}
\aznset{inflow-A-epsU-near-two-hi}{1.0}
\aznset{inflow-A-dCF-two-lo}{1.2}
\aznset{inflow-A-dCF-two-hi}{2.0}
\aznset{inflow-A-blobs-epsU-near}{6.0}
\aznset{inflow-A-blobs-dCF}{27}
\aznset{inflow-B-laminar-pct}{79}
\aznset{inflow-B-dCF}{19}
\aznset{inflow-3157-cf}{0.0031}
\aznset{inflow-3157-A-epsCF}{0.12}
\aznset{inflow-3157-B-epsCF}{1.15}

\aznset{fs-full-epsU}{0.440}
\aznset{fs-full-epsU-pct}{44.0}
\aznset{fs-full-epsU-released}{0.470}
\aznset{fs-full-epsp}{1.809}
\aznset{fs-full-epsCF}{0.869}
\aznset{fs-full-epsK}{0.994}
\aznset{fs-open-epsU}{0.046}
\aznset{fs-sphere-epsU}{0.039}
\aznset{fs-dual-epsU}{1.659}
\aznset{fs-crossflow-epsU}{0.947}
\aznset{fs-cases}{3{,}815}
\aznset{fs-full-n}{515}
\aznset{fs-force-gap}{5\times10^{-6}}

\aznset{single-rows}{15}
\aznset{single-gpu-hours}{1{,}654}
\aznset{ms-gpuh-total}{5{,}137}
\aznset{isla-joint-gpuh-total}{662}
\aznset{t-params-m}{9.0}
\aznset{gt-params-m}{27.6}
\aznset{isla-params-m}{67.5}
\aznset{gt-seeds}{six}
\aznset{gt-heldout-seeds}{three}
\aznset{isla-seeds}{six}
\aznset{isla-joint-seeds}{four}
\aznset{isla-heldout-seeds}{two}
\aznset{tr-gpu-hours}{25}
\aznset{gt-gpu-hours}{66}
\aznset{isla-gpu-hours}{113}
\aznset{isla-joint-gpu-hours}{166}
\aznset{gtb-infer-seconds}{12.9}
\aznset{gtb-infer-gib}{83}
\aznset{fs-full-epsU-pct}{44.0}
\aznset{t-nofa-epsU}{0.617}
\aznset{t-nofa-open}{0.181}
\aznset{t-nofa-sphere}{1.203}
\aznset{gt-notok-epsU}{0.898}
\aznset{gt-notok-epsU-pct}{89.8}
\aznset{gt-notok-epsU-near}{0.899}
\aznset{gt-notok-S}{1.438}
\aznset{gt-notok-boxes-lo}{0.52}
\aznset{gt-notok-boxes-hi}{0.62}
\aznset{gt-notok-spheres-lo}{1.14}
\aznset{gt-notok-over-gtb}{5.9}
\aznset{tr-full-epsU}{0.306}
\aznset{tr-full-S}{0.679}
\aznset{gt-full-epsU}{0.152}
\aznset{gt-full-epsU-pct}{15.2}
\aznset{gt-full-epsU-near}{0.179}
\aznset{gt-full-epsp}{0.427}
\aznset{gt-full-S}{0.290}
\aznset{gt-full-below-fs}{86}
\aznset{gt-full-epsp-below-fs}{94}
\aznset{gt-full-crossflow}{0.253}
\aznset{gt-full-dual}{0.396}
\aznset{isla-full-epsU}{0.082}
\aznset{isla-full-epsU-pct}{8.2}
\aznset{isla-full-epsU-near-pct}{11.2}
\aznset{isla-full-epsp}{0.196}
\aznset{isla-full-epsp-notop5}{0.109}
\aznset{isla-full-S}{0.153}
\aznset{isla-full-below-fs}{98}
\aznset{isla-full-epsp-below-fs}{99}
\aznset{isla-ho-full-epsU}{0.082}
\aznset{isla-ho-full-S}{0.186}
\aznset{isla-joint-epsp}{0.124}
\aznset{isla-joint-S}{0.123}
\aznset{isla-joint-p-drop-pct}{37}
\aznset{gt-over-isla-S-2dp}{1.90}
\aznset{gt-over-isla-S-lo}{1.87}
\aznset{gt-over-isla-S-hi}{1.94}
\aznset{gt-over-isla-epsU-2dp}{1.86}
\aznset{gt-over-isla-epsU-lo}{1.83}
\aznset{gt-over-isla-epsU-hi}{1.90}
\aznset{gt-over-isla-epsp-2dp}{2.17}
\aznset{gt-over-isla-epsp-lo}{2.10}
\aznset{gt-over-isla-epsp-hi}{2.25}
\aznset{pooled-body-gt-isla-S}{1.344}
\aznset{pooled-body-gt-isla-S-lo}{1.208}
\aznset{pooled-body-gt-isla-S-hi}{1.495}
\aznset{pooled-body-gt-isla-epsU}{1.503}
\aznset{pooled-body-gt-isla-epsU-lo}{1.337}
\aznset{pooled-body-gt-isla-epsU-hi}{1.689}
\aznset{pooled-body-gt-isla-epsp}{1.174}
\aznset{pooled-body-gt-isla-epsp-lo}{1.051}
\aznset{pooled-body-gt-isla-epsp-hi}{1.312}
\aznset{isla-mesh-ratio}{6}
\aznset{gt-mesh-ratio}{11}
\aznset{isla-near-mesh-ratio}{4}
\aznset{isla-sphere-pct}{2.6}
\aznset{isla-open-pct}{3.0}
\aznset{headroom-ratio-error-lo}{2}
\aznset{headroom-ratio-error-hi}{4}
\aznset{val-fp32-epsU}{0.157}
\aznset{val-bf16-epsU}{0.161}
\aznset{abl-geom-S}{0.997}
\aznset{abl-type-S}{0.338}
\aznset{abl-type-dual}{0.519}
\aznset{abl-notok-S}{0.369}
\aznset{abl-notok-crossflow}{0.499}
\aznset{isla-geom-gain-hi}{1.0}
\aznset{gt-geom-gain-lo}{0.9}
\aznset{gt-geom-gain-hi}{1.1}
\aznset{isla-newbc-gain-lo}{3.8}
\aznset{isla-newbc-gain-hi}{8.4}
\aznset{gt-newbc-gain-lo}{2.4}
\aznset{gt-newbc-gain-hi}{5.9}
\aznset{isla-newbc-S-lo}{0.77}
\aznset{isla-newbc-S-hi}{1.09}
\aznset{gt-newbc-S-lo}{0.90}
\aznset{gt-newbc-S-hi}{1.74}
\aznset{fs-full-epsU-opposed_hemispheres}{0.041}
\aznset{fs-full-epsU-dual_inlet}{1.586}
\aznset{fit-si-plain-train-U-lo}{0.98}
\aznset{fit-si-plain-train-U-hi}{1.00}
\aznset{fit-scaled-plain-train-U-lo}{0.11}
\aznset{fit-scaled-plain-train-U-hi}{0.15}
\aznset{fit-scaled-plain-val-U-lo}{0.57}
\aznset{fit-scaled-plain-val-U-hi}{0.62}
\aznset{fit-scaled-plain-val-p-lo}{2.03}
\aznset{fit-scaled-plain-val-p-hi}{5.18}
\aznset{fit-scaled-plain-best-epoch-lo}{31}
\aznset{fit-scaled-plain-best-epoch-hi}{46}
\aznset{fit-scaled-plain-gap-lo}{4}
\aznset{fit-scaled-plain-gap-hi}{6}
\aznset{fit-turb-train-share-pct}{64}
\aznset{fit-turb-val-share-pct}{68}
\aznset{fit-omega-train-share-pct}{32}
\aznset{fit-p-val-share-pct}{17}
\aznset{aux-fs-k}{0.0110}
\aznset{aux-fs-nut-m}{2.95}
\aznset{aux-gt-k}{0.0089}
\aznset{aux-gt-nut-m}{1.19}
\aznset{aux-isla-k}{0.0070}
\aznset{aux-isla-nut-m}{0.70}
\aznset{aux-isla-rel-gamma}{0.050}
\aznset{aux-isla-rel-retheta}{0.103}
\aznset{ho-re-gt-gain}{1.3}
\aznset{ho-re-isla-gain}{1.8}

\aznset{gs-cases}{nine}
\aznset{gs-released}{seven}
\aznset{gs-rp-max}{0.04}
\aznset{gs-rg-cases}{eight}
\aznset{gs-rg-median}{3.7}
\aznset{gs-rg-max}{19}
\aznset{gs-urg-median}{1.2}
\aznset{gs-urg-mean}{1.4}
\aznset{gs-urw-mean}{1.2}
\aznset{gs-unear-rg-mean}{2.5}
\aznset{gs-unear-rg-max}{5.3}
\aznset{gs-unear-rw-mean}{2.7}
\aznset{gs-unear-rw-max}{6.4}
\aznset{gs-urg-mean-released}{1.5}
\aznset{gs-self-interp-min}{0.2}
\aznset{gs-self-interp-max}{1.5}
\aznset{gs-friction-rg-max}{0.2}
\aznset{gs-extrap-min}{1.4}
\aznset{gs-extrap-max}{23}
\aznset{gs-rw-median}{6.9}
\aznset{gs-rw-max}{36}
\aznset{gs-plate-friction-ratio}{2.25}
\aznset{gs-core-hours-k}{30.1}
\aznset{gs-max-core-hours}{400}
\aznset{gs-pool}{196}
\aznset{gs-pool-pct}{5}
\aznset{gs-rw-max-unfloored-metric}{6.9}
\aznset{gs-rg-median-unfloored}{7.0}
\aznset{gs-rw-median-unfloored}{8.6}
\aznset{gs-rw-max-unfloored}{44}
\aznset{gs-small-force-cases}{three}
\aznset{gs-cost-ratio-wr}{3.2}
\aznset{gs-urg-min}{0.2}
\aznset{gs-urg-max}{4.0}
\aznset{gs-plate-force-ratio}{1.8}
\aznset{gs-ratio-realized-lo}{1.3}
\aznset{gs-ratio-realized-hi}{1.5}
\aznset{gs-re-min}{1.9\times10^{5}}
\aznset{gs-re-max}{1.4\times10^{6}}
\aznset{gs-extrap-cases}{five}
\aznset{gs-extrap-median}{9.2}
\aznset{gs-extrap-factor-lo}{1.5}
\aznset{gs-extrap-factor-hi}{3.4}
\aznset{gs-urw-median}{0.7}
\aznset{gs-rg-released-cases}{six}
\aznset{gs-rw-median-released}{3.5}
\aznset{gs-rw-median-unfloored-released}{4.6}

\aznset{sat-Lirr}{0.0309}
\aznset{sat-Lirr-lo}{0.0247}
\aznset{sat-Lirr-hi}{0.0365}
\aznset{sat-beta}{0.61}
\aznset{sat-fit-rms}{2.1}
\aznset{sat-fit-cells}{12}
\aznset{sat-fit-N1}{20}
\aznset{sat-fit-N2}{35}
\aznset{sat-noD12-E}{0.0345}
\aznset{sat-noD12-beta}{0.76}
\aznset{sat-Lirr-joint}{0.0318}
\aznset{sat-Lirr-joint-lo}{\ensuremath{<10^{-4}}}
\aznset{sat-Lirr-joint-hi}{0.0356}
\aznset{sat-alpha-joint}{3.0}
\aznset{sat-joint-cells}{15}
\aznset{sat-Lirr-all}{0.0338}
\aznset{sat-Lirr-C}{0.0327}
\aznset{sat-runs-C}{57}
\aznset{sat-floor400-C}{0.0380}
\aznset{sat-best}{0.0393}
\aznset{sat-within}{21}
\aznset{sat-cells}{50}
\aznset{sat-cells-conv}{15}
\aznset{sat-Nmax}{354}
\aznset{sat-runs}{483}
\aznset{sat-test-cars}{48}
\aznset{sat-val-cars}{35}
\aznset{sat-pool}{400}
\aznset{sat-commit}{e3dae21a}
\aznset{sat-gpuh}{17{,}913}
\aznset{sat-N6-400-k1-B16}{0.0416}
\aznset{sat-N6-400-k4-B16}{0.0399}
\aznset{sat-N7-400-k1-B16}{0.0412}
\aznset{sat-N7-400-k4-B16}{0.0397}
\aznset{sat-N8-400-k1-B16}{0.0407}
\aznset{sat-N9-400-k1-B16}{0.0418}
\aznset{sat-seed-spread}{3.5}
\aznset{sat-Lone}{0.0086}
\aznset{sat-Lone-ratio}{0.28}
\aznset{sat-triangle-sq-share}{52}
\aznset{sat-nonspatial-sq-share}{92}
\aznset{sat-gap}{0.0223}
\aznset{sat-strict}{0.0118}
\aznset{sat-strict-ratio}{0.38}
\aznset{sat-triangle-strict-sq-share}{38}
\aznset{sat-nonspatial-strict-sq-share}{85}
\aznset{sat-noise}{0.017}
\aznset{sat-noise-tau-lo}{3}
\aznset{sat-noise-tau-hi}{30}
\aznset{sat-noise-lo}{0.0079}
\aznset{sat-noise-hi}{0.025}
\aznset{sat-noise-R}{0.156}
\aznset{sat-noise-T-lo}{1.7}
\aznset{sat-noise-T-hi}{4.3}
\aznset{sat-noise-tau}{14}
\aznset{sat-noise-wheel-faces}{67}
\aznset{sat-noise-wheel-sigma}{88}
\aznset{sat-noise-share}{30}
\aznset{sat-bracket-ratio}{1.8}
\aznset{sat-size-spread-400}{2.6}
\aznset{sat-size-spread-400-ceil}{3}
\aznset{sat-k4-gain-200-lo}{5.5}
\aznset{sat-k4-gain-200-hi}{8.1}
\aznset{sat-k4-gain-400-lo}{3.8}
\aznset{sat-k4-gain-400-hi}{4.1}
\aznset{sat-k4-gain-400-about}{4}
\aznset{sat-doubling-gain-100-200}{11}
\aznset{sat-doubling-gain-200-400}{12}
\aznset{sat-power-cells}{six}
\aznset{sat-power-exponent}{0.18}
\aznset{sat-power-rms}{0.62}
\aznset{sat-floor-rms-same-cells}{1.3}

\aznset{stackup-rerun-u-mean-pct}{0.007}
\aznset{stackup-averaging-u-mean-pct}{0.34}

\aznset{posing-cx-open}{0.026}
\aznset{posing-cx-blockage-pct}{13}
\aznset{posing-cx-dual}{0.55}
\aznset{posing-epsu-oneinlet-lo}{3}
\aznset{posing-epsu-oneinlet-hi}{7}
\aznset{posing-epsu-dual-lo}{152}
\aznset{posing-epsu-dual-hi}{177}

\aznset{anatomy-alpha}{1.7}
\aznset{anatomy-beta}{5.4}
\aznset{anatomy-re}{3.5\times10^{6}}
\aznset{anatomy-cells-m}{4.2}

\aznset{pred-open-tok-u}{0.056}
\aznset{pred-open-fs-u}{0.047}
\aznset{pred-open-tok-unear}{0.10}
\aznset{pred-open-fs-unear}{0.15}
\aznset{pred-open-far-pct}{93}
\aznset{pred-open-isla-u}{0.025}

\icmltitlerunning{AeroZoo: Well-Posed Flow Problems for Foundation Models of Aerodynamics}

\begin{document}

\twocolumn[
  \icmltitle{AeroZoo: A Benchmark of Well-Posed Flow Problems\texorpdfstring{\\}{ }%
    for Foundation Models of Aerodynamics}

  \icmlsetsymbol{lead}{*}
  \begin{icmlauthorlist}
    \icmlauthor{Peter Sharpe}{lead,nv}
    \icmlauthor{Carmelo Gonzales}{lead,nv}
    \ificmlshowauthors\par\vspace{3pt}\fi
    \icmlauthor{Corey Adams}{nv}
    \icmlauthor{Mohammadmehdi Ataei}{nv}
    \icmlauthor{Sanjay Choudhry}{nv}
  \end{icmlauthorlist}

  \icmlaffiliation{nv}{NVIDIA, Santa Clara, California, USA}
  \icmlcorrespondingauthor{Peter Sharpe}{psharpe@nvidia.com}

  \icmlkeywords{computational fluid dynamics, neural surrogates, foundation models,
    benchmark, dataset, boundary-value problems, RANS, out-of-distribution generalization}

  \vskip 0.3in
]

\printAffiliationsAndNotice{\textsuperscript{*}Lead authors. }

\begin{abstract}
Neural surrogates for computational fluid dynamics (CFD) are trained and ranked on single-family,
fixed-domain datasets that cannot test the minimum bar for an aerodynamic foundation model:
predicting CFD-quality incompressible flow in seconds for new shapes in new enclosures. On
DrivAerML, larger models and longer training barely lower Transolver's error; doubling training
cases cuts it by about \azn{sat-doubling-gain-200-400}\% up to \azn{sat-pool} cases, exhausting
the dataset. We introduce \aerozoo{}, a benchmark and training corpus of \azn{complete} time-mean
unsteady Reynolds-averaged Navier--Stokes (URANS) solutions with a transition model across seven
geometry families, eight enclosures, and Reynolds numbers from $10^3$ to $10^8$. Each sample poses
a well-posed PDE problem as a solver receives it: boundary meshes grouped by condition type,
prescribed values, and fluid properties. The \azn{stable} time-mean-stable samples form the
canonical benchmark: two tasks, mean engineering metrics with an overall flow score
$S_\mathrm{flow}$ ($1.000$ for uniform inlet flow), and \azn{splits} frozen geometry-disjoint
splits holding out shapes, enclosures, boundary-condition layouts, and Reynolds ranges. On the
random split, boundary-condition tokens cut GeoTransolver's mean velocity error from
\azn{gt-notok-epsU-pct}\% to \azn{gt-full-epsU-pct}\% of inlet speed (uniform flow:
\azn{fs-full-epsU-pct}\%; $S_\mathrm{flow} = \azn{gt-full-S}$). ISLA, a boundary-driven
transformer, reaches \azn{isla-full-epsU-pct}\% ($S_\mathrm{flow} = \azn{isla-full-S}$;
\azn{isla-joint-S} with boundary targets), about \azn{isla-mesh-ratio} times mesh-refinement
sensitivity. On shared test samples, holding out a geometry family barely changes ISLA's mean
velocity error, whereas holding out enclosures or boundary-condition layouts raises it
\azn{isla-newbc-gain-lo} to \azn{isla-newbc-gain-hi} times. The dataset
\azpublic{is public}{will be released}; the scoring code
\azcodepublic{is public}{will be released}.
\end{abstract}

\section{Introduction}
\label{sec:introduction}

\begin{figure}[t]
  \centering
  \includegraphics[width=\columnwidth]{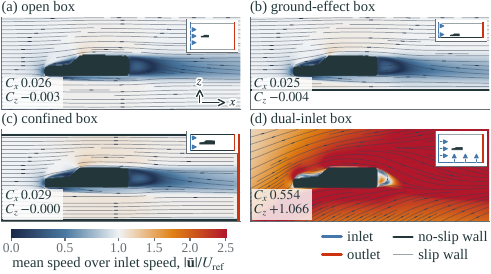}
  \caption{One car in four enclosures yields four distinct flows at the same inlet speed $\Uref$ and
    Reynolds number: (a) open, (b) ground-effect, (c) confined, and (d) dual-inlet boxes
    (center-plane speed with streamlines; insets show full domain boundaries; $(x, z)$ triad in (a)
    marks $x$ along the primary inlet and $z$ vertical). Within one car length of the body, the
    three single-inlet flows differ by \azn{posing-epsu-oneinlet-lo}--\azn{posing-epsu-oneinlet-hi}\%
    of $\Uref$ (more than mesh refinement; \cref{sec:trust}), no-slip tunnel walls in (c) raise drag
    by \azn{posing-cx-blockage-pct}\%, and the second inlet in (d) changes near-body velocity by
    \azn{posing-epsu-dual-lo}--\azn{posing-epsu-dual-hi}\% and raises the $x$-force coefficient
    $C_x$ (normalized by $\Lref^2$) from \azn{posing-cx-open} to \azn{posing-cx-dual}.}
  \label{fig:posing}
\end{figure}

\begin{figure*}[t]
  \centering
  \includegraphics[width=\textwidth]{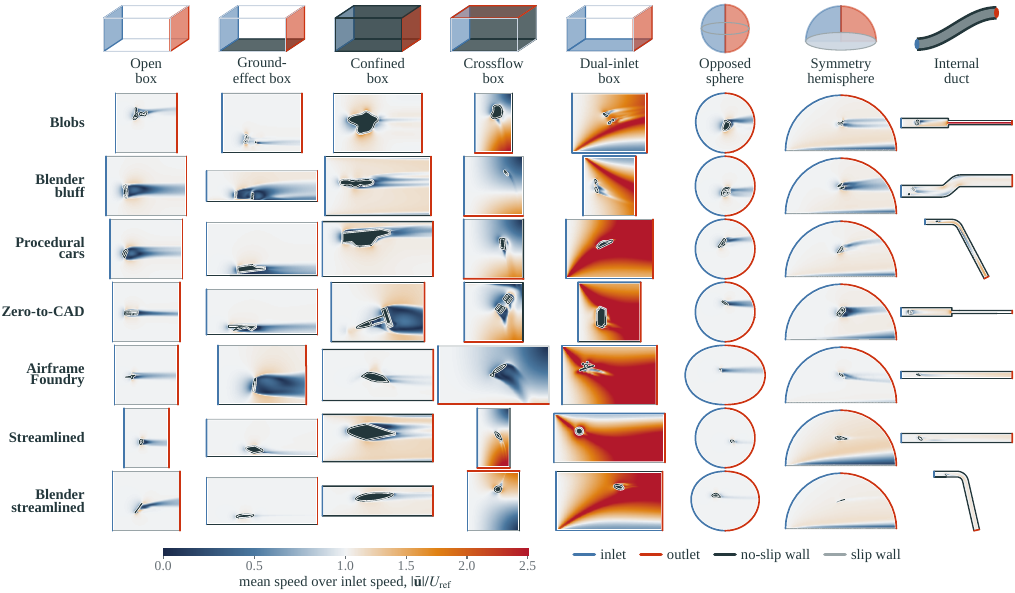}
  \caption{The zoo: every tile is a distinct geometry posed as its own boundary-value problem.
    Columns show the eight enclosure configurations (sketched above by boundary type; opposed
    spheres may be smooth or lumpy) and rows the seven geometry families, displaying the first
    \key{mean_stable} (\cref{sec:dataset}) sample per cell (with a body for ducts), cut along the
    primary inlet over its full fluid domain (colorbar linear on 0--1 and 1--2.5).}
  \label{fig:zoo}
\end{figure*}

\begin{table*}[t]
  \centering
  \caption{\aerozoo{} spans seven geometry families and eight enclosure configurations, whereas most
    public 3D CFD datasets vary one design family and several use costlier scale-resolving solvers
    (\cref{app:terms,app:datasets}; \textcolor{gray}{--}: not stated).}
  \label{tab:datasets}
  \scriptsize
  \setlength{\tabcolsep}{4pt}
  \renewcommand{\arraystretch}{0.90}
  \begin{tabular}{@{}lcclllr@{}}
    \toprule
    Dataset & Families & Enclosures & Operating points & Solver & Samples & Cells per sample \\
    \midrule
    \textbf{\aerozoo{}} & \textbf{7} & \textbf{8} & \textbf{$\boldsymbol{\Rey}$ $\boldsymbol{10^3}$--$\boldsymbol{10^8}$, continuous} & URANS mean, transition & \azn{complete} & \azn{cells-min-m}--\azn{cells-max-m}\,M \\
    \midrule
    DrivAerML~\citep{ashton2024drivaerml} & 1 & 1 & one & hybrid RANS--LES & 500 (484 public) & 160\,M \\
    WindsorML~\citep{ashton2024windsorml} & 1 & 1 & one & wall-modeled LES & 355 & 280--300\,M \\
    AhmedML~\citep{ashton2024ahmedml} & 1 & 1 & one & DDES & 500 & 15--20\,M \\
    DrivAerNet++~\citep{elrefaie2024drivaernetpp} & 1 & 1 & one & RANS & \num{8000} & 24\,M \\
    SHIFT-SUV~\citep{luminary2025shiftsuv} & 1 & \textcolor{gray}{--} & two vehicle scales & DES & $> $\num{1000} & \textcolor{gray}{--} \\
    DrivAerStar~\citep{qiu2025drivaerstar} & 1 & 1 & one & RANS & \num{12000} & 12\,M \\
    ShapeNet-Car~\citep{umetani2018learning} & 1 & 1 & one & URANS mean & 889 & \textcolor{gray}{--} \\
    \addlinespace[3pt]
    HiLiftAeroML~\citep{ashton2026hiliftaeroml} & 1 & 1 & 10 incidences & wall-modeled LES & \num{1800} & 300--500\,M \\
    BlendedNet~\citep{sung2025blendednet} & 1 & 1 & about nine per body & RANS & \num{8830} & 9--14\,M \\
    SuperWing~\citep{yang2026superwing} & 1 & 1 & Mach, incidence & RANS & \num{28856} & \textcolor{gray}{--} \\
    Emmi-Wing~\citep{paischer2025emmiwing} & 1 & 1 & one per wing & RANS & \num{29727} & \textcolor{gray}{--} \\
    \addlinespace[3pt]
    FlowBench, 3D part~\citep{tali2024flowbench} & 2 & 1 & $\Rey$ $10$--$10^3$ & DNS & 500 & \textcolor{gray}{--} \\
    WindTunnel-20k~\citep{inductiva2024windtunnel} & \textcolor{gray}{--} & 1 & four speeds, five yaw angles & steady RANS, 300 iterations & \num{19812} & \textcolor{gray}{--} \\
    \bottomrule
  \end{tabular}
\end{table*}

Neural surrogates for computational fluid dynamics (CFD) aim to predict in seconds flows and loads
that solvers compute in hours (\cref{app:terms} defines CFD terms). Most 3D datasets sweep one
design in a fixed domain and boundary-condition layout (\cref{tab:datasets}), such as 500 sedan
morphs at one Reynolds number in DrivAerML~\citep{ashton2024drivaerml}. These benchmarks made 3D
surrogates practical and suit single-product optimization.

On a fixed design family, however, scaling models and compute hits diminishing returns once
geometry count bottlenecks accuracy. On DrivAerML, scaling Transolver~\citep{wu2024transolver}
from \azn{sat-fit-N1} to \azn{sat-Nmax} million parameters at \azn{sat-pool} training cases changes
surface-pressure error by under \azn{sat-size-spread-400-ceil}\%, and quadrupling compute lowers it
by about \azn{sat-k4-gain-400-about}\%. Only adding cases still lowers error, by about
\azn{sat-doubling-gain-200-400}\% per doubling, toward a dataset-level floor bracketed between
target averaging noise and the scaling asymptote (\cref{app:saturation}).\reviewonly{ \rvTransfer}

An aerodynamic foundation model must instead predict accurate flows across new geometries,
enclosures, and operating points under any consistent boundary conditions. \aerozoo{} sets the
minimum bar for wide industrial use: CFD quality in seconds for incompressible flow at industrial
Reynolds numbers. Even in distribution, current surrogates miss their targets by several times
solver mesh sensitivity (\cref{sec:trust}). Measuring generalization faces two obstacles:
\begin{enumerate}[leftmargin=*,itemsep=1pt,topsep=2pt]
\item \emph{Breadth.} Holding out another morph of the same family tests interpolation rather than
  out-of-distribution generalization.
\item \emph{Posing.} Existing datasets omit fixed boundary conditions, reducing learning to a
  shape-to-field map rather than the PDE solution operator~\citep{sharpe2025globe}. One shape
  yields distinct flows above a road, inside a tunnel, or beside a second inlet (\cref{fig:posing});
  autoregressive models that roll flows forward like video frames lack this specification too.
\end{enumerate}

We introduce \aerozoo{} (\cref{fig:zoo}), \azn{complete} solved flow problems built to address both.
Every sample poses a distinct geometry from seven families of blobs, CAD parts, cars, aircraft, and
streamlined bodies in one of eight enclosures, plus \azn{body-free-ducts} body-free ducts, across
sizes of 2\,cm to 20\,m and Reynolds numbers from $10^3$ to $10^8$ under one frozen numerical
recipe. Each sample packages the complete boundary-value problem: boundary meshes grouped by
condition type, face-wise prescribed values, and fluid properties. As a benchmark and pretraining
corpus, \aerozoo{} contributes:
\begin{enumerate}[leftmargin=*,itemsep=0pt,topsep=2pt]
  \item \azn{complete} well-posed problems (\azn{cells-total-b}\,B cells) with stability labels and
    an error stack-up (\cref{sec:dataset,sec:trust});
  \item a \key{DomainMesh} schema and input rule separating PDE inputs from scoring scales (\cref{sec:schema});
  \item interior and boundary tasks, volume- and area-weighted mean metrics with overall flow score
    $S_\mathrm{flow}$, and \azn{splits} frozen splits (\cref{sec:benchmark}); and
  \item evaluations showing that boundary tokens cut GeoTransolver's mean velocity error
    \azn{gt-notok-over-gtb}-fold on the random split, that ISLA~\citep{sharpe2026isla} cuts it
    nearly in half again ($S_\mathrm{flow}=\azn{isla-full-S}$, or \azn{isla-joint-S} with boundary
    targets), and that both models degrade most on unseen enclosures and boundary-condition layouts
    (\cref{sec:baselines,sec:trust}).
\end{enumerate}

\section{A well-posed sample}
\label{sec:schema}

\begin{figure*}[t]
  \centering
  \begin{minipage}[c]{0.38\textwidth}
    \centering
    \includegraphics[width=\linewidth]{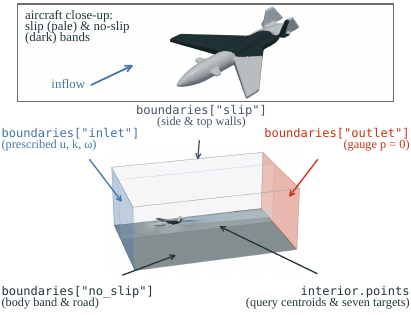}\\[-1pt]
    {\small (a) an annotated \key{DomainMesh} sample}
  \end{minipage}\hfill
  \begin{minipage}[c]{0.60\textwidth}
    \centering
    \footnotesize
    \setlength{\tabcolsep}{3pt}
    \renewcommand{\arraystretch}{1.00}
    \begin{tabular}{@{}>{\raggedright\arraybackslash}p{0.11\linewidth}>{\raggedright\arraybackslash}p{0.62\linewidth}>{\raggedright\arraybackslash}p{0.23\linewidth}@{}}
      \toprule
      Role & Tensors in each sample & Enters model? \\
      \midrule
      Input & four boundary meshes (\key{inlet}, \key{outlet}, \key{no_slip}, \key{slip}),
        face-wise \key{prescribed/*}, plus \key{kinematic_viscosity} and \key{density} & yes: poses PDE \\
      Query & \key{interior.points} (\azn{anatomy-cells-m}\,M cells here); boundary centroids & yes: query sites \\
      \midrule
      Target & seven mean fields ($\bar\vu, \bar p$, and five turbulence and transition fields
        $\bar k, \bar\omega, \bar\nu_t, \bar\gamma, \overline{\Rey}_{\theta t}$); wall shear on
        \key{no_slip} & no: predicted \\
      Weight & cell volumes (boundary face areas follow from triangle meshes) & no: loss/metrics \\
      Fact & case identity, window bounds, stability label, \key{body_id}, near-wall spacing $y^+$ & no: audit/splits \\
      Scale & \key{reference/*} ($\Lref, \Uref, \qref$, primary inlet vector, force axis $\mathbf e_F$) & no: scoring only \\
      \bottomrule
    \end{tabular}\\[3pt]
    {\small (b) declared tensor roles}
  \end{minipage}
  \caption{Every tensor has one declared role, and only PDE inputs and queries enter a model.
    (a) Airframe Foundry aircraft in a ground-effect box ($\Rey=\azn{anatomy-re}$,
    $\azn{anatomy-alpha}^\circ$ incidence, $\azn{anatomy-beta}^\circ$ sideslip; three slip/no-slip
    bands enlarged above; translucent speed cut on \cref{fig:zoo}'s scale). (b) Scales and metadata
    are withheld for scoring and auditing (\cref{app:schema}).}
  \label{fig:schema}
\end{figure*}

A learning task is \emph{well-posed} when its inputs determine its target, extending
\citet{sharpe2025globe} in the spirit of Hadamard~\citep{hadamard1902problemes}. An incompressible
RANS problem with a fixed closure is posed by three items: domain boundaries, each part's
boundary-condition type and prescribed values, and fluid density and kinematic viscosity. A solver
receives these and no more. For a learned model, the input is therefore geometry together with
boundary conditions and fluid (\cref{fig:posing}). Omitting conditions held fixed limits models to
those conditions; omitting varying conditions maps identical inputs to different targets, so no
model can fit both. \Cref{fig:schema} divides each sample's tensors by role.

\paragraph{Boundaries and their conditions.}
Parts of the boundary sharing one condition type merge into four triangle meshes: \key{inlet}
(prescribed velocity and turbulence), \key{outlet} (zero gauge pressure plus
\key{prescribed/reverse_flow/*} where flow re-enters), \key{no_slip} walls, and \key{slip} walls
(including symmetry planes). Because both bodies and enclosure walls may be slip or no-slip,
condition types cannot be inferred from shape. Each mesh stores its face-wise prescribed values. A
bent duct or dual-inlet box has no single inlet vector, so any model using one must compute it.

\paragraph{What a model may use.}
Only two scalars in \key{global_data} are inputs: kinematic viscosity $\nu$ and density $\rho$ (for
pressure and shear in pascals). Reference scales in \key{global_data/reference} ($\Lref, \Uref,
\qref=\rho\Uref^2/2$, and force axis $\mathbf e_F$) normalize metrics (\cref{sec:benchmark}) and are
withheld: because a wing's Reynolds number can use chord or span, handing $\Lref$ to a model would
leak dataset conventions. Under the input rule:
\begin{itemize}[leftmargin=*,itemsep=0pt,topsep=1pt]
  \item A model may compute any function of one sample's inputs: geometric features, a cheap solver
    (cost reported), or boundary-derived \emph{per-case scales} such as length, speed, and Reynolds
    number (\texttt{derive\_globals}, \cref{app:metrics}). Rescaling coordinates by a derived length
    requires passing that length or its Reynolds number so differently sized cases stay distinct.
  \item Withheld from inputs: targets, cell volumes, \key{body_id}, all other \key{global_data}
    keys, and any statistic fitted on test samples.
\end{itemize}

\paragraph{Physical units.}
Samples use SI units. Drawing length, speed, and Reynolds number independently spans
\azn{nu-decades} decades of viscosity $\nu$ (\azn{nu-air-water-pct}\% between water and air;
\cref{app:statistics}), so geometrically similar flows appear at different physical scales. This
prevents memorizing a fixed fluid constant and tests dimensional similarity (Buckingham's $\Pi$
theorem; \citealp{bakarji2022dimensionally,villar2023dimensionless,sharpe2025globe}).

\paragraph{Interior points and cell volumes.}
Interior targets are stored at finite-volume cell centroids, serving as query coordinates directly on
native cells without interpolation. Because point density reflects wake and wall-layer refinement,
models may also be evaluated in an independent-query mode that decodes each point from the
boundaries alone, preventing cues from neighboring query density (\cref{sec:baselines}). Cell volumes
span decades, weight metrics and optionally training losses, and are withheld at inference. Each
sample is a PhysicsNeMo \key{DomainMesh}~\citep[\cref{app:schema}]{physicsnemo}:
\begin{pycode}[unbreakable]
from physicsnemo.mesh import DomainMesh
sample = DomainMesh.load(path)   # lazy
inlet = sample.boundaries["inlet"]
bc = inlet.cell_data["prescribed", "dirichlet"]
u_in = bc["velocity"]            # [F, 3] m/s
physics = sample.global_data["physics"]
nu = physics["kinematic_viscosity"]  # m^2/s
x = sample.interior.points       # [N, 3] m
fields = sample.interior.point_data["solution"]
u = fields["velocity"]           # [N, 3] m/s
\end{pycode}

\section{The dataset}
\label{sec:dataset}

Every sample poses a different body, or body-free duct, in its own enclosure at its own size,
speed, and viscosity. Each planned \emph{case} is determined by one seed and a frozen recipe; each
solved, released case is a \emph{sample}, and failed cases are recorded without replacement
(\cref{app:recipe}).

\begin{figure}[t]
  \centering
  \includegraphics[width=\columnwidth]{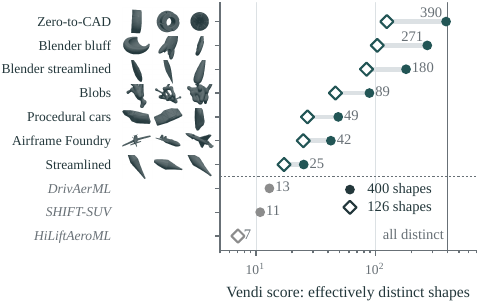}
  \caption{Every \aerozoo{} geometry family (three examples at left) exceeds single-design sweeps
    (gray) in Vendi score (effective distinct shapes)~\citep{friedman2023vendi} at 400 (dots) and
    126 shapes (diamonds, matching HiLiftAeroML's training geometries; \cref{app:datasets}).}
  \label{fig:vendi}
\end{figure}

\paragraph{What varies.}
Each case draws three independent ingredients: a body from seven geometry families
(\cref{fig:vendi}), one of eight enclosures (\cref{tab:configs}), and length, speed, and Reynolds
number setting viscosity (\cref{tab:draws}). Six geometry families are procedural; Zero-to-CAD
samples a public CAD corpus~\citep{ataei2026zerotocad,adsk2026zerotocad100k}. Cases are randomly
rotated in 3D with mostly random body orientation, so flow direction cannot be inferred from shape.
With probability 0.03, a scaled copy (0.25--1$\times$) adds a second body. Two thirds of bodies
split along parallel planes into two or three slip and no-slip bands (\cref{fig:schema}a,
\cref{tab:draws}); under uniform irrotational inflow, no-slip walls generate all vorticity, testing
whether models locate boundary-layer onset from wall types.

\paragraph{Breadth in shapes and flows.}
Under the Vendi score~\citep{friedman2023vendi} ($n$ identical shapes score 1; $n$ distinct score
$n$), even the least varied \aerozoo{} family scores at least \azn{vendi-ratio-min-k-lo} times the
more varied automotive dataset across bandwidths, and Zero-to-CAD reaches \azn{vendi-ceiling-pct}\%
distinct shapes (\cref{fig:vendi,app:datasets}). Flow regimes span laminar to turbulent: below
$\Rey=5\times10^4$, \azn{laminar-low-share-pct}\% of bodies are laminar over $\ge98\%$ of their
area; above $2\times10^5$, the median laminar share is \azn{laminar-high-median-pct}\%, and median
skin friction falls \azn{cf-median-fall}-fold across the five decades of $\Rey$
(\cref{fig:regimes}).

\paragraph{Meshing and solving.}
The \texttt{snappyHexMesh} mesher of OpenFOAM~\citep{weller1998tensorial,openfoam2026userguide}
builds multi-million-cell meshes refined on the body and along its wake, sizing wall layers before
the flow is known from flat-plate friction at the case's $\Rey$. Because many flows stay unsteady,
OpenFOAM~14 integrates the incompressible unsteady RANS (URANS) equations with the Langtry--Menter
transition SST model~\citep{menter1994two,menter2003ten,langtry2009correlation} and wall functions
(algebraic laws for near-wall shear), prescribing drawn inlet turbulence intensity and length scale,
for 15 flow-through (convective) times $\Tc=\Lref/\Uref$ of warm-up and 80 of averaging
(\cref{app:recipe}). Prescribed inflow eddy viscosity sets a median outer-wake effective Reynolds
number of \azn{re-eff-median} ($10^2$--$10^4$), whereas molecular viscosity $\nu$ governs near-wall
shear. Of \azn{planned} planned cases, \azn{complete} (\azn{complete-pct}\%) completed in
\azn{bytes-total-tb}\,TB; \azn{failed} failed checks and \azn{timed-out} timed out
(\azn{timed-out-top-decade} in the top $\Rey$ decade).

\paragraph{The stability label.}
Comparing the two halves of each averaging window labels a sample \key{mean_stable} if they agree
within $0.02\,\Uref$ in RMS velocity, $0.05\,\qref$ in RMS pressure, and 2\% in force coefficient
$C_F$ (\cref{sec:benchmark,app:stability}), and \key{mean_unstable} otherwise. Of \azn{complete}
samples, \azn{stable} (\azn{stable-share}\%) are \key{mean_stable}: \azn{stable-pct-other-min} to
\azn{stable-pct-other-max}\% in five configurations, \azn{stable-pct-crossflow}\% in crossflow
boxes, \azn{stable-pct-opposed_hemisphere_symmetry}\% in symmetry hemispheres, and
\azn{stable-pct-internal_duct}\% in ducts (\cref{fig:stability}). Both sets are released; the
canonical benchmark uses only \key{mean_stable} samples, and the label is computed from the
solution so inputs never reveal it.

\section{The benchmark}
\label{sec:benchmark}

Every \aerozoo{} evaluation pairs a task, metric, and split against a training-free reference.

\paragraph{Two tasks.}
Both tasks provide the posed problem (\cref{sec:schema}) for the time-mean solution. The
\emph{interior task} predicts seven mean fields at cell centroids (median \azn{cells-median-m}\,M
points): velocity and pressure are headline targets, and five turbulence fields (\cref{fig:schema})
are optional auxiliaries (\cref{app:metrics}). The \emph{boundary task} predicts boundary loads
without interior points: pressure on all four boundaries, wall shear on \key{no_slip}, and velocity
on \key{outlet} (median \azn{boundary-faces-median-m}\,M faces). Because \key{body_id} is withheld,
models predict across all boundary meshes; metrics score body pressure and wall shear plus duct
inlet pressure and outlet velocity, never scoring fields fixed by a prescribed (Dirichlet)
condition.

\paragraph{Metrics.}
Metrics use each sample's scoring scales (\cref{sec:schema}). With cell volumes $V_i$ and a hat on
each predicted value, the interior task is scored by
\begin{gather}
  \varepsilon_U=\frac{1}{\Uref}\biggl(\frac{\sum_i V_i\lVert\hatvu_i-\vu_i\rVert^2}{\sum_i V_i}\biggr)^{1/2},\nonumber\\
  \varepsilon_p=\frac{1}{\qref}\biggl(\frac{\sum_i V_i(\hat p_i-p_i)^2}{\sum_i V_i}\biggr)^{1/2},
  \label{eq:interior}
\end{gather}
over the whole fluid and over the near field within $\Lref$ of a body face centroid
($\varepsilon_U^{\mathcal N}$, $\varepsilon_p^{\mathcal N}$). Because median fluid volume spans a
factor of \azn{vol-enclosure-ratio} (from \azn{vol-med-confined}\,$\Lref^3$ in confined boxes to
\azn{vol-med-hemisphere}\,$\Lref^3$ in symmetry hemispheres), large far fields dilute
$\varepsilon_U$, whereas near-field errors weight body flow uniformly. \Cref{app:metrics} adds
unweighted relative $L^2$ errors ($r_U, r_p$) and auxiliary turbulence errors.

The boundary task scores body faces $\mathcal B$ (selected via withheld \key{body_id}) and no-slip
subset $\mathcal W\subseteq\mathcal B$, with areas $A_j$, outward normals $\vn_j$, and wall shear
$\vtau_j$ (wall-on-fluid traction):
\begin{gather}
  \vC=\frac{1}{\qref\Lref^2}\Bigl(\sum_{j\in\mathcal B}p_jA_j\vn_j-\sum_{j\in\mathcal W}A_j\vtau_j\Bigr),\label{eq:force}\\
  \varepsilon_{C_F}=\frac{\lvert\hat C_F-C_F\rvert}{\max(\lvert C_F\rvert,\,0.05)},\quad
  \varepsilon_{\vC}=\frac{\lVert\hat\vC-\vC\rVert}{\max(\lVert\vC\rVert,\,0.05)},\nonumber
\end{gather}
where $C_F=\vC\cdot\mathbf e_F$ projects onto the fastest inlet (drag in straight flow) and
$\varepsilon_{\vC}$ scores the 3D load. \Cref{app:metrics} adds surface pressure, shear, and duct
loss errors.

\begin{table}[t]
  \centering
  \caption{Seven headline splits: three random sizes and four held-out regions (below rule),
    counting canonical (\key{mean_stable}) samples (\key{bank_cars} test counts body samples).
    Identifiers: \key{family} enclosure group, \key{bank} geometry family, \key{profile} box layout.}
  \label{tab:splits}
  \scriptsize
  \setlength{\tabcolsep}{2.8pt}
  \begin{tabular}{@{}llrrr@{}}
    \toprule
    Split & Identifier & Train & Val. & Test \\
    \midrule
    Random & \key{full} & \azn{split-full-train-canonical} & \azn{split-full-val-canonical} & \azn{split-full-test-canonical} \\
    Random, $1/6$ train & \key{scarce} & \azn{split-scarce-train-canonical} & \azn{split-scarce-val-canonical} & \azn{split-scarce-test-canonical} \\
    Random, $1/36$ train & \key{super_scarce} & \azn{split-super_scarce-train-canonical} & \azn{split-super_scarce-val-canonical} & \azn{split-super_scarce-test-canonical} \\
    \midrule
    Non-box enclosures & \key{family_non_box} & \azn{split-family_non_box-train-canonical} & \azn{split-family_non_box-val-canonical} & \azn{split-family_non_box-test-canonical} \\
    Dual-inlet boxes & \key{profile_dual_inlet} & \azn{split-profile_dual_inlet-train-canonical} & \azn{split-profile_dual_inlet-val-canonical} & \azn{split-profile_dual_inlet-test-canonical} \\
    $\Rey\ge10^6$ & \key{reynolds} & \azn{split-reynolds-train-canonical} & \azn{split-reynolds-val-canonical} & \azn{split-reynolds-test-canonical} \\
    Procedural cars & \key{bank_cars} & \azn{split-bank_cars-train-canonical} & \azn{split-bank_cars-val-canonical} & \azn{split-bank_cars-test-canonical} \\
    \bottomrule
  \end{tabular}
\end{table}

\paragraph{Splits.}
\aerozoo{} freezes \azn{splits} geometry-disjoint splits by hashing primary bodies (\cref{tab:splits};
all \azn{splits} in \cref{app:splits}):
\begin{itemize}[leftmargin=*,itemsep=0pt,topsep=1pt,parsep=0pt]
  \item \emph{Three random splits:} \key{full} ($\approx$70/10/20\%), \key{scarce} ($1/6$ train),
    and \key{super_scarce} ($1/36$ train).
  \item \emph{Nineteen held-out splits:} holding out one geometry family, enclosure (or all
    non-boxes), Reynolds range, or three-band bodies; body-property splits score body samples only.
\end{itemize}
Validation is for selection only. Because held-out test sets overlap \key{full}, pretraining on
\aerozoo{} may use only the scored split's training set, and reusing \key{full} settings must be
stated (\cref{app:splits}).

\paragraph{Primary overall score and reporting.}
Metrics are arithmetic means over the canonical test samples on which each is defined. The
training-free \emph{free-stream reference} predicts the area-weighted mean prescribed inlet
velocity and turbulence with zero gauge pressure and wall shear. To summarize velocity and
pressure on one scale without field-range dominance, we define the \emph{overall flow score}
\begin{equation}
  S_\mathrm{flow} = \exp\Bigl(\frac{1}{4}\sum_{i\in\{U,\,U^{\mathcal N},\,p,\,p^{\mathcal N}\}}
  \ln\frac{\bar\varepsilon_i}{\bar\varepsilon_i^{\mathrm{fs}}}\Bigr),
  \label{eq:composite}
\end{equation}
where $\bar\varepsilon_i$ and $\bar\varepsilon_i^{\mathrm{fs}}$ are the model's and reference's
split means. Uniform flow scores $1.000$, lower is better, and halving any term lowers
$S_\mathrm{flow}$ equally (\cref{app:metrics} defines $S_\mathrm{all}$ adding turbulence).
\emph{We recommend $S_\mathrm{flow}$ as the primary overall metric for comparing interior-task
models on \aerozoo{}, reported with its four constituent means.} Evaluations should also report the
share beating uniform flow ($\varepsilon_U<\varepsilon_U^\mathrm{fs}$) and per-enclosure means
(reference $\varepsilon_U$ spans
\azn{fs-full-epsU-opposed_hemispheres}--\azn{fs-full-epsU-dual_inlet}; \cref{fig:results}).

\begin{table*}[t]
  \centering
  \caption{Interior task on \key{full} (\azn{fs-full-n} canonical test samples, arithmetic means).
    The overall flow score $S_\mathrm{flow}$ (\cref{eq:composite}; uniform flow scores $1.000$) is
    the primary metric for comparing interior-task models; near-field errors
    $\varepsilon_U^{\mathcal N}$ and $\varepsilon_p^{\mathcal N}$ score samples with a body.
    \emph{GeoTransolver + BC} adds boundary-condition tokens; \emph{ISLA} predicts interior targets
    unless noted. Boundary-conditioned scaled rows average six seeds (four for \emph{ISLA +
    boundary targets}); SI-unit and boundary-blind rows are single-seed float32 runs (SDs and
    unweighted $r_U, r_p$ in \cref{tab:results-rows}). Bold marks the strictly best learned entry
    per column (ties plain).}
  \label{tab:results}
  \footnotesize
  \setlength{\tabcolsep}{5.2pt}
  \begin{tabular}{@{}lcr >{\columncolor{overallbg}}r rrrr r@{}}
    \toprule
    & & & \multicolumn{1}{>{\columncolor{overallbg}}c}{\textbf{Overall}\,$\bm{\downarrow}$}
      & \multicolumn{4}{c}{Mean ref.-scaled $\downarrow$} & Share $\uparrow$ \\
    \cmidrule(lr){4-4}\cmidrule(lr){5-8}\cmidrule(l){9-9}
    Model & BC inputs & Params & $\bm{S_\mathrm{flow}}$
      & $\varepsilon_U$ & $\varepsilon_U^{\mathcal N}$
      & $\varepsilon_p$ & $\varepsilon_p^{\mathcal N}$
      & $\varepsilon_U{<}\text{ref.}$ \\
    \midrule
    Free-stream reference & yes & -- & 1.000 & 0.440 & 0.442 & 1.809 & 2.014 & -- \\
Zero predictor & no & -- & 1.648 & 1.203 & 1.193 & 1.809 & 2.014 & -- \\
\midrule
\multicolumn{9}{@{}l}{\emph{SI units (no per-case scales; single seed, float32)}} \\
Transolver (no flow axis) & no & 9.0\,M & 1.648 & 1.202 & 1.193 & 1.810 & 2.015 & 1\% \\
GeoTransolver (no BC tokens) & no & 27.6\,M & 1.638 & 1.184 & 1.180 & 1.811 & 2.016 & 1\% \\
GeoTransolver + BC & yes & 27.6\,M & 1.040 & 0.475 & 0.479 & 1.809 & 2.014 & 20\% \\
\midrule
\multicolumn{9}{@{}l}{\emph{Per-case scales or unit-invariant}} \\
Transolver (no flow axis) & no & 9.0\,M & 1.042 & 0.617 & 0.621 & 1.419 & 1.539 & 41\% \\
Transolver + flow axis & flow axis & 9.0\,M & 0.679 & 0.306 & 0.309 & 1.212 & 1.315 & 61\% \\
GeoTransolver (no BC tokens) & no & 27.6\,M & 1.438 & 0.898 & 0.899 & 1.839 & 2.041 & 13\% \\
GeoTransolver + BC & yes & 27.6\,M & 0.290 & 0.152 & 0.179 & 0.427 & 0.434 & 86\% \\
\addlinespace[2pt]
ISLA & yes & 67.5\,M & 0.153 & 0.082 & 0.112 & 0.196 & 0.214 & 98\% \\
ISLA + boundary targets & yes & 67.5\,M & \textbf{0.123} & \textbf{0.080} & \textbf{0.110} & \textbf{0.124} & \textbf{0.149} & 98\% \\
    \bottomrule
  \end{tabular}
\end{table*}

\section{What current models do}
\label{sec:baselines}

On \key{full}, we evaluate Transolver~\citep{wu2024transolver},
GeoTransolver~\citep{adams2025geotransolver}, and ISLA~\citep{sharpe2026isla} across boundary
inputs and unit scalings (\cref{tab:results}). Without boundary conditions, models in per-case
scales memorize their training cases without generalizing (validation error
\azn{fit-scaled-plain-gap-lo}--\azn{fit-scaled-plain-gap-hi}$\times$ training), whereas models in
raw SI units fail to fit even the training set because inflow speed varies \azn{speed-span}-fold in
random directions (\cref{app:baselines}). \Cref{sec:trust} examines held-out
generalization.\reviewonly{ \rvPlannedRows}

\begin{figure*}[t]
  \centering
  \includegraphics[width=\textwidth]{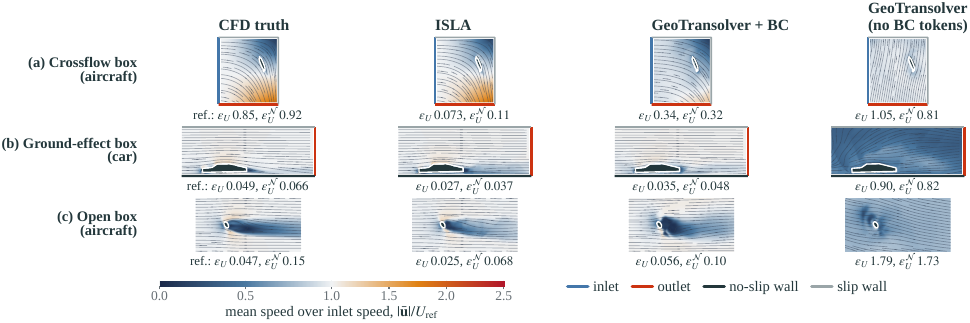}
  \caption{Predicted center-plane velocity and streamlines on three canonical \key{full} test
    samples (whole-domain $\varepsilon_U$, near-field $\varepsilon_U^{\mathcal N}$; ``ref.:''
    uniform flow; one seed per model, GeoTransolver + BC in per-case scales, float32). ISLA captures
    all three flows. GeoTransolver + BC captures the $90^\circ$ turn in (a) and the car wake and
    road boundary layer in (b); without BC tokens it predicts wrong flow directions. In (c),
    GeoTransolver + BC beats uniform flow near the aircraft (\azn{pred-open-tok-unear} vs.\
    \azn{pred-open-fs-unear}), whereas outer-volume drift (\azn{pred-open-far-pct}\% of volume)
    raises whole-domain $\varepsilon_U$ above uniform flow (\azn{pred-open-tok-u} vs.\
    \azn{pred-open-fs-u}; ISLA: \azn{pred-open-isla-u}).}
  \label{fig:predictions}
\end{figure*}

\paragraph{Setups.}
Every model trains on canonical \key{full} cases in PhysicsNeMo~\citep{physicsnemo} and is
evaluated at full mesh resolution on all \azn{fs-full-n} canonical test samples
(\cref{app:baselines}). We vary two axes: boundary inputs and unit scaling. Boundary-blind
Transolver and GeoTransolver receive only interior query points and signed distance to \key{no_slip}
walls; adding a global inlet-to-outlet flow axis (\emph{Transolver + flow axis}) provides flow
direction, whereas \emph{GeoTransolver + BC} (with boundary-condition tokens) and ISLA ingest
sampled boundary faces with their condition types and prescribed values. Raw \emph{SI units} pass
viscosity alone; \emph{per-case scales} nondimensionalize inputs and targets using boundary-derived
length, speed, and Reynolds number (ISLA computes these scales internally). Scaled
boundary-conditioned rows average \azn{gt-seeds} seeds (\azn{isla-joint-seeds} for \emph{ISLA +
boundary targets}); SI-unit and boundary-blind contrasts use single-seed float32 runs
(\cref{tab:results-rows,fig:predictions}).

\begin{enumerate}[leftmargin=*,itemsep=1pt,topsep=2pt]
  \item \emph{Boundary tokens drive the largest error drop.} In per-case scales, adding boundary
    tokens cuts GeoTransolver's mean $\varepsilon_U$ \azn{gt-notok-over-gtb}-fold, from
    \azn{gt-notok-epsU} to \azn{gt-full-epsU} ($S_\mathrm{flow}$ from \azn{gt-notok-S} to
    \azn{gt-full-S}), and cuts near-field error from \azn{gt-notok-epsU-near} to
    \azn{gt-full-epsU-near}. With tokens, GeoTransolver beats uniform flow on
    \azn{gt-full-below-fs}\% of samples and in every enclosure (\cref{fig:results}), gaining most
    in crossflow and dual-inlet boxes. Adding a global flow axis to Transolver halves its velocity
    error (from \azn{t-nofa-epsU} to \azn{tr-full-epsU}, $S_\mathrm{flow}=\azn{tr-full-S}$),
    leaving it well behind models given full boundary tokens.
  \item \emph{Condition types and prescribed values both matter.} In a controlled GeoTransolver
    ablation (\cref{tab:ablation}), zeroing condition types and prescribed values on the same
    boundary faces drops accuracy back to uniform flow ($S_\mathrm{flow}=\azn{abl-geom-S}$).
    Restoring condition types alone recovers most of the gain ($S_\mathrm{flow}=\azn{abl-type-S}$)
    yet still trails in dual-inlet boxes where prescribed speeds differ, whereas replacing tokens
    with a global mean inlet velocity loses outlet placement in crossflow boxes
    ($S_\mathrm{flow}=\azn{abl-notok-S}$).
  \item \emph{ISLA achieves the lowest errors across every metric and enclosure.} ISLA reaches
    $S_\mathrm{flow}=\azn{isla-full-S}$, $\varepsilon_U=\azn{isla-full-epsU}$, and
    $\varepsilon_p=\azn{isla-full-epsp}$, cutting $S_\mathrm{flow}$
    \azn{gt-over-isla-S-2dp}$\times$, $\varepsilon_U$ \azn{gt-over-isla-epsU-2dp}$\times$, and
    $\varepsilon_p$ \azn{gt-over-isla-epsp-2dp}$\times$ below GeoTransolver + BC (90\% Welch CIs
    in \cref{app:baselines}), beating uniform flow on \azn{isla-full-below-fs}\% of samples and in
    all eight enclosures (\cref{fig:results}). \emph{ISLA + boundary targets} cuts $\varepsilon_p$
    by \azn{isla-joint-p-drop-pct}\% to \azn{isla-joint-epsp} and lowers $S_\mathrm{flow}$ to
    \azn{isla-joint-S} overall and in every enclosure.
  \item \emph{Without boundary tokens, query-point density leaks wake direction in boxes.} Because
    the mesher refines a wedge from the body toward the outlet in boxes (unlike spheres and
    hemispheres), boundary-blind models in per-case scales achieve much lower errors in boxes than
    in spheres (e.g., Transolver scores \azn{t-nofa-open} in open boxes vs.\ \azn{t-nofa-sphere} in
    opposed spheres; \cref{app:baselines}).
  \needspace{3\baselineskip}
  \item \emph{Pressure separates models further, and SI-unit models learn none.} Zero gauge
    pressure is close to the truth in open boxes and spheres, whereas crossflow and dual-inlet
    boxes build large upstream pressure drops. No SI-unit model beats zero pressure
    (\azn{fs-full-epsp}). In per-case scales, GeoTransolver + BC lowers mean
    $\varepsilon_p$ to \azn{gt-full-epsp} (beating zero on \azn{gt-full-epsp-below-fs}\% of
    samples), ISLA reaches \azn{isla-full-epsp} (\azn{isla-full-epsp-below-fs}\%), and adding
    boundary targets lowers it to \azn{isla-joint-epsp}.
\end{enumerate}

\section{The error stack-up}
\label{sec:trust}

\begin{figure}[t]
  \centering
  \includegraphics[width=\columnwidth]{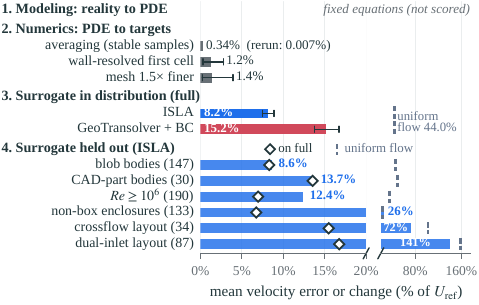}
  \caption{Four-link error stack-up (mean velocity error or change, \% of $\Uref$; x-scale
    compressed above 20\%). Solver sensitivities (Link 2; whiskers: case range) are slivers near
    zero beside \key{full} surrogate error (Link 3; 95\% bootstrap CIs; dashed: uniform flow,
    \azn{fs-full-epsU-pct}\%). On shared \key{full} test cases (Link 4; counts in parentheses),
    held-out ISLA (bars, \azn{isla-heldout-seeds} seeds) jumps past \key{full}-trained ISLA
    (diamonds) on unseen enclosures and layouts.}
  \label{fig:stackup}
\end{figure}

A surrogate's error with respect to a real flow builds up from four links, like the tolerances of
an assembly (\cref{fig:stackup}). First, the equations differ from reality (modeling error).
Second, the targets differ from the exact solution of those equations through discretization and a
finite averaging window. Third, the surrogate misses the targets on problems like its training
data. Finally, it adds out-of-distribution error on unfamiliar problems. \aerozoo{} states the
equations and measures the other three links, the last on its held-out splits.

The targets move little when averaged, rerun, or refined. Across stable samples, half-window means
differ by \azn{stackup-averaging-u-mean-pct}\% of $\Uref$, and cluster reruns shift velocity by
\azn{stackup-rerun-u-mean-pct}\%. In a \azn{gs-cases}-case grid study (\azn{gs-released} released
samples above $\Rey=10^5$ plus flat-plate and sphere controls; \cref{app:verification}),
$1.5\times$ mesh refinement shifts mean velocity by \azn{gs-urg-mean}\% (\azn{gs-unear-rg-mean}\%
in the near field) and a wall-resolved first cell shifts it by \azn{gs-urw-mean}\%. Because a
$1.5\times$ step underestimates total discretization error, Richardson-extrapolated force errors
are \azn{gs-extrap-factor-lo}--\azn{gs-extrap-factor-hi}$\times$ larger where convergence is
monotonic.

Surrogate error remains far larger. On \key{full}, ISLA's mean velocity error is
\azn{isla-full-epsU-pct}\% of $\Uref$ (near field \azn{isla-full-epsU-near-pct}\%; uniform flow
\azn{fs-full-epsU-pct}\%), roughly \azn{isla-mesh-ratio}$\times$ the mesh-refinement change
(\azn{isla-near-mesh-ratio}$\times$ in the near field, or
\azn{headroom-ratio-error-lo}--\azn{headroom-ratio-error-hi}$\times$ extrapolated discretization
error). GeoTransolver + BC has mean error \azn{gt-full-epsU-pct}\%
(\azn{gt-mesh-ratio}$\times$). Even in the easiest enclosures, opposed spheres
(\azn{isla-sphere-pct}\%) and open boxes (\azn{isla-open-pct}\%), ISLA stays roughly twice the
released-case mesh change. Surface forces are more sensitive to wall resolution, with median
$\varepsilon_{C_F}$ of \azn{gs-rg-median}\% under mesh refinement and \azn{gs-rw-median}\% with a
wall-resolved first cell (\cref{app:verification}).

On held-out splits, both GeoTransolver + BC and ISLA show the same sharp divide
(\cref{tab:heldout,fig:stackup}; ISLA's held-out suite uses its reference learning rate
$3.75\times10^{-4}$ across all splits so \key{full} and held-out runs are paired). Measured on the
exact test samples shared with \key{full}, holding out blobs or CAD parts leaves mean
$\varepsilon_U$ nearly unchanged for both ISLA ($\le\azn{isla-geom-gain-hi}\times$) and
GeoTransolver (\azn{gt-geom-gain-lo}--\azn{gt-geom-gain-hi}$\times$), and holding out
$\Rey\ge10^6$ raises it modestly (\azn{ho-re-isla-gain}$\times$ and \azn{ho-re-gt-gain}$\times$).
Holding out non-box enclosures, crossflow, or dual-inlet layouts instead multiplies mean
$\varepsilon_U$ by \azn{isla-newbc-gain-lo}--\azn{isla-newbc-gain-hi}$\times$ for ISLA
($S_\mathrm{flow}=\azn{isla-newbc-S-lo}$--\azn{isla-newbc-S-hi}) and
\azn{gt-newbc-gain-lo}--\azn{gt-newbc-gain-hi}$\times$ for GeoTransolver
($S_\mathrm{flow}=\azn{gt-newbc-S-lo}$--\azn{gt-newbc-S-hi}), leaving both worse than uniform flow
on unseen enclosures.

This stack-up sets \aerozoo{}'s priorities. A large CFD campaign is usually judged by its modeling
choices: the turbulence model, the inflow turbulence, and the choice between URANS and
scale-resolving methods such as LES or DDES. These choices change the targets and will limit a
surrogate's accuracy once its error approaches those differences. Today that point is far off: on
an unseen enclosure or boundary-condition layout, surrogate error comes mostly from failing to
generalize, the gap that \aerozoo{}'s held-out splits measure.

\section{Related work}
\label{sec:related}

\paragraph{Datasets and benchmarks.}
\Cref{tab:datasets} compares 3D aerodynamic datasets; 2D collections such as
AirfRANS~\citep{bonnet2022airfrans} and CFDBench~\citep{luo2023cfdbench} are summarized in
\cref{tab:datasets-2d}. Multi-PDE suites~\citep{takamoto2022pdebench,ohana2024well} vary physics
primarily on regular grids. PLAID standardizes simulation storage~\citep{casenave2025plaid}, and
FluidsBench, PhysicsBench, and CarBench build shared evaluations over existing
datasets~\citep{fluidsbench2026,lee2026physicsbench,elrefaie2026carbench}. \aerozoo{} adds a
declared tensor role for every PDE input and frozen splits holding out shapes, enclosures,
boundary-condition layouts, and Reynolds ranges.

\paragraph{Surrogates and what they assume.}
Surrogates for 3D aerodynamics~\citep{wu2024transolver,alkin2025abupt,ranade2025domino} mostly
condition on geometry and a global vector of operating parameters, with no input for
boundary-condition types. GNOT, BENO, and Unisolver ingest boundary shapes, boundary values, or
PDE components~\citep{hao2023gnot,wang2024beno,zhou2025unisolver}. GLOBE~\citep{sharpe2025globe}
argues that explicit boundary-condition inputs, unit invariance, and a global receptive field are
necessary for elliptic problems; \citet{li2026generalized} argue likewise for time-dependent
operators. Recent work also pretrains from geometry~\citep{wu2026geopt,medranonavarro2026gmp} or
pools steady CFD datasets, finding that transfer improves with pool
diversity~\citep{zhou2026crossdomain}.

\section{Conclusion}
\label{sec:conclusion}

On a single-family benchmark such as DrivAerML, Transolver's scaling floor is only
\azn{sat-within}\% below its achieved error and within a factor of \azn{sat-bracket-ratio} of the
targets' own averaging noise (\cref{app:saturation}). When \aerozoo{} poses \azn{complete} problems
varying shape, enclosure, and boundary conditions together, surrogates remain far from solver
accuracy: boundary conditioning unlocks a nearly \azn{gt-notok-over-gtb}-fold mean error drop for
GeoTransolver in distribution, yet even the best model, ISLA ($S_\mathrm{flow}=\azn{isla-full-S}$;
\azn{isla-joint-S} with boundary targets), remains roughly \azn{isla-mesh-ratio} times above
mesh-refinement sensitivity in distribution and degrades several-fold when enclosures or
boundary-condition layouts are held out. Incompressible, time-mean flow at industrial Reynolds
numbers is the minimum bar for an aerodynamic foundation model; compressible flow, scale-resolving
turbulence (LES), and unsteady flow come next. The same boundary-driven representation poses other
steady or time-mean boundary-value problems, from conjugate heat transfer to solid mechanics,
without dataset-specific conventions.

\newpage
\ifazpreprint\else
\section*{Impact Statement}

This paper presents a dataset and a benchmark for neural surrogates of fluid flow. Accurate fast
surrogates could shorten the design of vehicles, aircraft, buildings, and energy systems and cut
the energy that repeated CFD consumes. The released samples alone took \azn{core-hours-total-m}\,M
CPU core-hours of solver time, spent once so that others need not repeat it. Aerodynamic prediction
is dual-use; the release contains generic shapes rather than designs of any particular system, and
no personal data.
\fi

\bibliography{references}

\begin{thebibliography}{50}
\providecommand{\natexlab}[1]{#1}
\providecommand{\url}[1]{\texttt{#1}}
\expandafter\ifx\csname urlstyle\endcsname\relax
  \providecommand{\doi}[1]{doi: #1}\else
  \providecommand{\doi}{doi: \begingroup \urlstyle{rm}\Url}\fi

\bibitem[Adams et~al.(2026)Adams, Ranade, Nidhan, Nabian, Akhare, Cherukuri,
  and Choudhry]{adams2025geotransolver}
Adams, C., Ranade, R., Nidhan, S., Nabian, M.~A., Akhare, D., Cherukuri, R.,
  and Choudhry, S.
\newblock {GeoTransolver}: Learning physics on irregular domains using
  multi-scale geometry aware physics attention transformer.
\newblock In \emph{Advances in Neural Information Processing Systems}, 2026.
\newblock URL \url{https://arxiv.org/abs/2512.20399}.
\newblock To appear.

\bibitem[Alkin et~al.(2025)Alkin, Bleeker, Kurle, Kronlachner, Sonnleitner,
  Dorfer, and Brandstetter]{alkin2025abupt}
Alkin, B., Bleeker, M., Kurle, R., Kronlachner, T., Sonnleitner, R., Dorfer,
  M., and Brandstetter, J.
\newblock {AB-UPT}: Scaling neural {CFD} surrogates for high-fidelity
  automotive aerodynamics simulations via anchored-branched universal physics
  transformers.
\newblock \emph{Transactions on Machine Learning Research}, 2025.

\bibitem[Ashton et~al.(2024{\natexlab{a}})Ashton, Angel, Ghate, Kenway, Wong,
  Kiris, Walle, Maddix, and Page]{ashton2024windsorml}
Ashton, N., Angel, J.~B., Ghate, A.~S., Kenway, G. K.~W., Wong, M.~L., Kiris,
  C., Walle, A., Maddix, D.~C., and Page, G.
\newblock {WindsorML}: High-fidelity computational fluid dynamics dataset for
  automotive aerodynamics.
\newblock In \emph{Advances in Neural Information Processing Systems (NeurIPS)
  Datasets and Benchmarks Track}, volume~37, 2024{\natexlab{a}}.
\newblock \doi{10.52202/079017-1194}.

\bibitem[Ashton et~al.(2024{\natexlab{b}})Ashton, Maddix, Gundry, and
  Shabestari]{ashton2024ahmedml}
Ashton, N., Maddix, D.~C., Gundry, S., and Shabestari, P.~M.
\newblock {AhmedML}: High-fidelity computational fluid dynamics dataset for
  incompressible, low-speed bluff body aerodynamics.
\newblock arXiv:2407.20801, 2024{\natexlab{b}}.

\bibitem[Ashton et~al.(2024{\natexlab{c}})Ashton, Mockett, Fuchs, Fliessbach,
  Hetmann, Knacke, Schonwald, Skaperdas, Fotiadis, Walle, Hupertz, and
  Maddix]{ashton2024drivaerml}
Ashton, N., Mockett, C., Fuchs, M., Fliessbach, L., Hetmann, H., Knacke, T.,
  Schonwald, N., Skaperdas, V., Fotiadis, G., Walle, A., Hupertz, B., and
  Maddix, D.
\newblock {DrivAerML}: High-fidelity computational fluid dynamics dataset for
  road-car external aerodynamics.
\newblock arXiv:2408.11969, 2024{\natexlab{c}}.

\bibitem[Ashton et~al.(2026)Ashton, Clark, Goc, Heidt, Ivey, Agrawal, Bose,
  Adams, Sharpe, Leibovici, Akkurt, Nidhan, Ranade, and
  Kossaifi]{ashton2026hiliftaeroml}
Ashton, N., Clark, A., Goc, K., Heidt, L., Ivey, C., Agrawal, R., Bose, S.~T.,
  Adams, C., Sharpe, P., Leibovici, D., Akkurt, S., Nidhan, S., Ranade, R., and
  Kossaifi, J.
\newblock {HiLiftAeroML}: A high-fidelity computational fluid dynamics dataset
  for high-lift aircraft aerodynamics.
\newblock arXiv:2605.19565, 2026.

\bibitem[Ataei et~al.(2026)Ataei, Askari, Rahimi~Malekshan, and
  Jayaraman]{ataei2026zerotocad}
Ataei, M., Askari, F., Rahimi~Malekshan, K., and Jayaraman, P.~K.
\newblock {Zero-to-CAD}: Agentic synthesis of interpretable {CAD} programs at
  million-scale without real data.
\newblock arXiv:2604.24479, 2026.

\bibitem[{Autodesk AI Lab}(2026)]{adsk2026zerotocad100k}
{Autodesk AI Lab}.
\newblock Zero-to-{CAD} 100{K}: A curated subset of 100,000 geometrically
  diverse {CAD} construction sequences.
\newblock Hugging Face dataset \path{ADSKAILab/Zero-To-CAD-100k}, 2026.
\newblock URL \url{https://huggingface.co/datasets/ADSKAILab/Zero-To-CAD-100k}.
\newblock Apache License 2.0.

\bibitem[Bakarji et~al.(2022)Bakarji, Callaham, Brunton, and
  Kutz]{bakarji2022dimensionally}
Bakarji, J., Callaham, J., Brunton, S.~L., and Kutz, J.~N.
\newblock Dimensionally consistent learning with {Buckingham Pi}.
\newblock \emph{Nature Computational Science}, 2\penalty0 (12):\penalty0
  834--844, 2022.
\newblock \doi{10.1038/s43588-022-00355-5}.

\bibitem[Bonnet et~al.(2022)Bonnet, Mazari, Cinnella, and
  Gallinari]{bonnet2022airfrans}
Bonnet, F., Mazari, A.~J., Cinnella, P., and Gallinari, P.
\newblock {AirfRANS}: High fidelity computational fluid dynamics dataset for
  approximating {Reynolds}-averaged {Navier}--{Stokes} solutions.
\newblock In \emph{Advances in Neural Information Processing Systems (NeurIPS)
  Datasets and Benchmarks Track}, 2022.

\bibitem[Casenave et~al.(2025)Casenave, Roynard, Staber, Devaux-Rivi{\`e}re,
  Piat, Bucci, Akkari, Kabalan, Nguyen, Saverio, Carpintero~Perez, Kalaydjian,
  Fouch{\'e}, Gonon, Najjar, Daniel, Menier, Nastorg, Catalani, and
  Rey]{casenave2025plaid}
Casenave, F., Roynard, X., Staber, B., Devaux-Rivi{\`e}re, A., Piat, W., Bucci,
  M.~A., Akkari, N., Kabalan, A., Nguyen, X. M.~V., Saverio, L.,
  Carpintero~Perez, R., Kalaydjian, A., Fouch{\'e}, S., Gonon, T., Najjar, G.,
  Daniel, T., Menier, E., Nastorg, M., Catalani, G., and Rey, C.
\newblock {PLAID}: A unified data model for machine learning on heterogeneous
  physics simulations.
\newblock arXiv:2505.02974, 2025.
\newblock Presented at EuRIPS 2025; AI4Physics Workshop at ICML 2026.

\bibitem[Elrefaie et~al.(2024)Elrefaie, Morar, Dai, and
  Ahmed]{elrefaie2024drivaernetpp}
Elrefaie, M., Morar, F., Dai, A., and Ahmed, F.
\newblock {DrivAerNet++}: A large-scale multimodal car dataset with
  computational fluid dynamics simulations and deep learning benchmarks.
\newblock In \emph{Advances in Neural Information Processing Systems (NeurIPS)
  Datasets and Benchmarks Track}, 2024.

\bibitem[Elrefaie et~al.(2026)Elrefaie, Shu, Klenk, and
  Ahmed]{elrefaie2026carbench}
Elrefaie, M., Shu, D., Klenk, M., and Ahmed, F.
\newblock {CarBench}: A comprehensive benchmark for neural surrogates on
  high-fidelity {3D} car aerodynamics.
\newblock \emph{Communications Engineering}, 2026.
\newblock \doi{10.1038/s44172-026-00805-2}.

\bibitem[{FluidsBench}(2026)]{fluidsbench2026}
{FluidsBench}.
\newblock {FluidsBench}: Open benchmarks for scientific machine learning.
\newblock \url{https://fluidsbench.org/}, 2026.
\newblock Accessed 9 November 2026.

\bibitem[Friedman \& Dieng(2023)Friedman and Dieng]{friedman2023vendi}
Friedman, D. and Dieng, A.~B.
\newblock The {Vendi} score: A diversity evaluation metric for machine
  learning.
\newblock \emph{Transactions on Machine Learning Research}, July 2023.
\newblock URL \url{https://openreview.net/forum?id=g97OHbQyk1}.

\bibitem[Gebru et~al.(2021)Gebru, Morgenstern, Vecchione, Vaughan, Wallach,
  Daum{\'e}, and Crawford]{gebru2021datasheets}
Gebru, T., Morgenstern, J., Vecchione, B., Vaughan, J.~W., Wallach, H.,
  Daum{\'e}, H., and Crawford, K.
\newblock Datasheets for datasets.
\newblock \emph{Communications of the ACM}, 64\penalty0 (12):\penalty0 86--92,
  2021.
\newblock \doi{10.1145/3458723}.

\bibitem[Greenshields(2026)]{openfoam2026userguide}
Greenshields, C.~J.
\newblock \emph{{OpenFOAM} v14 User Guide}.
\newblock The OpenFOAM Foundation, London, UK, version 14 edition, July 2026.
\newblock URL \url{https://doc.cfd.direct/openfoam/user-guide-v14}.
\newblock Section 5.5: Mesh generation with the snappyHexMesh utility.

\bibitem[Hadamard(1902)]{hadamard1902problemes}
Hadamard, J.
\newblock Sur les probl{\`e}mes aux d{\'e}riv{\'e}es partielles et leur
  signification physique.
\newblock \emph{Princeton University Bulletin}, 13:\penalty0 49--52, 1902.

\bibitem[Hao et~al.(2023)Hao, Wang, Su, Ying, Dong, Liu, Cheng, Song, and
  Zhu]{hao2023gnot}
Hao, Z., Wang, Z., Su, H., Ying, C., Dong, Y., Liu, S., Cheng, Z., Song, J.,
  and Zhu, J.
\newblock {GNOT}: A general neural operator transformer for operator learning.
\newblock In \emph{Proceedings of the 40th International Conference on Machine
  Learning (ICML)}, volume 202 of \emph{Proceedings of Machine Learning
  Research}, pp.\  12556--12569, 2023.

\bibitem[{Inductiva}(2024)]{inductiva2024windtunnel}
{Inductiva}.
\newblock {WindTunnel-20k}.
\newblock Hugging Face dataset,
  \url{https://huggingface.co/datasets/inductiva/windtunnel-20k}, 2024.

\bibitem[Jordan et~al.(2024)Jordan, Jin, Boza, You, Cesista, Newhouse, and
  Bernstein]{jordan2024muon}
Jordan, K., Jin, Y., Boza, V., You, J., Cesista, F., Newhouse, L., and
  Bernstein, J.
\newblock Muon: An optimizer for hidden layers in neural networks.
\newblock \url{https://kellerjordan.github.io/posts/muon/}, 2024.

\bibitem[Kanchi et~al.(2025)Kanchi, Melanson, Somasekharan, Pan, and
  He]{kanchi2025unifoil}
Kanchi, R., Melanson, B., Somasekharan, N., Pan, S., and He, S.
\newblock {UniFoil}: A universal dataset of airfoils in transitional and
  turbulent regimes for subsonic and transonic flows.
\newblock In \emph{Advances in Neural Information Processing Systems (NeurIPS)
  Datasets and Benchmarks Track}, volume~38, 2025.
\newblock \doi{10.52202/085713-2467}.

\bibitem[Langtry \& Menter(2009)Langtry and Menter]{langtry2009correlation}
Langtry, R.~B. and Menter, F.~R.
\newblock Correlation-based transition modeling for unstructured parallelized
  computational fluid dynamics codes.
\newblock \emph{AIAA Journal}, 47\penalty0 (12):\penalty0 2894--2906, 2009.
\newblock \doi{10.2514/1.42362}.

\bibitem[Lee et~al.(2026)Lee, Jeong, and Kang]{lee2026physicsbench}
Lee, S.~W., Jeong, H., and Kang, N.
\newblock {PhysicsBench}: A unified leaderboard for generative and predictive
  models in engineering design and simulation.
\newblock arXiv:2608.24056, 2026.

\bibitem[Li et~al.(2026)Li, Sun, and Wang]{li2026generalized}
Li, R., Sun, Y., and Wang, W.
\newblock Generalized neural operator for parametric and boundary-value
  problems.
\newblock arXiv:2607.21932, 2026.

\bibitem[{Luminary Cloud}(2025)]{luminary2025shiftsuv}
{Luminary Cloud}.
\newblock {SHIFT-SUV}: High-fidelity computational fluid dynamics dataset for
  {SUV} external aerodynamics.
\newblock \url{https://huggingface.co/datasets/luminary-shift/SUV}, 2025.

\bibitem[Luo et~al.(2023)Luo, Chen, and Zhang]{luo2023cfdbench}
Luo, Y., Chen, Y., and Zhang, Z.
\newblock {CFDBench}: A large-scale benchmark for machine learning methods in
  fluid dynamics.
\newblock arXiv:2310.05963, 2023.

\bibitem[Medrano-Navarro et~al.(2026)Medrano-Navarro, Baldan, Liu, Holzschuh,
  Hagnberger, Niepert, and Thuerey]{medranonavarro2026gmp}
Medrano-Navarro, L., Baldan, G., Liu, Q., Holzschuh, B., Hagnberger, J.,
  Niepert, M., and Thuerey, N.
\newblock Geometry meets physics: Data-efficient pre-training for unstructured
  neural {PDE} solvers.
\newblock In \emph{Advances in Neural Information Processing Systems
  (NeurIPS)}, 2026.

\bibitem[Menter(1994)]{menter1994two}
Menter, F.~R.
\newblock Two-equation eddy-viscosity turbulence models for engineering
  applications.
\newblock \emph{AIAA Journal}, 32\penalty0 (8):\penalty0 1598--1605, 1994.
\newblock \doi{10.2514/3.12149}.

\bibitem[Menter et~al.(2003)Menter, Kuntz, and Langtry]{menter2003ten}
Menter, F.~R., Kuntz, M., and Langtry, R.
\newblock Ten years of industrial experience with the {SST} turbulence model.
\newblock In \emph{Turbulence, Heat and Mass Transfer 4}. Begell House, 2003.

\bibitem[Ohana et~al.(2024)Ohana, McCabe, Meyer, Morel, Agocs, Beneitez,
  Berger, Burkhart, Burns, Dalziel, Fielding, Fortunato, Goldberg, Hirashima,
  Jiang, Kerswell, Maddu, Miller, Mukhopadhyay, Nixon, Shen, Watteaux,
  R{\'e}galdo-Saint~Blancard, Rozet, Parker, Cranmer, and Ho]{ohana2024well}
Ohana, R., McCabe, M., Meyer, L., Morel, R., Agocs, F.~J., Beneitez, M.,
  Berger, M., Burkhart, B., Burns, K., Dalziel, S.~B., Fielding, D.~B.,
  Fortunato, D., Goldberg, J.~A., Hirashima, K., Jiang, Y.-F., Kerswell, R.~R.,
  Maddu, S., Miller, J., Mukhopadhyay, P., Nixon, S.~S., Shen, J., Watteaux,
  R., R{\'e}galdo-Saint~Blancard, B., Rozet, F., Parker, L.~H., Cranmer, M.,
  and Ho, S.
\newblock The well: a large-scale collection of diverse physics simulations for
  machine learning.
\newblock In \emph{Advances in Neural Information Processing Systems (NeurIPS)
  Datasets and Benchmarks Track}, 2024.

\bibitem[Paischer et~al.(2025)Paischer, Cotteleer, Dreze, Kurle, Rubini,
  Bleeker, Kronlachner, and Brandstetter]{paischer2025emmiwing}
Paischer, F., Cotteleer, L., Dreze, Y., Kurle, R., Rubini, D., Bleeker, M.,
  Kronlachner, T., and Brandstetter, J.
\newblock Going with the speed of sound: Pushing neural surrogates into
  highly-turbulent transonic regimes.
\newblock arXiv:2511.21474, 2025.
\newblock NeurIPS 2025 ML4PS Workshop.

\bibitem[{PhysicsNeMo Contributors}(2023)]{physicsnemo}
{PhysicsNeMo Contributors}.
\newblock {NVIDIA PhysicsNeMo}: An open-source framework for physics-based deep
  learning in science and engineering.
\newblock \url{https://github.com/NVIDIA/physicsnemo}, 2023.

\bibitem[Qiu et~al.(2025)Qiu, Kuang, Wang, Xu, Cui, Fu, Zhu, and
  Zhang]{qiu2025drivaerstar}
Qiu, J., Kuang, L., Wang, G., Xu, Y., Cui, L., Fu, S., Zhu, Y., and Zhang, R.
\newblock {DrivAerStar}: An industrial-grade {CFD} dataset for vehicle
  aerodynamic optimization.
\newblock In \emph{Advances in Neural Information Processing Systems (NeurIPS)
  Datasets and Benchmarks Track}, volume~38, 2025.
\newblock \doi{10.52202/085713-3293}.

\bibitem[Ranade et~al.(2025)Ranade, Nabian, Tangsali, Kamenev, Hennigh,
  Cherukuri, and Choudhry]{ranade2025domino}
Ranade, R., Nabian, M.~A., Tangsali, K., Kamenev, A., Hennigh, O., Cherukuri,
  R., and Choudhry, S.
\newblock {DoMINO}: A decomposable multi-scale iterative neural operator for
  modeling large scale engineering simulations.
\newblock arXiv:2501.13350, 2025.

\bibitem[Sharpe(2026{\natexlab{a}})]{sharpe2025globe}
Sharpe, P.
\newblock {GLOBE}: Accurate surrogates for boundary-driven {PDEs} via
  domain-inspired architectures and equivariance.
\newblock In \emph{Advances in Neural Information Processing Systems
  (NeurIPS)}, 2026{\natexlab{a}}.
\newblock URL \url{https://openreview.net/forum?id=YvUcvzejJe}.

\bibitem[Sharpe(2026{\natexlab{b}})]{sharpe2026isla}
Sharpe, P.
\newblock {ISLA}: A coordinate-free, measure-weighted transformer for
  boundary-driven {PDEs}.
\newblock arXiv:2611.17141, 2026{\natexlab{b}}.

\bibitem[Sung et~al.(2025)Sung, Spreizer, Elrefaie, Samuel, Jones, and
  Ahmed]{sung2025blendednet}
Sung, N., Spreizer, S., Elrefaie, M., Samuel, K., Jones, M.~C., and Ahmed, F.
\newblock {BlendedNet}: A blended wing body aircraft dataset and surrogate
  model for aerodynamic predictions.
\newblock In \emph{ASME International Design Engineering Technical Conferences
  (IDETC/CIE)}, 2025.
\newblock \doi{10.1115/DETC2025-168977}.

\bibitem[Takamoto et~al.(2022)Takamoto, Praditia, Leiteritz, MacKinlay,
  Alesiani, Pfl{\"u}ger, and Niepert]{takamoto2022pdebench}
Takamoto, M., Praditia, T., Leiteritz, R., MacKinlay, D., Alesiani, F.,
  Pfl{\"u}ger, D., and Niepert, M.
\newblock {PDEBench}: An extensive benchmark for scientific machine learning.
\newblock In \emph{Advances in Neural Information Processing Systems (NeurIPS)
  Datasets and Benchmarks Track}, 2022.

\bibitem[Tali et~al.(2025)Tali, Rabeh, Yang, Shadkhah, Karki, Upadhyaya,
  Dhakshinamoorthy, Saadati, Sarkar, Krishnamurthy, Hegde, Balu, and
  Ganapathysubramanian]{tali2024flowbench}
Tali, R., Rabeh, A., Yang, C.-H., Shadkhah, M., Karki, S., Upadhyaya, A.,
  Dhakshinamoorthy, S., Saadati, M., Sarkar, S., Krishnamurthy, A., Hegde, C.,
  Balu, A., and Ganapathysubramanian, B.
\newblock {FlowBench}: A large scale benchmark for flow simulation over complex
  geometries.
\newblock \emph{Journal of Data-centric Machine Learning Research}, 2, 2025.

\bibitem[Umetani \& Bickel(2018)Umetani and Bickel]{umetani2018learning}
Umetani, N. and Bickel, B.
\newblock Learning three-dimensional flow for interactive aerodynamic design.
\newblock \emph{ACM Transactions on Graphics}, 37\penalty0 (4):\penalty0 1--10,
  2018.
\newblock \doi{10.1145/3197517.3201325}.

\bibitem[Val{\'e}rio et~al.(2024)Val{\'e}rio, Penedones, and
  Sarmento]{valerio2024openfoam}
Val{\'e}rio, R., Penedones, H., and Sarmento, L.
\newblock Easily deploying {OpenFOAM} applications with {Python} and
  cloud-{HPC}.
\newblock In \emph{Proceedings of the 6th FOAM@Iberia Conference}, pp.\
  25--26, 2024.

\bibitem[Villar et~al.(2023)Villar, Yao, Hogg, Blum-Smith, and
  Dumitrascu]{villar2023dimensionless}
Villar, S., Yao, W., Hogg, D.~W., Blum-Smith, B., and Dumitrascu, B.
\newblock Dimensionless machine learning: Imposing exact units equivariance.
\newblock \emph{Journal of Machine Learning Research}, 24\penalty0
  (109):\penalty0 1--32, 2023.

\bibitem[Wang et~al.(2024)Wang, Li, Dwivedi, Hara, and Wu]{wang2024beno}
Wang, H., Li, J., Dwivedi, A., Hara, K., and Wu, T.
\newblock {BENO}: Boundary-embedded neural operators for elliptic {PDEs}.
\newblock In \emph{International Conference on Learning Representations
  (ICLR)}, 2024.

\bibitem[Weller et~al.(1998)Weller, Tabor, Jasak, and
  Fureby]{weller1998tensorial}
Weller, H.~G., Tabor, G., Jasak, H., and Fureby, C.
\newblock A tensorial approach to computational continuum mechanics using
  object-oriented techniques.
\newblock \emph{Computers in Physics}, 12\penalty0 (6):\penalty0 620--631,
  1998.
\newblock \doi{10.1063/1.168744}.

\bibitem[Wu et~al.(2024)Wu, Luo, Wang, Wang, and Long]{wu2024transolver}
Wu, H., Luo, H., Wang, H., Wang, J., and Long, M.
\newblock Transolver: A fast transformer solver for {PDEs} on general
  geometries.
\newblock In \emph{Proceedings of the 41st International Conference on Machine
  Learning (ICML)}, volume 235 of \emph{Proceedings of Machine Learning
  Research}, pp.\  53681--53705. PMLR, 2024.
\newblock URL \url{https://proceedings.mlr.press/v235/wu24r.html}.

\bibitem[Wu et~al.(2026)Wu, Guo, Li, Dou, Long, He, and Matusik]{wu2026geopt}
Wu, H., Guo, M., Li, Z., Dou, Z., Long, M., He, K., and Matusik, W.
\newblock {GeoPT}: Scaling physics simulation via lifted geometric
  pre-training.
\newblock arXiv:2602.20399, 2026.

\bibitem[Yang et~al.(2026)Yang, Tang, Liu, Thuerey, Zhang, and
  Chen]{yang2026superwing}
Yang, Y., Tang, W., Liu, M., Thuerey, N., Zhang, Y., and Chen, H.
\newblock {SuperWing}: A comprehensive transonic wing dataset for data-driven
  aerodynamic design.
\newblock \emph{Scientific Data}, 13:\penalty0 1113, 2026.
\newblock \doi{10.1038/s41597-026-07769-0}.

\bibitem[Zhou et~al.(2026)Zhou, Barati~Farimani, Ho, and
  Morel]{zhou2026crossdomain}
Zhou, A., Barati~Farimani, A., Ho, S., and Morel, R.
\newblock Cross-domain pretraining for steady-state neural {CFD} surrogates.
\newblock arXiv:2610.10398, 2026.

\bibitem[Zhou et~al.(2025)Zhou, Ma, Wu, Wang, and Long]{zhou2025unisolver}
Zhou, H., Ma, Y., Wu, H., Wang, H., and Long, M.
\newblock Unisolver: {PDE}-conditional transformers towards universal neural
  {PDE} solvers.
\newblock In \emph{Proceedings of the 42nd International Conference on Machine
  Learning}, volume 267 of \emph{Proceedings of Machine Learning Research},
  pp.\  79061--79088. PMLR, 2025.

\end{thebibliography}
\bibliographystyle{icml2026}

\newpage
\appendix
\crefalias{section}{appendix}
\crefalias{subsection}{appendix}
\onecolumn
\noindent\textbf{Guide to the appendices.}
\begin{center}\footnotesize
\begin{tabular}{@{}l>{\raggedright\arraybackslash}p{0.66\textwidth}l@{}}
  \toprule
  Appendix & Holds & Supports \\
  \midrule
  \ref{app:terms} & the CFD terms used, for readers in machine learning & the whole paper \\
  \ref{app:recipe} & how a case is posed, meshed, and solved; the stability label; what the recipe produced & \cref{sec:dataset} \\
  \ref{app:verification} & the measurements behind the error stack-up & \cref{sec:trust} \\
  \ref{app:schema} & the role of every key; the v1.3 conventions; a condensed key tree & \cref{sec:schema} \\
  \ref{app:metrics} & every metric, composite scores, reporting rules, and reference-implementation excerpt & \cref{sec:benchmark} \\
  \ref{app:splits} & all \azn{splits} splits and how they are drawn & \cref{sec:benchmark} \\
  \ref{app:baselines} & model inputs, unit scalings, multi-seed and diagnostic runs, held-out splits, ablation, and free-stream reference & \cref{sec:baselines,sec:trust} \\
  \ref{app:saturation} & scaling and the irreducible floor on DrivAerML: claims, bounds, fits, and protocol & \cref{sec:introduction,sec:conclusion} \\
  \ref{app:datasets} & sources for \cref{tab:datasets}; the shape-diversity score and its settings & \cref{sec:introduction,sec:dataset} \\
  \ref{app:datasheet} & the datasheet (v1.3 public release and license) & \\
  \bottomrule
\end{tabular}
\end{center}

\section{Terms}
\label{app:terms}

The paper is written for readers in machine learning. This appendix defines, in one or two lines
each, the terms from fluid dynamics and CFD that the main text uses, in the sense in which it uses
them.

\begin{multicols}{2}
\small
\setlength{\parskip}{0pt}
\paragraph{Flow and equations.}
\begin{description}[leftmargin=0pt,itemsep=1pt,topsep=1pt]
  \item[Navier--Stokes equations] The conservation of mass and momentum in a viscous fluid: the
    partial differential equations that every flow here obeys.
  \item[Incompressible flow] Flow of constant density: liquids, and gases well below the speed of
    sound. Its equations need one fluid property, the kinematic viscosity $\nu$ (m$^2$/s), the
    dynamic viscosity divided by the density.
  \item[Laminar and turbulent flow] Laminar flow is smooth and orderly. Turbulent flow is full of
    chaotic eddies over a wide range of sizes, which mix momentum far faster than viscosity alone.
  \item[Reynolds number] $\Rey=UL/\nu$ for a chosen speed $U$ and length $L$: the ratio of inertial
    to viscous effects. Geometrically similar problems at the same $\Rey$ have the same
    dimensionless solution (dynamic similarity).
  \item[RANS] Reynolds-averaged Navier--Stokes equations: equations for the mean flow, in which a
    turbulence model supplies the effect of the turbulent fluctuations.
  \item[URANS] Unsteady RANS: the same equations integrated in time, which resolves large, slow
    unsteadiness such as vortex shedding. An \aerozoo{} target is the time mean of a URANS solution
    over a finite window.
  \item[Hybrid RANS--LES] Methods that use RANS near walls and resolve the large eddies away from
    them.
  \item[DES and DDES] Detached-eddy simulation and its delayed form, the most common hybrid
    RANS--LES methods; the delayed form keeps attached boundary layers in RANS.
  \item[LES and wall-modeled LES] Large-eddy simulation resolves the large eddies everywhere and
    models the small ones; a wall-modeled LES also models the thin layer next to the wall.
  \item[DNS] Direct numerical simulation: every eddy is resolved and no turbulence is modeled.
  \item[Turbulence model] Here the $k$--$\omega$ SST model. It transports the turbulent kinetic
    energy $k$ (m$^2$/s$^2$) and the specific dissipation rate $\omega$ (1/s), and from them computes
    the eddy viscosity $\nu_t$ (m$^2$/s), the extra viscosity that represents turbulent mixing.
  \item[Inflow turbulence] The turbulence that the inlet prescribes, set by an intensity $I$, the
    size of the velocity fluctuations relative to the mean speed, and a length scale $\ell$, the
    size of the largest eddies.
  \item[Transition model] Here the Langtry--Menter model: two further equations that let a
    boundary layer start laminar and become turbulent. The intermittency $\gamma$ runs from 0
    (laminar) to 1 (turbulent); $\Rey_{\theta t}$ is a transported correlation variable that sets
    where transition begins.
  \item[Boundary layer] The thin layer next to a wall in which the velocity rises from zero to that
    of the outer flow, laminar or turbulent. Its momentum thickness measures the momentum it has
    lost. Where it leaves the wall, the flow \emph{separates}, and behind the body it forms a
    \emph{wake} of slowed, often unsteady flow.
  \item[Free stream] The flow far from any body; here, the state the inlet prescribes.
  \item[Pressures] The dynamic pressure is $q=\rho U^2/2$. Gauge pressure is measured from a
    reference, here zero at the outlet. Kinematic pressure, which OpenFOAM solves for, is pressure
    divided by density. With an eddy-viscosity model, the momentum equation carries only the
    deviatoric part of the turbulent stress, so the solved pressure includes its isotropic
    part, $(2/3)\rho k$; the stored pressure is the solver's.
  \item[Wall shear stress] $\vtau$ (Pa): the tangential shear force per unit area between fluid and
    wall, stored with OpenFOAM's sign as the traction of the wall on the fluid. The skin-friction
    coefficient is $\lVert\vtau\rVert/q$.
  \item[Force coefficient] A force divided by $q$ and an area. \aerozoo{} uses $\Lref^2$ as the
    area, so its coefficients are not the drag coefficients of the literature, which use a frontal
    or planform area.
  \item[Total-pressure loss coefficient] $K$: the drop in mass-flux-weighted total pressure,
    $p+\rho\lVert\vu\rVert^2/2$, from inlet to outlet, divided by $q$; the usual measure of a
    duct (\cref{app:metrics}).
\end{description}

\paragraph{Boundary conditions.}
\begin{description}[leftmargin=0pt,itemsep=1pt,topsep=1pt]
  \item[Dirichlet condition] The value of a field is prescribed on the boundary, such as the
    velocity on an inlet.
  \item[Zero gradient] The field's normal derivative is zero, so the boundary takes the value of
    the cell next to it (a homogeneous Neumann condition).
  \item[Inlet and outlet] Where flow enters, with prescribed velocity and turbulence, and where it
    leaves, at prescribed pressure.
  \item[No-slip and slip walls] A no-slip wall holds the fluid at rest. A slip wall allows no flow
    through it and exerts no shear, so it also models a plane of symmetry.
\end{description}

\paragraph{Mesh and numerics.}
\begin{description}[leftmargin=0pt,itemsep=1pt,topsep=1pt]
  \item[Finite-volume method] The domain is divided into cells; the solver balances mass and
    momentum over each cell and stores one value per cell, at its centroid.
  \item[Hex-dominant mesh] \texttt{snappyHexMesh} refines a background grid of hexahedra near the
    surfaces, snaps it onto them, and grows prism layers on walls. Cells cut by a surface become
    general polyhedra.
  \item[$y^+$ and the near-wall layers] $y^+$ is the distance from a wall to the first cell center
    in viscous units, $y\,u_\tau/\nu$, with friction velocity $u_\tau=(\lVert\vtau\rVert/\rho)^{1/2}$.
    By $y^+$, a turbulent boundary layer has three parts: the viscous sublayer below about 5, the
    buffer layer from about 5 to 30, and the logarithmic layer from about 30 to a few hundred. The
    stored \key{wall_y_plus} is the wall functions' own value, computed from the turbulent kinetic
    energy of the first cell.
  \item[Wall functions] Algebraic laws that supply the wall shear and the near-wall turbulence when
    the first cell lies in the logarithmic layer, so that the sublayer need not be resolved.
  \item[Convective time] $\Tc=\Lref/\Uref$, the time the free stream takes to pass the body.
  \item[Courant number] $U\,\Delta t/\Delta x$: how many cells the flow crosses in one time step.
  \item[Modeling and discretization error] Modeling error is the distance of the exact solution of
    the stated equations from the real flow. Discretization error is the distance of a numerical
    solution from that exact solution.
  \item[Grid-convergence study] The same case solved on successively finer meshes, to estimate the
    error of the discretization. When three meshes converge monotonically, Richardson extrapolation
    estimates the value on an infinitely fine mesh, and the observed order is the rate at which the
    error shrinks with cell size.
\end{description}

\paragraph{Enclosures.}
\begin{description}[leftmargin=0pt,itemsep=1pt,topsep=1pt]
  \item[Boxes] An open box has an inlet face opposite its outlet and slip side faces. A
    ground-effect box makes one side face a no-slip ground; a confined box makes all four no-slip;
    a crossflow box puts the outlet on a face beside the inlet; a dual-inlet box adds a second inlet
    beside the first (\cref{tab:configs}).
  \item[Sphere and hemisphere] An opposed sphere has inlet and outlet caps on opposite sides. A
    symmetry hemisphere is half a sphere closed by a symmetry (slip) plane.
  \item[Internal duct] A passage of one of 10 shapes with an inlet and an outlet end and no-slip
    walls, with a body in half of them.
\end{description}

\paragraph{Other.}
\begin{description}[leftmargin=0pt,itemsep=1pt,topsep=1pt]
  \item[Case and sample] A case is a planned flow problem. A sample is a case that was solved and
    released.
  \item[Well-posed] For a mathematical problem, after Hadamard: a solution exists, is unique, and
    depends continuously on the data. For a learning task (\cref{sec:schema}): the inputs determine
    the target.
  \item[Vendi score] The exponential of the entropy of the eigenvalues of a normalized similarity
    matrix~\citep{friedman2023vendi}: the effective number of distinct items in a collection.
\end{description}
\end{multicols}

\needspace{8\baselineskip}
\section{The generation recipe}
\label{app:recipe}

This appendix gives the parameters behind \cref{sec:dataset}: how a case is posed, meshed,
solved, and labeled, and what the recipe produced. Every number is a constant of the frozen recipe
that produced every released sample (mesh policy v14, solver policy v10, artifact schema 5).

\subsection{Posing a case}

From a 64-bit seed, the generator derives named, independent random streams for the geometry, the
pose, the enclosure, the body-wall layout, the boundary values and viscosity, the inflow
turbulence, and the mesh. The posed problem, including every stream key, is written to a manifest
before meshing. \Cref{tab:configs} lists the enclosures, \cref{fig:ducts} shows the duct shapes,
and \cref{tab:draws} lists what is drawn for each case.

\begin{table}[!ht]
  \centering
  \caption{The eight enclosure configurations. Box distances are from the body to each face, in
    reference lengths, drawn uniformly per case; the whole case is then rotated at random.}
  \label{tab:configs}
  \footnotesize
  \renewcommand{\arraystretch}{1.2}
  \begin{tabular}{@{}l>{\raggedright\arraybackslash}p{0.4\textwidth}>{\raggedright\arraybackslash}p{0.34\textwidth}@{}}
    \toprule
    Configuration (share) & What its boundaries do & Distances to the faces \\
    \midrule
    Open box (14.4\%) & inlet opposite outlet; the four side faces slip & inlet 0.8--2.2, outlet 2--5, sides 0.8--2.5 \\
    Ground-effect box (14.4\%) & as the open box, with one side face a no-slip ground & inlet 0.7--2, outlet 1.5--4, sides 0.5--2, ground 0.05--0.3 \\
    Confined box (14.4\%) & inlet opposite outlet; all four side faces no-slip & inlet 0.3--1, outlet 0.7--2, walls 0.1--0.7 \\
    Crossflow box (14.4\%) & outlet on a face adjacent to the inlet; one to three other faces no-slip & inlet and outlet 0.5--2.5, sides 0.4--2, no-slip 0.1--0.8 \\
    Dual-inlet box (14.4\%) & a second inlet, 0.5 to 1.5 times as fast as the first, on a face adjacent to the first; outlet opposite the first; zero to three faces no-slip & inlets 0.5--2, outlet 1--3, sides 0.4--2, no-slip 0.15--1 \\
    Opposed sphere (9\%) & a smooth or lumpy sphere with inlet and outlet caps on opposite sides & sampled radius \\
    Symmetry hemisphere (9\%) & half a sphere closed by a symmetry plane; inlet and outlet caps on the dome & sampled radius \\
    Internal duct (10\%) & a passage of one of 10 shapes with inlet and outlet ends and no-slip walls (sometimes slip span walls); a body in half of them & shape-dependent \\
    \bottomrule
  \end{tabular}
\end{table}

\begin{table}[!ht]
  \centering
  \caption{What is drawn for each case. ``Log-uniform'' is uniform in the logarithm.}
  \label{tab:draws}
  \footnotesize
  \begin{tabular}{@{}l>{\raggedright\arraybackslash}p{0.72\textwidth}@{}}
    \toprule
    Quantity & How it is set \\
    \midrule
    Enclosure & one of eight configurations, with the shares of \cref{tab:configs} \\
    Reference length $\Lref$ & log-uniform, 0.02 to 20\,m. The body is scaled so that its largest
      extent before rotation is $\Lref$; in a body-free duct, $\Lref$ is the inlet's hydraulic diameter \\
    Inlet speed $\Uref$ & log-uniform, 0.25 to 50\,m/s, on the primary inlet (stays on the primary inlet even when a dual-inlet box's second inlet is faster) \\
    Reynolds number $\Rey$ & one of the five decades from $10^3$ to $10^8$, in equal shares;
      log-uniform within it \\
    Viscosity $\nu$ & $\Uref\Lref/\Rey$ \\
    Density $\rho$ & 1.225\,kg/m$^3$ in every case \\
    Inflow turbulence & intensity $I$ log-uniform, 0.5\% to 10\%; length scale $\ell$
      log-uniform, 0.01 to $1\,\Lref$, with $k=(3/2)(I U_\mathrm{in})^2$ and
      $\omega=k^{1/2}/(C_\mu^{1/4}\ell)$ ($C_\mu=0.09$), using each inlet's own speed $U_\mathrm{in}$ \\
    Body orientation & in the open, ground-effect, and confined boxes, aligned with the flow with
      probability 0.5 (longest area-weighted principal axis along the inlet direction, with Gaussian
      incidence and sideslip of about $6^\circ$; the sign of its third moment of area picks fore and
      aft when skewness $\ge 0.1$, or at random otherwise, the roll about it is uniform in open and
      confined boxes, and in ground-effect boxes the vertical sign is arbitrary, so a wide-span
      aircraft is posed spanwise and about half of those bodies are inverted); otherwise, and in
      every other configuration, uniformly random \\
    Mirroring & with probability 0.5, with the triangle winding restored \\
    Body wall & all no-slip, two bands, or three bands, each with probability $1/3$. Bands are
      contiguous, alternately slip and no-slip, and cut by planes normal to a random direction \\
    Second body & with probability 0.03: a copy of the case's own body, 0.25 to 1 times its size and
      posed independently, 0.1 to 1 reference lengths away \\
    Whole case & rotated at random in 3D \\
    \bottomrule
  \end{tabular}
\end{table}

\needspace{5\baselineskip}
The inlet prescribes a uniform velocity (\cref{tab:operators}). From $I$, $\ell$, and the inlet
speed, the generator sets the inlet's turbulent kinetic energy, specific dissipation rate, and
transition-onset Reynolds number (by the Langtry--Menter correlation), with intermittency
$\gamma=1$. The resulting eddy viscosity at the inlet is
$\nu_t=\azn{nut-inlet-coefficient}\,I\,\ell\,\Uref$, so
$\nu_t/\nu=\azn{nut-inlet-coefficient}\,I\,(\ell/\Lref)\,\Rey$. The solver caps this ratio at
$10^5$ (\cref{fig:inflow}).

\begin{table}[!ht]
  \centering
  \caption{What each boundary type imposes.}
  \label{tab:operators}
  \footnotesize
  \begin{tabular}{@{}llll@{}}
    \toprule
    Boundary & Velocity & Pressure & Turbulence fields \\
    \midrule
    \key{inlet} & prescribed (Dirichlet) & zero gradient & prescribed \\
    \key{outlet} & zero gradient; prescribed where flow re-enters & zero (Dirichlet) & as velocity \\
    \key{no_slip} & zero & zero gradient & wall functions ($k$, $\omega$, $\nu_t$); zero gradient ($\gamma$, $\Rey_{\theta t}$) \\
    \key{slip} & no penetration, free tangential slip & zero gradient & zero gradient \\
    \bottomrule
  \end{tabular}
\end{table}

\subsection{Meshing}

Every input surface passed to the mesher must be closed, consistently oriented, manifold, and free
of self-intersection.
\texttt{snappyHexMesh} builds each mesh from a uniform, near-isotropic background grid sized to a
budget of two to three million cells (about \azn{cells-median-m} million cells at the median after
refinement). It refines the mesh by two levels at the body's surface and by one
level in a wedge that follows the wake to the outlet. The wedge starts $0.25\,\Lref$ upstream of
the body and widens by half the body's transverse extent. Wall layers cover the no-slip walls and
the whole body, slip bands included, and are sized before the flow is known:
\begin{itemize}
  \item the first cell's center is placed at $y^+=30$ for a friction velocity that is
    \azn{yplus-safety-factor} times the larger of the Blasius and Prandtl--Schlichting flat-plate
    estimates at the case's Reynolds number. Because the friction velocity is doubled, the target
    on a flat plate is $y^+=\azn{yplus-plate-target}$;
  \item the first layer is at most $0.005\,\Lref$ thick and at least $1/800$ of the surface cell;
  \item each further layer is 1.5 times thicker, until a layer reaches half a surface cell, which
    gives 3 to 16 layers.
\end{itemize}
Above $\Rey=10^5$, the median over the samples with a body of each body's median $y^+$ is
\azn{yplus-median} (quartiles \azn{yplus-q1} to \azn{yplus-q3}). In \azn{yplus-5-30-pct}\% of those samples it lies between 5 and 30, in the
buffer layer (\cref{app:terms}).

A mesh is admitted only if one connected fluid region holds at least 99.9\% of the cells, every
boundary region that the manifest poses is present, and the mesh is within fixed quality limits. The
mesher works to limits of $80^\circ$ on non-orthogonality and concavity. Admission requires
concavity below $90^\circ$, skewness below 20 on boundaries and 10 inside, and OpenFOAM
\texttt{checkMesh}'s normalized cell determinant (\texttt{minDeterminant}) above $10^{-6}$.

\subsection{Solving and averaging}

\Cref{tab:numerics} gives the numerical policy, which is the same for every case.

\begin{table}[!ht]
  \centering
  \caption{The fixed numerical policy (OpenFOAM~14).}
  \label{tab:numerics}
  \footnotesize
  \begin{tabular}{@{}ll@{}}
    \toprule
    Component & Fixed choice \\
    \midrule
    Time integration & second-order backward, adaptive step \\
    Courant limit & 10; initial step $0.002\,\Tc$, at most $0.005\,\Tc$ \\
    Pressure--velocity coupling & PIMPLE: one outer, two pressure, one non-orthogonal corrector \\
    Momentum convection & bounded \texttt{limitedLinearV 1} (limited second order) \\
    Turbulence convection & bounded upwind (first order) \\
    Gradients, interpolation & linear; Laplacian \texttt{limited corrected 0.33} \\
    Closure & \texttt{kOmegaSSTLM}, wall functions on no-slip walls \\
    Schedule & warm-up $0$--$15\,\Tc$, mean over $15$--$95\,\Tc$ \\
    \bottomrule
  \end{tabular}
\end{table}

\subsection{The stability label}
\label{app:stability}

The stability label compares the means $\bar{(\cdot)}_1$ and $\bar{(\cdot)}_2$ over the two halves
of the averaging window (\cref{tab:numerics}). A sample is \key{mean_stable} if
\begin{equation*}
  \frac{\lVert\bar{\vu}_2-\bar{\vu}_1\rVert_V}{\Uref}\le0.02,\qquad
  \frac{\lVert\bar p_2-\bar p_1\rVert_V}{\qref}\le0.05,\qquad
  \frac{\lvert\bar C_{F,2}-\bar C_{F,1}\rvert}{\max(\lvert\bar C_F\rvert,\,0.05)}\le0.02,
\end{equation*}
with $\lVert\cdot\rVert_V$ the volume-weighted root mean square, $\bar C_F$ the full-window mean
force coefficient, and every field finite; otherwise it is \key{mean_unstable}. In a body-free duct,
stability is checked on the force over the duct's no-slip walls rather than on the scored duct loss
$K$. If a duct's first window is \key{mean_unstable}, the solver continues the same
flow and averages a second window of $80\,\Tc$. That window starts at $95\,\Tc$ or after two
turnover times (the duct's volume over its inlet's volume flow), whichever is later. Only ducts
get this second window. Of the unstable samples, \azn{unstable-fail-velocity-pct}\% fail the
velocity test, \azn{unstable-fail-force-pct}\% the force test, and \azn{unstable-fail-pressure-pct}\%
the pressure test.

\subsection{What the recipe produced}
\label{app:statistics}

\begin{figure}[!ht]
  \centering
  \begin{minipage}[t]{0.485\textwidth}
    \vspace{0pt}\centering
    \includegraphics{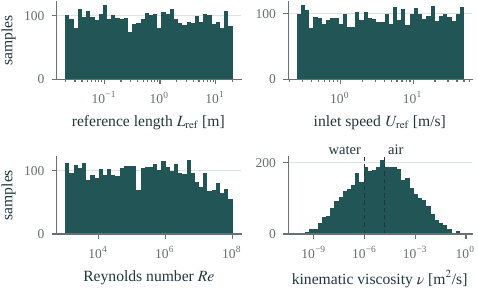}
    \caption{The three drawn scales and the derived viscosity over the released samples
      (logarithmic bins). Within each Reynolds decade the draws are log-uniform; the dip in the top
      decade comes from the timed-out cases. Dashed lines mark the viscosities of water and air at
      room temperature; \azn{nu-air-water-pct}\% of samples lie between them.}
    \label{fig:scales}
  \end{minipage}\hfill
  \begin{minipage}[t]{0.485\textwidth}
    \vspace{0pt}\centering
    \includegraphics{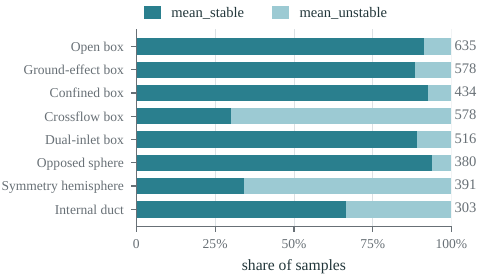}
    \caption{Crossflow boxes and symmetry hemispheres are mostly \key{mean_unstable};
      five configurations are about 90\% stable. Share of released samples labeled
      \key{mean_stable}, by enclosure configuration, with the number of samples at the right. The
      label records whether the finite-window mean had settled; samples with either label passed
      the solver's checks.}
    \label{fig:stability}
  \end{minipage}
\end{figure}

\begin{figure}[!ht]
  \centering
  \includegraphics[width=\textwidth]{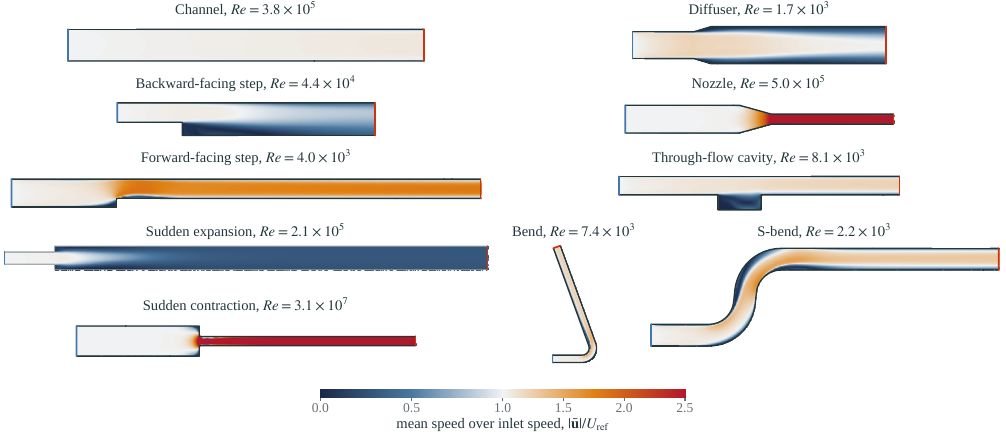}
  \caption{One body-free internal-duct sample of each of the \azn{duct-shapes} duct shapes,
    cut along the passage. Each panel is cropped to its own passage and drawn to its own scale; its
    title gives the shape and the Reynolds number, built on the inlet's hydraulic diameter. All
    \azn{duct-shapes} samples are \key{mean_stable}.}
  \label{fig:ducts}
\end{figure}

\begin{figure}[!ht]
  \centering
  \begin{minipage}[t]{0.485\textwidth}
    \vspace{0pt}\centering
    \includegraphics[width=\linewidth]{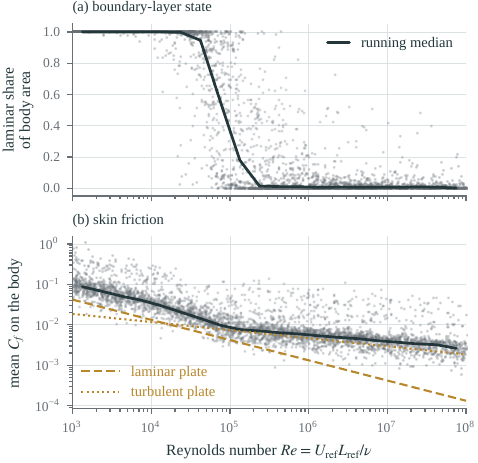}
    \caption{The flows span regimes: one dot per released sample with a body.
      (a) Share of the body's no-slip area where the transition model's
      intermittency $\bar\gamma$ is below one half. (b) Area-mean skin-friction coefficient
      $C_f=\lVert\vtau\rVert/\qref$ on the body's no-slip faces, with the
      mean friction of a laminar and a turbulent flat plate of length $\Lref$.}
    \label{fig:regimes}
  \end{minipage}
\end{figure}

\begin{figure}[!ht]
  \centering
  \includegraphics[width=0.8\textwidth]{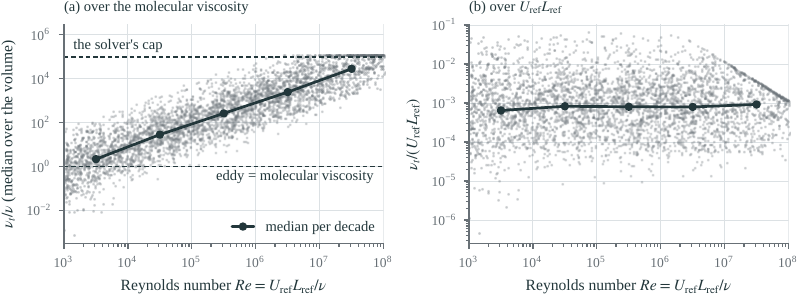}
  \caption{The eddy viscosity of each released sample against its Reynolds number: the
    median of $\nu_t$ over the sample's volume, (a) over the molecular viscosity $\nu$ and (b) over
    $\Uref\Lref$, with the median of each Reynolds decade. The dashed lines in (a) mark $\nu_t=\nu$
    and the solver's cap of $10^5$, which is the sloping edge at the right of (b). The inlet's eddy
    viscosity is about $\azn{nut-inlet-coefficient}\,I\,\ell\,\Uref$ whatever $\nu$ is, so (b) is
    level and (a) rises in proportion to the Reynolds number.}
  \label{fig:inflow}
\end{figure}

\Cref{tab:composition} and \cref{fig:scales,fig:stability,fig:ducts} break down the
released samples by geometry family and enclosure configuration, show the distributions of the drawn
scales and of the stability label, and display one solved duct of each shape; \cref{fig:vendi}
shows example bodies from each geometry family. \Cref{fig:regimes} shows the flow regimes of \cref{sec:dataset}, and
\cref{fig:inflow} the eddy viscosity that the inflow turbulence gives each sample.

\paragraph{Why the viscosity varies.}
Each case draws its own viscosity, for two reasons. First, a real fluid's viscosity is not one
number: by Sutherland's law at sea-level pressure, the kinematic viscosity of air rises
by \azn{air-nu-change-15-25-pct}\% from 15 to 25\,\textdegree C and by
\azn{air-nu-change-0-35-pct}\% from 0 to 35\,\textdegree C, and that of water is an order of
magnitude lower. Second, an incompressible flow depends on its length, speed, and viscosity only
through the Reynolds number $UL/\nu$, the similarity of \cref{sec:schema}. Two samples with the
same geometry in units of $L$, the same boundary conditions in units of $U$, and the same Reynolds
number have the same flow in those units, whatever their SI values. Drawing the three scales
independently spreads the samples of each Reynolds number over many combinations of SI values
(\cref{tab:draws,fig:scales}), so a model is rewarded for learning the similarity, a basic
acceptance test of a flow solver, and cannot reach a low error by memorizing the viscosity of one
fluid.

\paragraph{The inflow's eddy viscosity.}
The inlet's eddy viscosity does not depend on $\nu$, and in the typical sample it persists
through the volume. Over the samples, the median ratio of the volume median of $\nu_t$ to the
inlet's value is \azn{nut-volume-over-inlet-median}.
\begin{itemize}[itemsep=1pt]
  \item \emph{In units of $\Uref\Lref$,} the median over the release of the volume median is
    \azn{nut-over-ul-median} (5th to 95th percentile \azn{nut-over-ul-p05} to \azn{nut-over-ul-p95}), and
    the inlet's value a median of \azn{nut-inlet-over-ul-median}.
  \item \emph{Over $\nu$,} the volume median therefore rises in proportion to the Reynolds number.
    Its median is \azn{nut-median-low} in the lowest Reynolds decade and \azn{nut-median-high} in
    the highest. It exceeds 100 in \azn{nut-above-100-pct}\% of samples, and in
    \azn{nut-top-decade-above-90k-pct}\% of the top decade's samples it is within a tenth of the
    solver's cap.
  \item \emph{At the walls,} friction still depends on $\nu$: the body's mean skin friction falls
    \azn{cf-median-fall}-fold from the lowest to the highest Reynolds decade. Its median is
    \azn{cf-over-laminar-lo} to \azn{cf-over-laminar-hi} times the laminar flat-plate law in the
    two lowest decades and \azn{cf-over-turbulent-lo} to \azn{cf-over-turbulent-hi} times the
    turbulent law in the three highest (\cref{fig:regimes}b).
\end{itemize}

\begin{table}[!ht]
  \centering
  \caption{Released samples by geometry family and enclosure configuration. The planned counts in each
    row follow the configuration weights; the shortfalls are failed and timed-out cases. By count the
    shortfall is largest in the confined box, which has the most cells
    (\azn{shortfall-confined} of \azn{shortfall-confined-planned} planned cases,
    \azn{shortfall-confined-pct}\%, \azn{shortfall-confined-timed-out} of them timed out), and by
    share in the internal duct (\azn{shortfall-duct} of \azn{shortfall-duct-planned},
    \azn{shortfall-duct-pct}\%, \azn{shortfall-duct-failed} of them failed). The internal-duct column
    includes the \azn{body-free-ducts} body-free ducts, each counted under the family of the body
    drawn for it.}
  \label{tab:composition}
  \footnotesize
  \setlength{\tabcolsep}{4pt}
  \begin{tabular}{@{}lrrrrrrrrr@{}}
    \toprule
    Geometry family & \shortstack[r]{Open\\box} & \shortstack[r]{Ground-effect\\box} & \shortstack[r]{Confined\\box} & \shortstack[r]{Crossflow\\box} & \shortstack[r]{Dual-inlet\\box} & \shortstack[r]{Opposed\\sphere} & \shortstack[r]{Symmetry\\hemisphere} & \shortstack[r]{Internal\\duct} & Total \\
    \midrule
    Blobs & 229 & 203 & 133 & 206 & 184 & 143 & 153 & 116 & 1{,}367 \\
Blender bluff & 78 & 70 & 55 & 67 & 51 & 49 & 38 & 31 & 439 \\
Procedural cars & 79 & 57 & 47 & 62 & 58 & 34 & 43 & 36 & 416 \\
Zero-to-CAD & 58 & 64 & 32 & 53 & 39 & 39 & 23 & 25 & 333 \\
Airframe Foundry & 56 & 56 & 55 & 51 & 46 & 40 & 41 & 22 & 367 \\
Streamlined & 65 & 56 & 52 & 56 & 67 & 47 & 45 & 46 & 434 \\
Blender streamlined & 70 & 72 & 60 & 83 & 71 & 28 & 48 & 27 & 459 \\
\midrule
Total & 635 & 578 & 434 & 578 & 516 & 380 & 391 & 303 & 3{,}815 \\
    \bottomrule
  \end{tabular}
\end{table}

\FloatBarrier

\section{Solution verification}
\label{app:verification}

This appendix gives the measurements behind the second link of the error stack-up
(\cref{sec:trust,fig:stackup}): how close the released solutions are to the exact solutions of their
equations. Its main part is a grid study of reruns, finer meshes, and wall-resolved first cells.
Under $1.5\times$ mesh refinement, the velocity changes by a mean of \azn{gs-urg-mean}\% of $\Uref$
(median \azn{gs-urg-median}\%; near-field mean \azn{gs-unear-rg-mean}\%), and the force changes by a
median of \azn{gs-rg-median-unfloored}\% unfloored (\azn{gs-rg-median}\% in the benchmark's metric,
whose denominator is at least 0.05).

\paragraph{Design.}
A grid study re-solved \azn{gs-cases} cases with the frozen recipe: one released \key{mean_stable}
sample from each of \azn{gs-released} strata (four open boxes: Streamlined, Blender streamlined,
Airframe Foundry, and a bluff body; a car in ground effect; a duct with a body; and a body in an
opposed sphere), plus two canonical controls (a flat plate at $\Rey=10^6$ with 1\% inflow
turbulence and a sphere at $\Rey=10^5$ with 0.5\%). Released cases were drawn by fixed hash before
seeing any result from the \azn{gs-pool} \key{mean_stable} samples (\azn{gs-pool-pct}\% of the
release, all above $\Rey=10^5$; drawn cases span $\azn{gs-re-min}$--$\azn{gs-re-max}$) whose
production solve took at most \azn{gs-max-core-hours} core-hours and whose first wall layer met the
$y^+$ target rather than the $0.005\,\Lref$ cap. That selection pool excludes crossflow,
dual-inlet, confined, and symmetry-hemisphere cases. Each case was solved four more times, changing
only the mesh:
\begin{description}[leftmargin=2em,itemsep=0pt]
  \item[R] the production mesh, solved again on the study's cluster;
  \item[C, G] meshes 1.5 times coarser and 1.5 times finer in each direction, with the production
    wall rule;
  \item[W] the production mesh with its first cell at $y^+\approx1$.
\end{description}
The posing, solver policy, and averaging window match production. The finer of two solutions is
taken as truth and the coarser as prediction; interior errors (\cref{eq:interior}) evaluate on the
truth's cell centroids after reconstructing the coarser field by nearest cell plus a limited
least-squares gradient over 12 neighbors (self-interpolation floor \azn{gs-self-interp-min} to
\azn{gs-self-interp-max}\% of $\Uref$ on the production meshes). On the Airframe Foundry case, finer
mesh G hit the cluster's two-day wall limit at 41\% of its averaging window and is marked ``--'' in
\cref{tab:gridstudy}. The study used \azn{gs-core-hours-k}\,k CPU core-hours.

\begin{table}[!ht]
  \centering
  \caption{Grid study by case: Reynolds number; cells of the production (R), finer (G), and
    wall-resolved (W) meshes; change from production to finer in force and velocity; where C, R,
    and G converge monotonically, observed order of $C_F$ and production force error relative to
    Richardson extrapolation; change from production to wall-resolved in force; and laminar share
    of the body's boundary layer on both meshes. Changes, errors, and shares in percent; ``--'':
    not defined or (Airframe Foundry G) did not finish within the two-day wall limit. A released
    body with no enclosure named is in an open box.}
  \label{tab:gridstudy}
  \footnotesize
  \setlength{\tabcolsep}{3pt}
  \begin{tabular}{@{}lrrrrrrrrr@{}}
    \toprule
    & & & \multicolumn{4}{c}{production $\to$ finer}
      & \shortstack{production $\to$\\wall-resolved} & \multicolumn{2}{c}{laminar share} \\
    \cmidrule(lr){4-7}\cmidrule(lr){8-8}\cmidrule(l){9-10}
    Case & $\Rey$ & cells R/G/W (M) & $\varepsilon_{C_F}$ & $\varepsilon_U$ & order & extrapolated
      & $\varepsilon_{C_F}$ & production & \shortstack{wall-\\resolved} \\
    \midrule
    streamlined body & $4.1\times10^{5}$ & 3.2/10.8/3.4 & 0.7 & 0.2 & 0.9 & 2.3 & 0.5 & 0 & 0 \\
car in ground effect & $7.2\times10^{5}$ & 2.8/8.7/3.3 & 0.5 & 1.6 & -- & -- & 2.7 & 0 & 0 \\
Airframe Foundry body & $1.9\times10^{5}$ & 5.9/--/6.0 & -- & -- & -- & -- & 3.5 & 0 & 7 \\
Blender bluff body & $1.4\times10^{6}$ & 4.4/14.8/4.5 & 5.3 & 1.6 & 2.2 & 9.2 & 0.6 & 0 & 0 \\
body in an opposed sphere & $2.6\times10^{5}$ & 3.8/9.6/3.8 & 2.0 & 0.7 & -- & -- & 8.6 & 0 & 0 \\
duct with a body & $2.6\times10^{5}$ & 1.3/3.0/1.8 & 6.8 & 4.0 & 1.0 & 22.9 & 12.0 & 0 & 0 \\
Blender streamlined body & $2.6\times10^{5}$ & 3.0/10.0/3.1 & 8.0 & 1.1 & 1.9 & 14.7 & 36.1 & 30 & 11 \\
plate (control) & $1.0\times10^{6}$ & 3.0/9.6/3.2 & 0.9 & 0.3 & 2.7 & 1.4 & 6.9 & 0 & 0 \\
sphere (control) & $1.0\times10^{5}$ & 3.7/12.4/3.9 & 19.1 & 1.3 & -- & -- & 20.5 & 72 & 67 \\
    \bottomrule
  \end{tabular}
\end{table}

\needspace{6\baselineskip}
\paragraph{Findings.}
\Cref{tab:gridstudy} gives the results by case.
\begin{itemize}
  \item \emph{Reruns.} On another cluster, with another rank count and decomposition, the force
    moves by at most \azn{gs-rp-max}\% in the steady cases (mean velocity change
    \azn{stackup-rerun-u-mean-pct}\% of $\Uref$).
  \item \emph{Mesh.} Refining the mesh $1.5\times$ per direction (realized cell-count ratio
    \azn{gs-ratio-realized-lo}--\azn{gs-ratio-realized-hi}) shifts velocity by a mean of
    \azn{gs-urg-mean}\% of $\Uref$ (range \azn{gs-urg-min}--\azn{gs-urg-max}\%, median
    \azn{gs-urg-median}\%; \azn{gs-urg-mean-released}\% over the \azn{gs-rg-released-cases}
    released cases alone) and near-field velocity by a mean of \azn{gs-unear-rg-mean}\% (up to
    \azn{gs-unear-rg-max}\%). Force shifts by a median of \azn{gs-rg-median}\% in the benchmark's
    metric over \azn{gs-rg-cases} cases (\azn{gs-rg-median-unfloored}\% unfloored;
    \azn{gs-small-force-cases} of the \azn{gs-cases} cases have $\lvert C_F\rvert<0.05$, where the
    denominator floor applies), almost all from pressure (friction moves by at most
    \azn{gs-friction-rg-max}\%). Where C, R, and G converge monotonically (\azn{gs-extrap-cases}
    cases), production force lies \azn{gs-extrap-min}--\azn{gs-extrap-max}\% from the
    Richardson-extrapolated value (median \azn{gs-extrap-median}\%, or
    \azn{gs-extrap-factor-lo}--\azn{gs-extrap-factor-hi}$\times$ the refinement change). The
    sphere's \azn{gs-rg-max}\% jump reflects a regime switch: its production run is
    \key{mean_unstable}, whereas its finer mesh settles to steady flow.
  \item \emph{Wall cell.} Setting the first cell to $y^+\approx1$ shifts velocity by a mean of
    \azn{gs-urw-mean}\% of $\Uref$ (median \azn{gs-urw-median}\%; near-field mean
    \azn{gs-unear-rw-mean}\%, up to \azn{gs-unear-rw-max}\%) and force by a median of
    \azn{gs-rw-median}\% in the metric (\azn{gs-rw-median-unfloored}\% unfloored;
    \azn{gs-rw-median-released}\% and \azn{gs-rw-median-unfloored-released}\% on the released cases
    alone). It also shifts boundary-layer transition: the flat plate's force rises
    \azn{gs-plate-force-ratio}$\times$ (friction \azn{gs-plate-friction-ratio}$\times$;
    \azn{gs-rw-max-unfloored}\% unfloored, \azn{gs-rw-max-unfloored-metric}\% floored), and the
    Blender streamlined body loses most of its laminar boundary layer (\azn{gs-rw-max}\% force
    change). A wall-resolved mesh takes \azn{gs-cost-ratio-wr}$\times$ the median production solver
    time.
\end{itemize}

\paragraph{Additional solver checks.}
Four further checks vary inflow eddy viscosity, convection order, Courant limit, and job restarts.
\begin{itemize}
  \item \emph{Inflow eddy viscosity.} On \azn{inflow-cases} released \key{mean_stable} samples at
    $\Rey\ge10^6$ (Streamlined, Blender bluff, and Blobs), raising inlet $\omega$ at fixed $k$ on
    the identical mesh so that inlet $\nu_t/\nu=1$ shifts near-body velocity
    $\varepsilon_U^{\mathcal N}$ by \azn{inflow-A-epsU-near-two-lo}--\azn{inflow-A-epsU-near-two-hi}\%
    and unfloored force $\lvert\Delta C_F\rvert/\lvert C_F\rvert$ by
    \azn{inflow-A-dCF-two-lo}--\azn{inflow-A-dCF-two-hi}\% on the two lower-intensity cases, and by
    \azn{inflow-A-blobs-epsU-near}\% and \azn{inflow-A-blobs-dCF}\% on the high-intensity Blobs
    case. Lowering inlet $k$ at fixed length scale on the Streamlined case leaves
    \azn{inflow-B-laminar-pct}\% of the body laminar and shifts unfloored force by
    \azn{inflow-B-dCF}\%; because $\lvert C_F\rvert = \azn{inflow-3157-cf} \ll 0.05$ on that case,
    its floored $\varepsilon_{C_F}$ changes by \azn{inflow-3157-A-epsCF}\% and
    \azn{inflow-3157-B-epsCF}\% in the two variants.
  \item \emph{Scheme (steady-RANS sanity check).} On three meshes (up to \azn{fine-cells-max-m}\,M
    cells), \azn{study-cases} external-box cases were solved to steady RANS once with first-order
    upwind momentum convection and once with the recipe's limited second-order scheme. Over
    unweighted wake probes on the finest mesh, the RMS velocity difference is \azn{scheme-u-min} to
    \azn{scheme-u-max}\% of $\Uref$; because this compares a degraded first-order steady run against
    the second-order scheme, it verifies scheme sensitivity rather than estimating the production
    URANS discretization error in \cref{fig:stackup}.
  \item \emph{Time step.} One external-box URANS case was solved twice on the same mesh with Courant
    limits of 5 and 10. The volume-weighted RMS difference between the two mean velocity fields is
    \azn{courant-u-pct}\% of $\Uref$; the force coefficient differs by \azn{courant-cf}.
  \item \emph{Restarts.} One external-box URANS case was solved twice on the same mesh, once without
    interruption and once stopped and restarted \azn{restart-resumes} times. The volume-weighted RMS
    difference between the two mean velocity fields is \azn{restart-u-pct}\% of $\Uref$.
\end{itemize}
\FloatBarrier

\needspace{24\baselineskip}
\section{The sample schema}
\label{app:schema}

Every key of a sample has one role (\cref{fig:schema}b), and only the input keys may enter a model
(the input rule of \cref{sec:schema}): \cref{tab:key-roles} gives each role, followed by the
conventions and the key tree of one released sample.

\begin{table}[!ht]
  \centering
  \caption{The role of each key of a sample, in the order of \cref{fig:schema}b. \key{*} stands for
    every key under a group.}
  \label{tab:key-roles}
  \footnotesize
  \begin{tabular}{@{}l>{\raggedright\arraybackslash}p{0.84\textwidth}@{}}
    \toprule
    Role & Keys \\
    \midrule
    Input & \key{boundaries/*/points} and \key{cells}; \key{boundaries/*/cell_data/prescribed/*};
      \key{global_data/physics/kinematic_viscosity} and \key{density} \\
    Query & \key{interior/points}; the centroids of \key{boundaries/*/cells} \\
    Target & \key{interior/point_data/solution/*}; \key{boundaries/*/cell_data/solution/*};
      \key{no_slip/cell_data/derived/wall_shear_stress} \\
    Weight & \key{interior/point_data/_target_quadrature_measure} (cell volumes); face areas,
      computed from the triangles \\
    Fact & \key{cell_data/annotation/body_id}; \key{no_slip/cell_data/derived/wall_y_plus};
      \key{global_data/schema_version}, \key{identity/*}, \key{solution_definition/*}, and
      \key{diagnostics/*}; \key{global_data/physics/closure_model_id} and \key{body_acceleration} (zero) \\
    Scale & \key{global_data/reference/*} \\
    \bottomrule
  \end{tabular}
\end{table}

\paragraph{Conventions.}
\begin{itemize}[itemsep=1pt]
  \item \emph{Rows.} Interior arrays have one row per cell; boundary \key{cell_data} arrays have one
    row per triangle. Every sample has the same keys, and a boundary with no faces keeps its keys
    with zero rows.
  \item \emph{Units and precision.} SI throughout: lengths in m, velocities in m/s, pressure in Pa
    (gauge, zero at the outlet), $k$ in m$^2$/s$^2$, $\omega$ in 1/s, $\nu_t$ in m$^2$/s; $\gamma$
    and $\Rey_{\theta t}$ are dimensionless. Coordinates, quadrature weights, and field tensors are
    \texttt{float32}.
  \item \emph{Signs.} Triangle normals point out of the fluid. Wall shear stress is the traction of
    the wall on the fluid (OpenFOAM's sign), so the friction force on a body is $-\sum_j\vtau_jA_j$.
  \item \emph{Seams and boundary closure (v1.3).} Points repeat along former patch seams because
    patches are merged without joining coincident vertices. Release v1.3 replaces affected boundary
    polygons in 3{,}396 samples by positive-area triangle fans so that 3{,}781 of 3{,}816 released
    samples (including the mascot) have closed, edge-manifold boundaries (33 keep native OpenFOAM
    four-polygon edges; one sample keeps a double-sheet fan and one keeps three open edges where
    coincident \texttt{float32} points prevent positive-area fans). Computing triangle areas from
    \texttt{float64} edge cross-products avoids underflow on thin fan triangles (813 triangles in 76
    samples). Interior tensors, cell volumes, stability labels, and splits are byte-identical
    between v1.2 and v1.3.
  \item \emph{Symmetry planes.} In the \azn{symmetry-samples} symmetry hemispheres, the mesher
    leaves the symmetry plane as background-grid steps within \azn{symmetry-offset-max-pct}\% of
    $\Lref$ of the fitted plane.
\end{itemize}

\needspace{0.85\textheight}
\paragraph{The key tree.}
The listing below is the key tree of one released sample, read from the v1.3 artifact itself; array
sizes are that sample's. The sample is \texttt{case-0000000000000003016-edcd559b0f11}%
.

\begin{treecode}
DomainMesh
|-- interior  (Mesh)
|   |-- points                                              [4550479, 3] float32
|   |-- cells                                               [0, 1] int64
|   `-- point_data
|       |-- solution
|       |   |-- velocity                                    [4550479, 3] float32
|       |   |-- gauge_pressure                              [4550479] float32
|       |   |-- turbulent_kinetic_energy                    [4550479] float32
|       |   |-- specific_dissipation_rate                   [4550479] float32
|       |   |-- turbulent_kinematic_viscosity               [4550479] float32
|       |   |-- intermittency                               [4550479] float32
|       |   `-- momentum_thickness_reynolds_number          [4550479] float32
|       `-- _target_quadrature_measure                      [4550479] float32
|-- boundaries
|   |-- inlet  (Mesh)
|   |   |-- points                                          [12800, 3] float32
|   |   |-- cells                                           [25122, 3] int64
|   |   `-- cell_data
|   |       |-- solution/*                                  [25122, ...] float32 (seven fields, as interior)
|   |       `-- prescribed
|   |           `-- dirichlet
|   |               |-- velocity                            [25122, 3] float32
|   |               |-- turbulent_kinetic_energy            [25122] float32
|   |               |-- specific_dissipation_rate           [25122] float32
|   |               |-- intermittency                       [25122] float32
|   |               `-- momentum_thickness_reynolds_number  [25122] float32
|   |-- outlet  (Mesh)
|   |   |-- points                                          [17309, 3] float32
|   |   |-- cells                                           [34140, 3] int64
|   |   `-- cell_data
|   |       |-- solution/*                                  [34140, ...] float32 (seven fields, as interior)
|   |       `-- prescribed
|   |           |-- dirichlet
|   |           |   `-- gauge_pressure                      [34140] float32
|   |           `-- reverse_flow/*                          [34140, ...] float32 (five inlet fields)
|   |-- no_slip  (Mesh)
|   |   |-- points                                          [55559, 3] float32
|   |   |-- cells                                           [110368, 3] int64
|   |   `-- cell_data
|   |       |-- solution/*                                  [110368, ...] float32 (seven fields, as interior)
|   |       |-- prescribed
|   |       |   `-- dirichlet
|   |       |       `-- velocity                            [110368, 3] float32
|   |       |-- annotation
|   |       |   `-- body_id                                 [110368] int64
|   |       `-- derived
|   |           |-- wall_shear_stress                       [110368, 3] float32
|   |           `-- wall_y_plus                             [110368] float32
|   `-- slip  (Mesh)
|       |-- points                                          [68480, 3] float32
|       |-- cells                                           [135042, 3] int64
|       `-- cell_data
|           |-- solution/*                                  [135042, ...] float32 (seven fields, as interior)
|           `-- annotation
|               `-- body_id                                 [135042] int64
`-- global_data
    |-- schema_version                                      () int64
    |-- identity/*                                          (provenance & run diagnostics)
    |-- physics
    |   |-- density                                         () float32
    |   |-- kinematic_viscosity                             () float32
    |   |-- body_acceleration                               [3] float32
    |   `-- closure_model_id                                () int64
    |-- reference
    |   |-- velocity                                        [3] float32
    |   |-- secondary_inlet_velocities                      [0, 3] float32
    |   |-- force_projection_direction                      [3] float32
    |   |-- speed                                           () float32
    |   |-- length                                          () float32
    |   `-- dynamic_pressure                                () float32
    |-- solution_definition/*                               (provenance & run diagnostics)
    `-- diagnostics/*                                       (provenance & run diagnostics)
\end{treecode}

Three more kinds of file come with the samples. \key{index.json} has one record per sample
(identity, hashes, stability label, and quality facts), so subsets can be chosen without opening an
artifact. The release manifest (\key{release-v1.3.json}) gives every planned case its status and
every split its effective sizes. Each sample folder holds its posed problem
(\key{case-manifest.json}), its execution history, its validation receipt, and (for the 3{,}396
samples modified in v1.3) a \key{boundary-resplit.json} lineage record.
\FloatBarrier

\section{Metrics in full}
\label{app:metrics}

Every metric is a volume- or area-weighted error in the sample's own reference units, so every
sample counts on the same scale (\cref{tab:metrics}); this appendix defines the composite interior
scores, states the reporting rules, and excerpts the reference implementation.

\begin{table}[!ht]
  \centering
  \caption{Every metric: the quantity whose error it measures, the points or faces it covers, its
    weight, its unit, and the test samples on which it is defined.}
  \label{tab:metrics}
  \footnotesize
  \setlength{\tabcolsep}{2.6pt}
  \begin{tabular}{@{}ll>{\raggedright}p{0.12\textwidth}>{\raggedright}p{0.18\textwidth}l>{\raggedright}p{0.15\textwidth}>{\raggedright\arraybackslash}p{0.135\textwidth}@{}}
    \toprule
    Metric & Task & Error of & Over & Weight & In units of & Defined on \\
    \midrule
    $\varepsilon_U$, $\varepsilon_p$ & interior & velocity, pressure & every interior point & cell volume & $\Uref$, $\qref$ & every sample \\
    $\varepsilon_U^{\mathcal N}$, $\varepsilon_p^{\mathcal N}$ & interior & velocity, pressure & points within $\Lref$ of a body face centroid & cell volume & $\Uref$, $\qref$ & samples with a body \\
    $S_\mathrm{flow}$ & interior (overall) & velocity \& pressure means & split means of $\varepsilon_U, \varepsilon_U^{\mathcal N}, \varepsilon_p, \varepsilon_p^{\mathcal N}$ & $1/4$ each in $\ln$ & free-stream split means & split summary \\
    $S_\mathrm{all}$ & interior (overall) & flow \& turbulence means & adds $\varepsilon_k, \varepsilon_{\nu_t}, \varepsilon_\gamma, \varepsilon_{\Rey_{\theta t}}$ & $1/3$ per group in $\ln$ & free-stream split means & split summary \\
    $\varepsilon_k$, $\varepsilon_\omega$, $\varepsilon_{\nu_t}$ & interior (aux.) & $k$, $\omega$, $\nu_t$ & every interior point & cell volume & $\Uref^2$, $\Uref/\Lref$, $\Uref\Lref$ & every sample \\
    $\varepsilon_\gamma$, $\varepsilon_{\Rey_{\theta t}}$ & interior (aux.) & $\gamma$, $\Rey_{\theta t}$ & every interior point & cell volume & the true field's RMS & every sample \\
    $r_U$, $r_p$ & interior & velocity, pressure & every interior point & none & the true field's norm & every sample \\
    $\varepsilon_{C_F}$, $\varepsilon_{\vC}$ & boundary & force along $\mathbf e_F$; force vector & the body's faces & face area & $\max(\lvert C_F\rvert,0.05)$; $\max(\lVert\vC\rVert,0.05)$ & samples with a body \\
    $\varepsilon_{C_p}$ & boundary & surface pressure & the body's faces & face area & $\qref$ & samples with a body \\
    $\varepsilon_\tau$ & boundary & wall shear stress & the body's no-slip faces & face area & $\qref$ & samples with a body \\
    $\varepsilon_K$ & boundary & total-pressure loss & inlet and outlet faces & mass flux & $\max(\lvert K\rvert,0.05)$ & ducts \\
    \bottomrule
  \end{tabular}
\end{table}

Every metric is computed per test sample in that sample's reference units and then summarized by its
arithmetic mean over the canonical test set. Inlet speeds span a factor of about \azn{speed-span}
across the release and dynamic pressures a factor of about \azn{q-span}, so errors in SI units
averaged over a test set would measure only the fastest samples. In units of $\Uref$ and $\qref$,
every sample counts on the same scale, which is also the scale of the stability label and of
\cref{sec:trust}. Errors are weighted by cell volume or face area, so they approximate integrals
over the fluid or the body's surface rather than sums over a mesh that is finest where the mesher
chose.

\paragraph{Interior task.}
$\varepsilon_U$ and $\varepsilon_p$ are given by \cref{eq:interior}. In a large enclosure most of
the fluid is free stream, which dilutes the errors that matter near the body, so
$\varepsilon_U^{\mathcal N}$ and $\varepsilon_p^{\mathcal N}$ restrict the sums of
\cref{eq:interior} to the near field,
\begin{equation*}
  \mathcal N=\Bigl\{\,i\;:\;\min_{j\in\mathcal B}\lVert\vx_i-\mathbf c_j\rVert\le\Lref\Bigr\},
\end{equation*}
with $\mathbf c_j$ the centroids of the body's faces $\mathcal B$ (defined below; body-free ducts
have no near field). The relative $L^2$ errors common in machine learning count every point equally:
$r_U=\lVert\hatvu-\vu\rVert_2/\lVert\vu\rVert_2$, and $r_p$ likewise for the pressure.

\paragraph{Auxiliary targets.}
The five turbulence fields of the interior task each have an error. With
$\lVert f\rVert_V=\bigl(\sum_iV_if_i^2/\sum_iV_i\bigr)^{1/2}$ the volume-weighted root mean square
over the interior points,
\begin{gather*}
  \varepsilon_k=\frac{\lVert\hat k-k\rVert_V}{\Uref^2},\qquad
  \varepsilon_\omega=\frac{\lVert\hat\omega-\omega\rVert_V}{\Uref/\Lref},\qquad
  \varepsilon_{\nu_t}=\frac{\lVert\hat\nu_t-\nu_t\rVert_V}{\Uref\Lref},\\
  \varepsilon_\gamma=\frac{\lVert\hat\gamma-\gamma\rVert_V}{\lVert\gamma\rVert_V},\qquad
  \varepsilon_{\Rey_{\theta t}}=\frac{\lVert\widehat{\Rey}_{\theta t}-\Rey_{\theta t}\rVert_V}{\lVert\Rey_{\theta t}\rVert_V}.
\end{gather*}
These errors are not on the scale of $\varepsilon_U$. A prediction of zero scores 1 on $\gamma$
and $\Rey_{\theta t}$ only, and the eddy viscosity of a typical sample is about 0.001 in its unit
(\cref{fig:inflow}), so each auxiliary error is compared with the free-stream reference's mean on
the same split rather than with 1.

\paragraph{Overall composite scores.}
To summarize multiple interior metrics on one dimensionless scale without letting any single field's
numeric range dominate, we normalize each split-level mean error $\bar\varepsilon_i$ by the
free-stream reference's mean error $\bar\varepsilon_i^{\mathrm{fs}}$ on the same split and take a
weighted geometric mean:
\begin{equation*}
  S = \exp\Bigl(\sum_i w_i \ln\frac{\bar\varepsilon_i}{\bar\varepsilon_i^{\mathrm{fs}}}\Bigr),
  \qquad \sum_i w_i = 1.
\end{equation*}
Two groupings are defined:
\begin{itemize}[itemsep=1pt]
  \item \emph{Overall flow score $S_\mathrm{flow}$ (\cref{eq:composite}; primary):} equal group
    weight ($1/2$ each) to velocity ($\varepsilon_U, \varepsilon_U^{\mathcal N}$) and pressure
    ($\varepsilon_p, \varepsilon_p^{\mathcal N}$), giving $w_i = 1/4$ for each of the four metrics.
  \item \emph{All-field overall score $S_\mathrm{all}$:} equal group weight ($1/3$ each)
    to velocity ($w_i=1/6$ each), pressure ($w_i=1/6$ each), and turbulence
    ($\varepsilon_k, \varepsilon_{\nu_t}, \varepsilon_\gamma, \varepsilon_{\Rey_{\theta t}}$,
    $w_i=1/12$ each). Specific dissipation $\varepsilon_\omega$ is omitted because $\omega$ spans
    many orders of magnitude near walls and would dilute the remaining turbulence fields.
\end{itemize}
Uniform inlet flow scores $1.000$ on both, and lower is better. Halving any metric within a group
lowers $S$ by the same factor, and changing the reference rescales every model by one constant
without changing their ordering. A composite score is reported only alongside its constituent
per-field means; if a model omits any required field, that composite score is undefined with no
imputation. The boundary task has no composite score.

\paragraph{Boundary task.}
$\mathcal B$ is the set of body faces of \key{no_slip} and \key{slip} (\key{annotation/body_id}
$\ge1$, both bodies when a case has two) and $\mathcal W$ its no-slip subset (body-free ducts have
no body force or body surface errors). The force errors are \cref{eq:force}; note that $C_F =
\vC\cdot\mathbf e_F$ can be negative (e.g., in crossflow or dual-inlet boxes). The surface errors are
\begin{equation*}
  \varepsilon_{C_p}=\frac{1}{\qref}\biggl(\frac{\sum_{j\in\mathcal B}A_j(\hat p_j-p_j)^2}{\sum_{j\in\mathcal B}A_j}\biggr)^{1/2},\qquad
  \varepsilon_\tau=\frac{1}{\qref}\biggl(\frac{\sum_{j\in\mathcal W}A_j\lVert\hat\vtau_j-\vtau_j\rVert^2}{\sum_{j\in\mathcal W}A_j}\biggr)^{1/2}.
\end{equation*}
A duct is scored by its total-pressure loss coefficient $K$, which is positive in every released
duct. Its static pressure drop is not: it is negative in some ducts, where the flow slows enough to
recover more pressure than it loses. The coefficient is
\begin{equation*}
  K=\frac{\langle p_0\rangle_\mathrm{inlet}-\langle p_0\rangle_\mathrm{outlet}}{\qref},\qquad
  \langle p_0\rangle=\frac{\sum_j\bigl(p_j+\rho\lVert\vu_j\rVert^2/2\bigr)\dot m_j}{\sum_j\dot m_j},\qquad
  \dot m_j=\rho A_j\,\vu_j\cdot\vn_j,
\end{equation*}
with $\varepsilon_K=\lvert\hat K-K\rvert/\max(\lvert K\rvert,0.05)$ ($K>0$ on all released ducts).
Because boundary normals $\vn_j$ point out of the fluid, $\dot m_j$ is negative on the inlet and
positive on the outlet; the sign cancels in the ratio $\langle p_0\rangle_\mathrm{inlet}$. On the
inlet, velocity is prescribed and pressure is predicted; on the outlet, the reverse holds. If the
net predicted outlet mass flux satisfies
$\sum_{j\in\mathrm{outlet}}\hat{\dot m}_j \le 0.01\lvert\sum_{j\in\mathrm{inlet}}\dot m_j\rvert$
(including zero or reversed net flow), $\hat K$ is treated as invalid and $\varepsilon_K$ counts as
1 so near-zero denominators cannot dominate a split mean.

The boundary task's metrics evaluate pressure and wall shear on the body and, in a duct, inlet
pressure and outlet velocity. The 0.05 in the denominators of the force errors matches the stability
test (\cref{app:stability}), and $\lvert C_F\rvert<0.05$ in \azn{force-small-pct}\% of samples with
a body. Boundary-task results must be scored on release v1.3, whose boundary meshes are closed.

\paragraph{Reporting.}
A result names the task, the split, the release, and the hash of its catalog (v1.3:
\texttt{\azn{index-sha}}\ldots), the effective sizes, any filtering, and any input outside the
input rule of \cref{sec:schema}. It reports:
\begin{itemize}
  \item each metric's arithmetic mean over the canonical (\key{mean_stable}) test samples on which
    it is defined: field errors on every sample, near-field and body metrics on samples with a body,
    and $\varepsilon_K$ on ducts;
  \item for the interior task, the primary overall flow score $S_\mathrm{flow}$ (and $S_\mathrm{all}$
    when turbulence fields are predicted), the share of samples with
    $\varepsilon_U<\varepsilon_U^\mathrm{fs}$, and the mean $\varepsilon_U$ within each of the eight
    enclosure configurations;
  \item for $\varepsilon_{C_F}$ and $\varepsilon_{\vC}$, the shares within 0.02 (the stability
    test's force tolerance) and 0.05;
  \item the five auxiliary errors if the model predicts the turbulence fields; a model may predict
    velocity and pressure alone, and its result says so;
  \item its number of training seeds, across-seed dispersion (standard deviation or range), how any
    random input such as boundary-face subsampling was averaged, and whether interior query points
    were decoded independently or with cross-query attention;
  \item whether its settings were chosen on the validation part of its own split or reused from
    \key{full} (\cref{app:splits});
  \item optionally, the same scores over all released test samples.
\end{itemize}
Three details matter when scores are compared.
\begin{enumerate}[itemsep=1pt]
  \item A boundary score must leave out fields that a Dirichlet condition fixes on that boundary
    (velocity and the five prescribed turbulence quantities on \key{inlet}, with $\nu_t$ fixed via
    $k/\omega$; velocity on \key{no_slip}; pressure on \key{outlet}), because there the stored mean
    equals the input. If an auxiliary boundary metric scores velocity on \key{slip} walls, it must
    exclude the symmetry hemisphere's plane, where the stored velocity is the adjacent cell's value
    and the plane itself is stepped (\cref{app:schema}); none of \cref{tab:metrics}'s boundary
    metrics scores \key{slip} velocity.
  \item The relative $L^2$ errors $r_U$ and $r_p$ weight every interior point equally, as most
    machine-learning papers do, so they weight the finely meshed boundary layers more than
    $\varepsilon_U$ and $\varepsilon_p$ do. Also, $r_p$ divides by a gauge pressure whose zero is
    at the outlet.
  \item Rank correlations of $C_F$ are not recommended. Across different bodies,
    $\lvert C_F\rvert$ spans a factor of about \azn{force-span}, so a ranking measures the spread
    of the bodies more than the model.
\end{enumerate}

\paragraph{Reference implementation.}
The excerpt below computes per-case scales from the posed problem alone (\texttt{derive\_globals})
and the free-stream reference (\texttt{free\_stream}). Its length is the boundary length $L_b$, the
root-mean-square distance between two points of the boundary surface, which does not change when a
case is rotated. \texttt{free\_stream} computes eddy viscosity as $\bar k/\bar\omega$ from the
area-weighted mean inlet $k$ and $\omega$ (capped at $10^5\nu$, matching the solver). In production
scoring on v1.3, triangle areas and normals should be computed from \texttt{float64} edge
cross-products so thin fan triangles do not underflow in \texttt{float32}, and body-free ducts are
skipped by body and near-field metrics. The remaining metric functions follow
\cref{eq:interior,eq:force} and are tested on problems with analytic answers.
\begin{pycode}
def free_stream(sample):
    """The zero-skill prediction, one value per field for every query point.

    The area-weighted mean of what the inlet prescribes, the eddy viscosity k / omega that
    goes with it (capped at 10^5 nu, as in the solver), and zero gauge pressure. Its wall
    shear stress is zero.
    """
    inlet = sample.boundaries["inlet"]
    area = inlet.cell_areas.double()
    mean = {
        name: torch.einsum("f,f...->...", area, value.double()) / area.sum()
        for name, value in inlet.cell_data["prescribed", "dirichlet"].items()
    }
    nu = float(sample.global_data["physics"]["kinematic_viscosity"])
    mean["gauge_pressure"] = torch.zeros((), dtype=torch.float64)
    mean["turbulent_kinematic_viscosity"] = (
        mean["turbulent_kinetic_energy"] / mean["specific_dissipation_rate"]
    ).clamp(max=1e5 * nu)
    return mean



def derive_globals(sample):
    """Global scales from the posed problem alone: allowed inputs, unlike global_data/reference."""
    inlet = sample.boundaries["inlet"]
    area = inlet.cell_areas.double()
    u = inlet.cell_data["prescribed", "dirichlet", "velocity"].double()
    speed = float((area * u.norm(dim=-1)).sum() / area.sum())  # area-weighted mean inlet speed
    # Length: the RMS distance between two points uniform on all boundary faces, sqrt(2 Var x),
    # from each triangle's exact moments, so a rotation of the case does not change it
    surfaces = [sample.boundaries[n] for n in ("inlet", "outlet", "no_slip", "slip")]
    o = torch.cat([b.points for b in surfaces]).double().mean(0)  # limits cancellation
    total, first, second = 0.0, 0.0, 0.0
    for b in surfaces:
        p, q, r = (b.points.double()[b.cells[:, k]] - o for k in range(3))
        w = b.cell_areas.double()
        total = total + w.sum()
        first = first + (w[:, None] * (p + q + r)).sum(0) / 3
        second = second + (w * (p * p + q * q + r * r + p * q + q * r + r * p).sum(-1)).sum() / 6
    length = float(2 * (second / total - (first / total).pow(2).sum())) ** 0.5
    nu = float(sample.global_data["physics", "kinematic_viscosity"])
    return {
        "inlet_velocity": (area[:, None] * u).sum(0) / area.sum(),  # area-weighted mean
        "inlet_speed": speed,
        "length": length,
        "reynolds_number": speed * length / nu,
    }
\end{pycode}
\FloatBarrier

\section{All splits}
\label{app:splits}

\Cref{tab:all-splits} lists every split with the question it asks and its sizes, in four groups:
the random splits, the held-out enclosures, the held-out geometry families, and the splits by
Reynolds range or body-wall layout. \Cref{tab:freestream-splits} uses the
same order.

\begin{table}[!ht]
  \centering
  \caption{All \azn{splits} splits: the canonical (\key{mean_stable}), released, and planned sizes
    of each part. A held-out split's test set is a whole region of the population, so the test sets
    differ in size: the canonical ones by a factor of \azn{test-size-ratio-canonical}
    (\azn{test-canonical-min} to \azn{test-canonical-max}), the released ones by
    \azn{test-size-ratio-released}, and the planned ones by \azn{test-size-ratio-planned}. They
    differ in difficulty by more; compare models only on the same split. For the eight splits
    that hold out a property of the body, the canonical test size counts samples with a body,
    and the released and planned sizes count all.}
  \label{tab:all-splits}
  \scriptsize
  \setlength{\tabcolsep}{1.9pt}
  \begin{tabular}{@{}l>{\raggedright\arraybackslash}p{0.17\textwidth}rrrrrrrrr@{}}
    \toprule
    Split (\texttt{split\_id}) & Question & \multicolumn{3}{c}{Train} & \multicolumn{3}{c}{Validation} & \multicolumn{3}{c}{Test} \\
    \cmidrule(lr){3-5}\cmidrule(lr){6-8}\cmidrule(lr){9-11}
    & & canon. & released & planned & canon. & released & planned & canon. & released & planned \\
    \midrule
    \multicolumn{11}{@{}l}{\emph{Random}} \\
\texttt{full} & random split of every case, about 70/10/20\% & 2{,}011 & 2{,}713 & 3{,}104 & 290 & 386 & 440 & 515 & 716 & 816 \\
\texttt{scarce} & as \texttt{full}, with $1/6$ of its training part & 321 & 440 & 506 & 290 & 386 & 440 & 515 & 716 & 816 \\
\texttt{super\_scarce} & as \texttt{full}, with $1/36$ of its training part & 53 & 74 & 84 & 290 & 386 & 440 & 515 & 716 & 816 \\
\addlinespace
\multicolumn{11}{@{}l}{\emph{Held-out enclosure}} \\
Open box (\texttt{profile\_open\_opposed}) & hold out the open box & 1{,}943 & 2{,}775 & 3{,}249 & 293 & 405 & 466 & 580 & 635 & 645 \\
Ground-effect box (\texttt{profile\_ground\_effect}) & hold out the ground-effect box & 2{,}011 & 2{,}837 & 3{,}281 & 294 & 400 & 455 & 511 & 578 & 624 \\
Confined box (\texttt{profile\_confined\_opposed}) & hold out the confined box & 2{,}114 & 2{,}963 & 3{,}281 & 301 & 418 & 462 & 401 & 434 & 617 \\
Crossflow box (\texttt{profile\_crossflow}) & hold out the crossflow box & 2{,}298 & 2{,}820 & 3{,}251 & 345 & 417 & 473 & 173 & 578 & 636 \\
Dual-inlet box (\texttt{profile\_dual\_inlet}) & hold out the dual-inlet box & 2{,}044 & 2{,}875 & 3{,}270 & 313 & 424 & 469 & 459 & 516 & 621 \\
Opposed sphere (\texttt{family\_opposed\_sphere}) & hold out the opposed sphere & 2{,}158 & 3{,}017 & 3{,}494 & 302 & 418 & 480 & 356 & 380 & 386 \\
Symmetry hemisphere (\texttt{family\_opposed\_hemisphere}) & hold out the symmetry hemisphere & 2{,}330 & 2{,}982 & 3{,}463 & 352 & 442 & 504 & 134 & 391 & 393 \\
Internal duct (\texttt{family\_duct}) & hold out the internal duct & 2{,}266 & 3{,}055 & 3{,}416 & 348 & 457 & 506 & 202 & 303 & 438 \\
Every enclosure but the boxes (\texttt{family\_non\_box}) & train on boxes; test on every other enclosure & 1{,}850 & 2{,}390 & 2{,}743 & 274 & 351 & 400 & 692 & 1{,}074 & 1{,}217 \\
\addlinespace
\multicolumn{11}{@{}l}{\emph{Held-out geometry family}} \\
Blobs (\texttt{bank\_blobs}) & hold out the Blobs family & 1{,}662 & 2{,}156 & 2{,}434 & 237 & 292 & 326 & 872 & 1{,}367 & 1{,}600 \\
Blender bluff (\texttt{bank\_blender\_bluff}) & hold out the Blender bluff family & 2{,}149 & 2{,}935 & 3{,}363 & 331 & 441 & 497 & 326 & 439 & 500 \\
Procedural cars (\texttt{bank\_cars}) & hold out the Procedural cars family & 2{,}173 & 2{,}965 & 3{,}390 & 321 & 434 & 490 & 310 & 416 & 480 \\
Zero-to-CAD (\texttt{bank\_zero\_to\_cad}) & hold out the Zero-to-CAD family & 2{,}250 & 3{,}040 & 3{,}464 & 332 & 442 & 496 & 224 & 333 & 400 \\
Airframe Foundry (\texttt{bank\_airframes}) & hold out the Airframe Foundry family & 2{,}198 & 3{,}018 & 3{,}471 & 321 & 430 & 489 & 288 & 367 & 400 \\
Streamlined (\texttt{bank\_streamlined}) & hold out the Streamlined family & 2{,}132 & 2{,}940 & 3{,}383 & 332 & 441 & 497 & 337 & 434 & 480 \\
Blender streamlined (\texttt{bank\_blender\_streamlined}) & hold out the Blender streamlined family & 2{,}148 & 2{,}938 & 3{,}385 & 310 & 418 & 475 & 346 & 459 & 500 \\
\addlinespace
\multicolumn{11}{@{}l}{\emph{Reynolds range and body-wall layout}} \\
$\Rey\ge10^6$ (\texttt{reynolds}) & train below $\Rey=10^6$; test above & 1{,}577 & 2{,}119 & 2{,}282 & 238 & 311 & 333 & 1{,}001 & 1{,}385 & 1{,}745 \\
$10^5\le\Rey<10^6$ (\texttt{reynolds\_interior}) & hold out $10^5\le\Rey<10^6$ & 1{,}935 & 2{,}614 & 3{,}048 & 292 & 380 & 440 & 589 & 821 & 872 \\
Three body bands (\texttt{layout}) & train on one or two body bands; test on three & 1{,}647 & 2{,}233 & 2{,}559 & 239 & 311 & 347 & 892 & 1{,}271 & 1{,}454 \\
    \bottomrule
  \end{tabular}
\end{table}

Splits are drawn by hashing. Each case maps to a number in $[0,1)$ through the SHA-256 of the
release's design seed, the name of a split stream, and the digest of the case's source geometry.
Each split cuts that number at its fractions, so proportions hold in expectation rather than
exactly. There are three streams. The first draws the in-distribution split and the second the two
scarce subsets. The third draws the validation part of every held-out split: an eighth of its
training side, never drawn from its held-out region. Because the held-out splits share this
stream, a case drawn for validation in one of them is validation in every other held-out split
whose test region does not contain it.

The hash uses the primary body, which no two cases share, so every partition is disjoint in
geometry. Hashing also keeps a case's part fixed when cases are added. The second body that a case
adds with probability 0.03 is a scaled copy of its own primary body, so it shares no geometry
with any other case, in any part.

\paragraph{Splits that hold out a property of the body.}
The seven geometry-family splits and \key{layout} assign a body-free duct too, by the body or the
body-wall layout that was drawn for it, although no body is posed. Their lists keep those ducts in
the training and validation parts, and their test scores leave them out, so these eight splits score
only test samples that contain a body. The canonical test sizes of \cref{tab:all-splits} count samples with a body. The
\key{mean_stable} body-free ducts that a geometry-family split leaves out of its score number
\azn{bank-test-bodiless-min} to \azn{bank-test-bodiless-max} (\azn{bank-cars-test-bodiless} for
\key{bank_cars}, which scores \azn{split-bank_cars-test-canonical} samples). \key{layout} leaves
out \azn{layout-test-bodiless}, and \azn{layout-test-bodiless-planned} of its planned test cases
are ducts without a body.

\paragraph{Headline splits and model selection.}
The seven headline splits are \key{full}, \key{scarce}, \key{super_scarce}, \key{family_non_box},
\key{profile_dual_inlet}, \key{reynolds}, and \key{bank_cars} (\cref{tab:splits});
\key{profile_dual_inlet} is the one that holds out a layout of boundary conditions. The
geometry-family test sets together cover every case, so every validation sample of one split is a
test sample of another. Of \key{full}'s \azn{split-full-val-canonical} canonical validation
samples, \azn{split-full-val-in-held-out-test} are test samples of \key{family_non_box},
\key{reynolds}, or \key{bank_cars}; conversely, the shared held-out validation stream contains
\azn{split-ho-val-in-full-test} of \key{full}'s \azn{split-full-test-canonical} canonical test
samples (\azn{split-ho-val-in-full-test-pct}\%), so models evaluated on \key{full} must select
strictly on \key{full}'s validation part. Settings chosen on \key{full}'s
validation part may be reused on the other splits if the result says so; otherwise they are chosen
on the validation part of the split that is scored. Failed and timed-out cases keep their places, so
effective sizes are known exactly, and a later release that finishes the timed-out cases moves no
case.

\section{Details of the runs}
\label{app:baselines}

This appendix documents the inputs, unit scalings, training settings, per-enclosure means, held-out
generalization results, controlled boundary-token ablation, auxiliary turbulence means, and
free-stream reference behind \cref{sec:baselines,sec:trust}. All evaluation scores are arithmetic
means over canonical (\key{mean_stable}) test samples (averaged across training seeds for
multi-seed rows). Every interior-task run below was trained and scored on release v1.2; because
interior fields, cell volumes, and splits are byte-identical in v1.3, these interior scores carry
over unchanged.\reviewonly{ \rvPlannedScope}

\paragraph{What each model is given.}
\Cref{tab:adapters} lists the inputs of the model setups of \cref{tab:results}.
The inflow's direction is random from case to case. Prescribed boundary values give it directly,
and in boxes the density of interior query points also marks it because the mesher refines a wedge
from the body to the outlet. Without boundary tokens or a flow axis, a model receives nothing of
the inlets, outlets, or slip walls.

With boundary tokens, GeoTransolver uses an extension of the public model: its ball query takes each
token's centroid, and the token keeps all 22 channels:
\begin{itemize}[itemsep=1pt]
  \item the face centroid (3) and a one-hot condition type (4);
  \item the prescribed velocity, and the velocity prescribed where flow re-enters (3 each);
  \item the prescribed pressure, $k$, $\omega$, intermittency, and $\Rey_{\theta t}$ (5), plus the
    last four where flow re-enters (4).
\end{itemize}
A value that a face does not prescribe is 0. The ball query uses six radii (0.01, 0.05, 0.25, 1,
2.5, and 5) with at most 8, 16, 32, 32, 64, and 128 neighbors, in meters under SI units and in
multiples of the boundary length $L_b$ under per-case scales.

\begin{table}[!ht]
  \centering
  \caption{What each model setup of \cref{tab:results} is given. Transolver and GeoTransolver also
    take one global scalar: standardized log viscosity in SI units and standardized $\log_{10}\Rey_b$
    in per-case scales; \emph{Transolver + flow axis} adds the unit vector from the
    flow-rate-weighted inlet centroid to the area-weighted outlet centroid, and
    \emph{GeoTransolver + BC} adds boundary-condition tokens. ISLA takes $\log_{10}\Rey_b$ and
    computes $L_b, U_b$ inside the model.\reviewonly{ \rvAdaptersNote}}
  \label{tab:adapters}
  \footnotesize
  \begin{tabular}{@{}>{\raggedright\arraybackslash}p{0.23\textwidth}>{\raggedright\arraybackslash}p{0.43\textwidth}>{\raggedright\arraybackslash}p{0.12\textwidth}>{\raggedright\arraybackslash}p{0.11\textwidth}@{}}
    \toprule
    Model setup & Inputs & BC inputs & Parameters \\
    \midrule
    Transolver~\citep{wu2024transolver} (no flow axis) & query points, each with signed distance to
      \key{no_slip} walls and its gradient & no & \azn{t-params-m}\,M \\
    Transolver + flow axis & the same point features, plus a global inlet-to-outlet unit flow-axis
      vector & flow axis & \azn{t-params-m}\,M \\
    GeoTransolver~\citep{adams2025geotransolver} (no BC tokens) & the same point features; geometry
      input is the interior query points & no & \azn{gt-params-m}\,M \\
    GeoTransolver + BC & up to 50{,}000 sampled faces per boundary as 22-channel tokens (centroid,
      condition type, prescribed values); the same point features & yes & \azn{gt-params-m}\,M \\
    ISLA~\citep{sharpe2026isla} & all four boundary meshes (200{,}000 area-weighted tokens
      water-filled across roles) with condition types and prescribed values, plus interior query
      coordinates & yes & \azn{isla-params-m}\,M \\
    \reviewonly{\rvAdapterRows}%
    \bottomrule
  \end{tabular}
\end{table}

\paragraph{The two unit scalings.}
Transolver and GeoTransolver are evaluated under two unit scalings.
\begin{itemize}[itemsep=1pt]
  \item \emph{SI units.} Coordinates stay in meters and are not centered. The only global input is
    standardized $\log_{10}\nu$, fitted on the training samples. Targets and prescribed boundary
    values are scaled by one constant per field across the dataset.
  \item \emph{Per-case scales.} Before entering the model, three scales are computed from the
    boundary faces of each posed problem (face $i$ with area $a_i$, centroid $\vx_i$, and prescribed
    inlet velocity $\vu_i$):
    \begin{equation*}
      L_b=\bigl(2\operatorname{Var}_A(\vx)\bigr)^{1/2},\qquad
      U_b=\biggl(\frac{\sum_{i\in\mathrm{inlet}}a_i\lVert\vu_i\rVert^2}{\sum_{i\in\mathrm{inlet}}a_i}\biggr)^{1/2},\qquad
      \Rey_b=\frac{U_b L_b}{\nu},
    \end{equation*}
    where $\operatorname{Var}_A(\vx)$ is the area-weighted variance of position over all faces of
    the four boundary meshes together, each triangle's second moment included. Coordinates and
    signed distances are divided by $L_b$, velocities by $U_b$, pressure by $\rho U_b^2/2$
    ($\rho=1.225$\,kg/m$^3$), $k$ by $U_b^2$, $\omega$ by $U_b/L_b$, and $\nu_t$ by $U_b L_b$.
    Standardized $(\log_{10}\Rey_b-m)/s$ (constants $m, s$ fitted on the split's training set)
    replaces viscosity as the global input; $L_b$ and $U_b$ themselves are not passed. Predictions
    are returned to SI units before scoring.
\end{itemize}
Because $L_b$, $U_b$, and $\Rey_b$ depend only on the posed problem and read no key of
\key{global_data/reference}, the input rule of \cref{sec:schema} allows them. ISLA computes these
exact $L_b$ and $U_b$ inside the model~\citep{sharpe2026isla}. \texttt{derive\_globals}
(\cref{app:metrics}) computes the same $L_b$ and uses the area-weighted mean of $\lVert\vu_i\rVert$
for speed (identical to $U_b$ whenever inlet speed is uniform across inlets).

\paragraph{Training, validation selection, and scoring.}
Every run trained on the canonical training samples of its split in PhysicsNeMo~\citep{physicsnemo}
for 250 epochs (125{,}500 optimizer steps on \key{full}), data-parallel across four NVIDIA GB300
GPUs (one case per GPU, four cases per step), drawing 200{,}000 interior query points per case and
step. All runs minimize the unweighted per-case relative $L^2$ loss summed across all seven fields
(velocity weighted 3, others 1) using Muon~\citep{jordan2024muon} on matrix weights and AdamW
(weight decay $10^{-4}$) on remaining parameters. All test scores evaluate the final checkpoint
(epoch 250) in float32 at full mesh resolution, reduced in float64 with cell-volume weights.
\begin{itemize}[itemsep=1pt]
  \item \emph{Multi-seed runs on \key{full} (bfloat16; \cref{tab:results,tab:results-rows}).}
    We trained Transolver (width 256, 12 layers, eight heads, 256 slices, \azn{t-params-m}\,M
    parameters; \azn{tr-gpu-hours}\,GPU-h per run), GeoTransolver (same backbone plus six-radius
    ball query, \azn{gt-params-m}\,M; \azn{gt-gpu-hours}\,GPU-h), and ISLA (width 512, 12 layers,
    128 slices, \azn{isla-params-m}\,M; \azn{isla-gpu-hours}\,GPU-h). Each ran a six-arm validation
    sweep on \key{full} comparing StepLR against five-epoch warm-up + cosine decay at $1/3$, $1$,
    and $3\times$ each model's reference rate ($10^{-3}$ for Transolver and GeoTransolver,
    $3.75\times10^{-4}$ for ISLA). All three selected warm-up + cosine: $10^{-3}$ for
    Transolver + flow axis and GeoTransolver + BC, and $1.125\times10^{-3}$ for ISLA, each trained
    for \azn{isla-seeds} seeds (42 to 47) and scored with one boundary draw. Over those
    \azn{isla-seeds} seeds, GeoTransolver + BC's errors are \azn{gt-over-isla-S-2dp}
    [\azn{gt-over-isla-S-lo}, \azn{gt-over-isla-S-hi}] ($S_\mathrm{flow}$),
    \azn{gt-over-isla-epsU-2dp} [\azn{gt-over-isla-epsU-lo}, \azn{gt-over-isla-epsU-hi}]
    (velocity), and \azn{gt-over-isla-epsp-2dp} [\azn{gt-over-isla-epsp-lo},
    \azn{gt-over-isla-epsp-hi}] (pressure) times ISLA's (90\% Welch CIs). A float32 check of
    GeoTransolver + BC (step decay, seed 42) agreed closely with bfloat16 on \key{full}'s
    validation set (mean $\varepsilon_U$ \azn{val-fp32-epsU} vs.\ \azn{val-bf16-epsU}).
    \emph{ISLA + boundary targets} (\azn{isla-joint-seeds} seeds, 42 to 45;
    \azn{isla-joint-gpu-hours}\,GPU-h, $1.46\times$ step time) also fits boundary pressure,
    velocity, and wall shear, selecting $1.125\times10^{-3}$ over $3.75\times10^{-4}$ on
    \key{full}'s validation set. These runs match \citet{sharpe2026isla} digit for digit; five of
    the 515 test samples account for nearly half of ISLA's mean $\varepsilon_p$
    (\azn{isla-full-epsp} overall vs.\ \azn{isla-full-epsp-notop5} without them). Total compute for
    the multi-seed interior-only sweep, headline, ablation, and GeoTransolver held-out runs was
    \azn{ms-gpuh-total}\,GPU-h, plus \azn{isla-joint-gpuh-total}\,GPU-h for the four
    boundary-target ISLA runs.
  \item \emph{Single-seed sensitivity runs on \key{full} (\cref{tab:results-rows}).}
    An additional \azn{single-rows} single-seed runs (\azn{single-gpu-hours}\,GPU-h, averaged over
    two boundary-face scoring draws; \azn{gtb-infer-seconds}\,s and \azn{gtb-infer-gib}\,GiB per
    sample for GeoTransolver + BC) span three precision/schedule settings across the
    SI-unit and boundary-blind per-case setups, supplying those contrast rows of \cref{tab:results}
    (a single-seed float32 GeoTransolver + BC run supplies the qualitative slices in
    \cref{fig:predictions}). Without boundary tokens in per-case scales, GeoTransolver's box errors
    are \azn{gt-notok-boxes-lo}--\azn{gt-notok-boxes-hi} vs.\ over \azn{gt-notok-spheres-lo} in
    spheres, aided by ball-query radii up to five domain lengths.
\end{itemize}

\begin{table}[!ht]
  \centering
  \caption{All runs on \key{full} (\azn{fs-full-n} canonical test samples): reference predictors,
    multi-seed headline runs (with across-seed SD in parentheses), and single-seed
    precision/schedule sensitivity runs for the SI-unit and boundary-blind contrasts. GPU-hours are
    per run. Test columns give the primary overall flow score $S_\mathrm{flow}$ (\cref{eq:composite}),
    arithmetic means of $\varepsilon_U$ and $\varepsilon_p$, and unweighted relative $L^2$ errors
    $r_U, r_p$. The last four columns give end-of-training relative $L^2$ errors on training and
    validation (200{,}000 sampled points; zero scores 1) and the lowest validation-loss epoch
    (``--'': not recorded for multi-seed or boundary-target runs).}
  \label{tab:results-rows}
  \scriptsize
  \setlength{\tabcolsep}{2.7pt}
  \begin{tabular}{@{}llr >{\columncolor{overallbg}}r rrrr rrr r@{}}
    \toprule
    & & & \multicolumn{1}{>{\columncolor{overallbg}}c}{\textbf{Overall}\,$\bm{\downarrow}$} & \multicolumn{4}{c}{Test means (\key{full}) $\downarrow$} & \multicolumn{3}{c}{Last-epoch rel.\ $L^2$ $\downarrow$} & \\
    \cmidrule(lr){4-4}\cmidrule(lr){5-8}\cmidrule(lr){9-11}
    Model / unit scaling & LR / setting & GPU-h & $\bm{S_\mathrm{flow}}$
      & $\varepsilon_U$ & $\varepsilon_p$ & $r_U$ & $r_p$
      & train $r_U$ & val.\ $r_U$ & val.\ $r_p$ & \shortstack{best\\ep.} \\
    \midrule
    Free-stream reference & -- & -- & 1.000 & 0.440 & 1.809 & 0.389 & 1.000 & -- & -- & -- & -- \\
Zero predictor & -- & -- & 1.648 & 1.203 & 1.809 & 1.000 & 1.000 & -- & -- & -- & -- \\
\midrule
\multicolumn{12}{@{}l}{\emph{Multi-seed runs on \texttt{full} (six seeds, four for ISLA + boundary targets; per-case scales or unit-invariant, bfloat16, cosine)}} \\
Transolver + flow axis & $10^{-3}$ & 25 & 0.679 ($\pm$0.015) & 0.306 ($\pm$0.006) & 1.212 ($\pm$0.036) & 0.276 & 1.419 & 0.110 & 0.290 & 2.10 & -- \\
GeoTransolver + BC & $10^{-3}$ & 66 & 0.290 ($\pm$0.004) & 0.152 ($\pm$0.002) & 0.427 ($\pm$0.009) & 0.180 & 0.623 & 0.072 & 0.175 & 0.60 & -- \\
ISLA & $1.125{\times}10^{-3}$ & 113 & 0.153 ($\pm$0.003) & 0.082 ($\pm$0.002) & 0.196 ($\pm$0.008) & 0.110 & 0.304 & 0.060 & 0.110 & 0.30 & -- \\
ISLA + boundary targets & $1.125{\times}10^{-3}$ & 166 & 0.123 ($\pm$0.002) & 0.080 ($\pm$0.001) & 0.124 ($\pm$0.003) & 0.106 & 0.281 & -- & -- & -- & -- \\
\addlinespace
\multicolumn{12}{@{}l}{\emph{Transolver (no flow axis) (9.0\,M params; single-seed sensitivity runs)}} \\
SI units & float32, cosine & 78 & 1.648 & 1.202 & 1.810 & 0.999 & 1.009 & 0.999 & 0.999 & 1.00 & 250 \\
SI units & float32, step & 74 & 1.647 & 1.201 & 1.808 & 0.998 & 1.015 & 0.997 & 0.997 & 1.01 & 250 \\
SI units & bfloat16, step & 31 & 1.648 & 1.200 & 1.811 & 0.997 & 1.092 & 0.997 & 0.996 & 1.06 & 211 \\
per-case scales & float32, cosine & 78 & 1.042 & 0.617 & 1.419 & 0.539 & 2.192 & 0.113 & 0.594 & 3.36 & 46 \\
per-case scales & float32, step & 73 & 1.060 & 0.622 & 1.455 & 0.543 & 1.833 & 0.134 & 0.568 & 2.03 & 41 \\
per-case scales & bfloat16, step & 31 & 1.112 & 0.633 & 1.561 & 0.554 & 2.015 & 0.133 & 0.587 & 2.24 & 46 \\
\addlinespace
\multicolumn{12}{@{}l}{\emph{GeoTransolver (no BC tokens) (27.6\,M params; single-seed sensitivity runs)}} \\
SI units & float32, cosine & 182 & 1.638 & 1.184 & 1.811 & 0.981 & 1.015 & 0.981 & 0.983 & 1.01 & 250 \\
SI units & float32, step & 188 & 1.644 & 1.196 & 1.809 & 0.993 & 1.027 & 0.993 & 0.994 & 1.01 & 250 \\
SI units & bfloat16, step & 59 & 1.643 & 1.191 & 1.812 & 0.988 & 1.133 & 0.987 & 0.990 & 1.06 & 250 \\
per-case scales & float32, cosine & 179 & 1.438 & 0.898 & 1.839 & 0.753 & 1.935 & 0.106 & 0.620 & 5.18 & 36 \\
per-case scales & float32, step & 173 & 1.461 & 0.918 & 1.856 & 0.774 & 1.937 & 0.154 & 0.621 & 2.76 & 31 \\
per-case scales & bfloat16, step & 57 & 1.519 & 1.000 & 1.840 & 0.843 & 1.829 & 0.148 & 0.617 & 3.37 & 31 \\
\addlinespace
\multicolumn{12}{@{}l}{\emph{GeoTransolver + BC (27.6\,M params; SI units only; single-seed runs)}} \\
SI units & float32, cosine & 190 & 1.040 & 0.475 & 1.809 & 0.401 & 1.002 & 0.373 & 0.395 & 1.00 & 250 \\
SI units & float32, step & 192 & 1.599 & 1.128 & 1.816 & 0.932 & 1.188 & 0.925 & 0.932 & 1.09 & 226 \\
SI units & bfloat16, step & 69 & 1.644 & 1.140 & 1.907 & 0.943 & 1.989 & 0.927 & 0.940 & 1.53 & 211 \\
    \bottomrule
  \end{tabular}
\end{table}

\begin{figure}[!ht]
  \centering
  \includegraphics[width=0.76\textwidth]{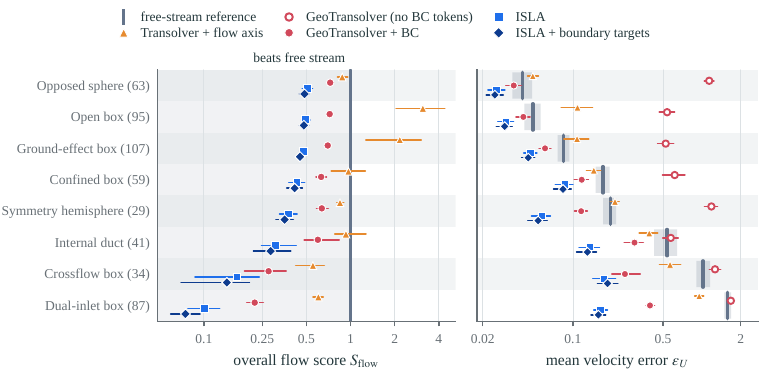}
  \caption{Per-enclosure accuracy on \key{full} (test counts in parentheses; 95\% bootstrap
    intervals; six seeds for Transolver + flow axis, GeoTransolver + BC, and ISLA; four for ISLA +
    boundary targets; single-seed float32 for GeoTransolver without BC tokens). Left: overall flow
    score $S_\mathrm{flow}$ ($<1$, shaded: beats free stream); GeoTransolver + BC, ISLA, and ISLA +
    boundary targets beat uniform flow in all eight enclosures, with ISLA + boundary targets lowest
    in every enclosure. Right: mean $\varepsilon_U$ beside the free-stream reference (vertical bar)
    and boundary-blind GeoTransolver.}
  \label{fig:results}
\end{figure}

\paragraph{Held-out splits.}
\Cref{tab:heldout} reports both GeoTransolver + BC and ISLA across \key{full} and six held-out
splits. On each held-out split, GeoTransolver + BC selected between StepLR and warm-up + cosine at
$10^{-3}$ on that split's validation set (StepLR on \key{family_non_box} and
\key{profile_dual_inlet}, cosine on the other four) and trained \azn{gt-heldout-seeds} seeds
(42 to 44). ISLA trained \azn{isla-heldout-seeds} seeds (42 to 43) per held-out split with warm-up
+ cosine at its reference rate $3.75\times10^{-4}$ and no per-split tuning; \cref{tab:heldout} and
\cref{fig:stackup} therefore pair ISLA's held-out runs with ISLA trained at $3.75\times10^{-4}$ on
\key{full} (two seeds; mean $\varepsilon_U=\azn{isla-ho-full-epsU}$,
$S_\mathrm{flow}=\azn{isla-ho-full-S}$).

With two to three seeds per held-out split, seed variance is sizable on the hardest splits
(per-seed SD of $\ln\varepsilon$ up to 0.26 on \key{family_non_box}, \key{reynolds}, and
\key{profile_crossflow}), leaving per-split velocity and pressure differences on those splits and
\key{profile_dual_inlet} inconclusive under 90\% Welch intervals (on \key{profile_crossflow},
$S_\mathrm{flow}$ still favors ISLA). Pooled across all seven splits at
ISLA's matched $3.75\times10^{-4}$ rate (Welch--Satterthwaite 95\% CIs from observed seed spreads),
GeoTransolver + BC's geometric-mean error is \azn{pooled-body-gt-isla-S}
[\azn{pooled-body-gt-isla-S-lo}, \azn{pooled-body-gt-isla-S-hi}] times ISLA's on
$S_\mathrm{flow}$, \azn{pooled-body-gt-isla-epsU} [\azn{pooled-body-gt-isla-epsU-lo},
\azn{pooled-body-gt-isla-epsU-hi}] times ISLA's on velocity, and \azn{pooled-body-gt-isla-epsp}
[\azn{pooled-body-gt-isla-epsp-lo}, \azn{pooled-body-gt-isla-epsp-hi}] times ISLA's on pressure
across the canonical test sets \citep[matching][]{sharpe2026isla}.

\begin{table}[!ht]
  \centering
  \caption{Unified held-out generalization of GeoTransolver + BC (\azn{gt-seeds} seeds on
    \key{full}, \azn{gt-heldout-seeds} per held-out split) and ISLA (interior targets only,
    $3.75\times10^{-4}$ across all rows, \azn{isla-heldout-seeds} seeds per split). Columns give
    split training and canonical test counts (with shared \key{full}-test count in parentheses),
    free-stream reference means, each model's overall flow score $S_\mathrm{flow}$
    (\cref{eq:composite}), arithmetic mean $\varepsilon_U$ and $\varepsilon_p$, and the paired
    degradation factor $\times$ (held-out mean $\varepsilon_U$ divided by \key{full}-trained mean
    $\varepsilon_U$ on the exact shared test cases).}
  \label{tab:heldout}
  \scriptsize
  \setlength{\tabcolsep}{2.1pt}
  \begin{tabular}{@{}llrr rr >{\columncolor{overallbg}}rrrr@{\hspace{5pt}}>{\columncolor{overallbg}}rrrr@{}}
    \toprule
    & & & & \multicolumn{2}{c}{Reference $\downarrow$} & \multicolumn{4}{c}{GeoTransolver + BC $\downarrow$} & \multicolumn{4}{c}{ISLA ($3.75{\times}10^{-4}$) $\downarrow$} \\
    \cmidrule(lr){5-6}\cmidrule(lr){7-10}\cmidrule(l){11-14}
    Split & Held out & Train & Test (sh.) & $\varepsilon_U$ & $\varepsilon_p$
      & $\bm{S_\mathrm{flow}}$ & $\varepsilon_U$ & $\varepsilon_p$ & $\times$
      & $\bm{S_\mathrm{flow}}$ & $\varepsilon_U$ & $\varepsilon_p$ & $\times$ \\
    \midrule
    \texttt{full} & nothing (random) & 2{,}011 & 515 & 0.440 & 1.809 & 0.290 & 0.152 & 0.427 & -- & 0.186 & 0.082 & 0.297 & -- \\
\midrule
\texttt{bank\_blobs} & blob bodies & 1{,}662 & 872 (147) & 0.409 & 1.655 & 0.303 & 0.154 & 0.355 & 1.1$\times$ & 0.205 & 0.085 & 0.294 & 1.0$\times$ \\
\texttt{bank\_zero\_to\_cad} & CAD-part bodies & 2{,}250 & 224 (30) & 0.466 & 2.004 & 0.304 & 0.164 & 0.452 & 0.9$\times$ & 0.236 & 0.117 & 0.362 & 1.0$\times$ \\
\texttt{reynolds} & $\Rey \ge 10^6$ & 1{,}577 & 1{,}001 (190) & 0.379 & 1.640 & 0.364 & 0.161 & 0.498 & 1.3$\times$ & 0.352 & 0.125 & 0.576 & 1.8$\times$ \\
\texttt{family\_non\_box} & non-box enclosures & 1{,}850 & 692 (133) & 0.238 & 0.595 & 1.745 & 0.328 & 0.979 & 2.4$\times$ & 1.085 & 0.257 & 0.747 & 3.8$\times$ \\
\texttt{profile\_crossflow} & crossflow layout & 2{,}298 & 173 (34) & 0.947 & 3.490 & 1.243 & 1.450 & 3.605 & 5.9$\times$ & 0.773 & 0.656 & 2.906 & 4.7$\times$ \\
\texttt{profile\_dual\_inlet} & dual-inlet layout & 2{,}044 & 459 (87) & 1.659 & 9.134 & 0.904 & 1.456 & 8.463 & 3.6$\times$ & 0.909 & 1.450 & 8.379 & 8.4$\times$ \\
    \bottomrule
  \end{tabular}
\end{table}

\paragraph{Controlled boundary-token ablation.}
\Cref{tab:ablation} isolates what boundary tokens carry for GeoTransolver on \key{full} (bfloat16,
warm-up + cosine at $10^{-3}$, epoch 250). Showing both the \azn{gt-seeds}-seed reference and seed
42 alone alongside three seed-42 ablations with identical random draws: zeroing condition types and
prescribed values leaves face centroids alone and raises $S_\mathrm{flow}$ to \azn{abl-geom-S}
(matching uniform flow); restoring condition types without values recovers \azn{abl-type-S} yet
trails full tokens in dual-inlet boxes (\azn{abl-type-dual} vs.\ \azn{gt-full-dual}); and removing
boundary tokens and passing the area-weighted mean inlet velocity as a global input reaches
\azn{abl-notok-S} overall yet misses outlet placement in crossflow boxes
(\azn{abl-notok-crossflow} vs.\ \azn{gt-full-crossflow}).

\begin{table}[!ht]
  \centering
  \caption{Controlled boundary-token ablation of GeoTransolver on \key{full} (per-case scales,
    bfloat16, warm-up + cosine at $10^{-3}$, epoch 250, \azn{fs-full-n} canonical test samples,
    means). The first row averages six seeds; all remaining rows use seed 42 with identical random
    draws. \emph{Condition types only} and \emph{Face centroids only} keep the full architecture
    with the remaining token channels zeroed; \emph{No boundary tokens + global mean inlet $\vu$}
    removes the boundary tokens and ball-query branches and passes area-weighted mean inlet velocity
    beside $\Rey_b$ as a global input.}
  \label{tab:ablation}
  \scriptsize
  \setlength{\tabcolsep}{3.8pt}
  \begin{tabular}{@{}lr >{\columncolor{overallbg}}r rrrr rrr@{}}
    \toprule
    & & \multicolumn{5}{c}{All canonical test samples (\azn{fs-full-n})}
      & \multicolumn{3}{c}{Mean $\varepsilon_U$ by enclosure $\downarrow$} \\
    \cmidrule(lr){3-7}\cmidrule(l){8-10}
    Variant & Params & $\bm{S_\mathrm{flow}}{\downarrow}$ & $\varepsilon_U{\downarrow}$
      & $\varepsilon_U^{\mathcal N}{\downarrow}$ & $\varepsilon_p{\downarrow}$ & ${<}$ref.\,$\uparrow$
      & Open box & Crossflow & Dual inlet \\
    \midrule
    Full boundary tokens (six-seed ref.) & 27.6\,M & 0.290 & 0.152 & 0.179 & 0.427 & 86\% & 0.041 & 0.253 & 0.396 \\
Full boundary tokens (seed 42) & 27.6\,M & 0.296 & 0.155 & 0.181 & 0.437 & 86\% & 0.041 & 0.263 & 0.408 \\
\midrule
Condition types only (values zeroed) & 27.6\,M & 0.338 & 0.192 & 0.207 & 0.484 & 66\% & 0.051 & 0.267 & 0.519 \\
No boundary tokens + global mean inlet $\vu$ & 9.0\,M & 0.369 & 0.202 & 0.213 & 0.569 & 75\% & 0.044 & 0.499 & 0.504 \\
Face centroids only (types \& values zeroed) & 27.6\,M & 0.997 & 0.519 & 0.519 & 1.544 & 42\% & 0.121 & 0.955 & 1.130 \\
    \bottomrule
  \end{tabular}
\end{table}

\paragraph{The free-stream reference.}
The free-stream reference's mean error depends strongly on the enclosure (\cref{tab:freestream-splits}):
it is small in open boxes and opposed spheres (\azn{fs-open-epsU} and \azn{fs-sphere-epsU}) and
large in crossflow and dual-inlet boxes (\azn{fs-crossflow-epsU} and \azn{fs-dual-epsU}). The
reference was scored on all \azn{fs-cases} released samples with the reference implementation
(\cref{app:metrics}); on boundary meshes unchanged in v1.3, body forces agree with the dataset's
case table to within $\azn{fs-force-gap}$ of $\max(\lvert C_F\rvert, 0.05)$, and v1.3 body-force
reference scores use the exact v1.3 case forces. On \key{full}, the reference's mean $\varepsilon_U$
is \azn{fs-full-epsU} on canonical test samples and \azn{fs-full-epsU-released} on all released test
samples.

\begin{table}[!ht]
  \centering
  \caption{The free-stream reference on the canonical test samples of every split, in the order of
    \cref{tab:all-splits}: arithmetic means of the four flow interior errors and share of samples
    with a body whose force error $\varepsilon_{C_F}\le0.02$. \key{scarce} and \key{super_scarce}
    share the test set of \key{full}. On \key{full}, the reference's mean $\varepsilon_{C_F}$ is
    \azn{fs-full-epsCF} and its mean duct-loss error $\varepsilon_K$ is \azn{fs-full-epsK}.}
  \label{tab:freestream-splits}
  \footnotesize
  \setlength{\tabcolsep}{4.5pt}
  \begin{tabular}{@{}lrrrrrr@{}}
    \toprule
    & & \multicolumn{4}{c}{Mean interior error $\downarrow$} & $\varepsilon_{C_F}$ \\
    \cmidrule(lr){3-6}\cmidrule(l){7-7}
    Test set (split) & Samples & $\varepsilon_U$ & $\varepsilon_U^{\mathcal N}$
      & $\varepsilon_p$ & $\varepsilon_p^{\mathcal N}$ & $\le0.02$ $\uparrow$ \\
    \midrule
    \multicolumn{7}{@{}l}{\emph{Random}} \\
Random cases (\texttt{full}, \texttt{scarce}, \texttt{super\_scarce}) & 515 & 0.440 & 0.442 & 1.809 & 2.014 & 0.2\% \\
\addlinespace
\multicolumn{7}{@{}l}{\emph{Held-out enclosure}} \\
Open box (\texttt{profile\_open\_opposed}) & 580 & 0.046 & 0.115 & 0.026 & 0.071 & 0.2\% \\
Ground-effect box (\texttt{profile\_ground\_effect}) & 511 & 0.091 & 0.139 & 0.051 & 0.090 & 0.2\% \\
Confined box (\texttt{profile\_confined\_opposed}) & 401 & 0.186 & 0.188 & 0.143 & 0.158 & 0.0\% \\
Crossflow box (\texttt{profile\_crossflow}) & 173 & 0.947 & 0.907 & 3.490 & 3.856 & 0.6\% \\
Dual-inlet box (\texttt{profile\_dual\_inlet}) & 459 & 1.659 & 1.612 & 9.134 & 9.886 & 0.0\% \\
Opposed sphere (\texttt{family\_opposed\_sphere}) & 356 & 0.039 & 0.129 & 0.022 & 0.081 & 0.0\% \\
Symmetry hemisphere (\texttt{family\_opposed\_hemisphere}) & 134 & 0.179 & 0.143 & 0.095 & 0.098 & 0.0\% \\
Internal duct (\texttt{family\_duct}) & 202 & 0.627 & 0.225 & 1.937 & 2.329 & 0.0\% \\
Every enclosure but the boxes (\texttt{family\_non\_box}) & 692 & 0.238 & 0.147 & 0.595 & 0.430 & 0.0\% \\
\addlinespace
\multicolumn{7}{@{}l}{\emph{Held-out geometry family}} \\
Blobs (\texttt{bank\_blobs}) & 872 & 0.409 & 0.426 & 1.655 & 1.838 & 0.0\% \\
Blender bluff (\texttt{bank\_blender\_bluff}) & 326 & 0.367 & 0.398 & 1.268 & 1.419 & 0.0\% \\
Procedural cars (\texttt{bank\_cars}) & 310 & 0.460 & 0.477 & 2.008 & 2.218 & 0.0\% \\
Zero-to-CAD (\texttt{bank\_zero\_to\_cad}) & 224 & 0.466 & 0.507 & 2.004 & 2.234 & 0.0\% \\
Airframe Foundry (\texttt{bank\_airframes}) & 288 & 0.391 & 0.394 & 1.793 & 1.980 & 0.0\% \\
Streamlined (\texttt{bank\_streamlined}) & 337 & 0.468 & 0.449 & 2.338 & 2.600 & 0.3\% \\
Blender streamlined (\texttt{bank\_blender\_streamlined}) & 346 & 0.490 & 0.484 & 2.446 & 2.616 & 0.6\% \\
\addlinespace
\multicolumn{7}{@{}l}{\emph{Reynolds range and body-wall layout}} \\
$\Rey\ge10^6$ (\texttt{reynolds}) & 1{,}001 & 0.379 & 0.382 & 1.640 & 1.773 & 0.3\% \\
$10^5\le\Rey<10^6$ (\texttt{reynolds\_interior}) & 589 & 0.463 & 0.465 & 2.137 & 2.391 & 0.0\% \\
Three body bands (\texttt{layout}) & 892 & 0.435 & 0.442 & 2.049 & 2.233 & 0.2\% \\
    \bottomrule
  \end{tabular}
\end{table}

\paragraph{Training fit and auxiliary turbulence errors.}
End-of-training relative $L^2$ diagnostics (\cref{tab:results-rows}) separate the failure modes of
\cref{sec:baselines}. In SI units without boundary tokens, training $r_U$ ends at
\azn{fit-si-plain-train-U-lo}--\azn{fit-si-plain-train-U-hi}, so no model fits the training set. In
per-case scales without boundary tokens, training $r_U$ reaches
\azn{fit-scaled-plain-train-U-lo}--\azn{fit-scaled-plain-train-U-hi} whereas validation $r_U$ stays at
\azn{fit-scaled-plain-val-U-lo}--\azn{fit-scaled-plain-val-U-hi} (best validation at epochs
\azn{fit-scaled-plain-best-epoch-lo}--\azn{fit-scaled-plain-best-epoch-hi}, validation $r_p$
\azn{fit-scaled-plain-val-p-lo}--\azn{fit-scaled-plain-val-p-hi}). Under the seven-field loss weights
(velocity 3, others 1), the five turbulence fields contribute \azn{fit-turb-train-share-pct}\% of
training loss ($\omega$ alone \azn{fit-omega-train-share-pct}\%) and \azn{fit-turb-val-share-pct}\%
of validation loss, leaving \azn{fit-p-val-share-pct}\% on pressure; ISLA uses the same weights and
learns pressure well, so loss weighting alone does not explain Transolver's and GeoTransolver's
pressure errors.

\Cref{tab:results-aux} reports arithmetic means of the auxiliary turbulence errors on \key{full}.
Because $\varepsilon_\omega$ is normalized by $\Uref/\Lref$, far below wall dissipation rates, it
remains near $2\times10^3$ for every model. On $k$ and $\nu_t$, GeoTransolver + BC
lowers mean error below the free-stream reference (\azn{aux-gt-k} vs.\ \azn{aux-fs-k} and
\azn{aux-gt-nut-m} vs.\ \azn{aux-fs-nut-m}$\times10^{-3}$), and ISLA lowers both further
(\azn{aux-isla-k} and \azn{aux-isla-nut-m}$\times10^{-3}$), reaching mean relative $L^2$ errors of
\azn{aux-isla-rel-gamma} on $\gamma$ and \azn{aux-isla-rel-retheta} on $\Rey_{\theta t}$
(unweighted relative $L^2$, as in \citet{sharpe2026isla}).

\begin{table}[!ht]
  \centering
  \caption{Arithmetic mean auxiliary turbulence errors on \key{full} (\azn{fs-full-n} canonical test
    samples): reference-scaled $\varepsilon_k$, $\varepsilon_\omega$, and $\varepsilon_{\nu_t}$
    ($\times10^3$), plus unweighted relative $L^2$ errors $r_\gamma$ and $r_{\Rey_{\theta t}}$ for
    the multi-seed runs (as in \citet{sharpe2026isla}; six seeds per multi-seed row, four for ISLA
    + boundary targets, one for SI-unit Transolver). Bold marks the strictly lowest learned error
    per column (ties are plain).}
  \label{tab:results-aux}
  \footnotesize
  \setlength{\tabcolsep}{4.5pt}
  \begin{tabular}{@{}llrrrrr@{}}
    \toprule
    & & \multicolumn{5}{c}{Mean auxiliary error $\downarrow$} \\
    \cmidrule(l){3-7}
    Model & Units & $\varepsilon_k$ & $\varepsilon_\omega$ & $\varepsilon_{\nu_t}\times10^3$
      & $r_\gamma$ & $r_{\Rey_{\theta t}}$ \\
    \midrule
    Free-stream reference & -- & 0.0110 & 2{,}033.4 & 2.95 & -- & -- \\
\midrule
Transolver (no flow axis) & SI & 0.0120 & 2{,}033.7 & 3.54 & -- & -- \\
Transolver + flow axis & scaled & 0.0114 & 2{,}010.4 & 3.47 & 0.099 & 0.868 \\
GeoTransolver + BC & scaled & 0.0089 & 2{,}002.7 & 1.19 & 0.065 & 0.145 \\
ISLA & unit-inv. & 0.0070 & \textbf{1{,}960.4} & 0.70 & \textbf{0.050} & 0.103 \\
ISLA + boundary targets & unit-inv. & \textbf{0.0069} & 1{,}962.4 & \textbf{0.69} & 0.051 & 0.103 \\
    \bottomrule
  \end{tabular}
\end{table}
\FloatBarrier

\reviewonly{\rvFurtherResults}

\section{Scaling and the irreducible floor on DrivAerML}
\label{app:saturation}

On DrivAerML~\citep{ashton2024drivaerml}, Transolver~\citep{wu2024transolver} saturates in model
size and compute; only adding training cases still lowers test error (\cref{tab:saturation-claims}).
Combining a scaling-law grid over model size $N$, training-set size $D$, and compute multiple
$\kappa$ ($B$ cases per step) with a single-case capacity test and an analytic target-noise floor
brackets DrivAerML's intrinsic irreducible error from above and below.

\begin{table}[!ht]
  \centering
  \caption{Summary of the DrivAerML saturation findings and the evidence for each. Errors are
    relative $L^2$ errors of surface pressure.}
  \label{tab:saturation-claims}
  \footnotesize
  \begin{tabular}{@{}>{\raggedright\arraybackslash}p{0.3\textwidth}>{\raggedright\arraybackslash}p{0.43\textwidth}>{\raggedright\arraybackslash}p{0.2\textwidth}@{}}
    \toprule
    Statement & Evidence & Where \\
    \midrule
    At 400 cases the error changes by \azn{sat-size-spread-400}\% from \azn{sat-fit-N1} to
      \azn{sat-Nmax} million parameters
      & at equal budget ($\kappa=1$, 16 cases per step): \azn{sat-N6-400-k1-B16},
        \azn{sat-N7-400-k1-B16}, \azn{sat-N8-400-k1-B16}, and \azn{sat-N9-400-k1-B16} for 20, 35,
        105, and 354 million parameters
      & \cref{fig:saturation}b \\
    Four times the training compute lowers it by \azn{sat-k4-gain-400-lo} to
      \azn{sat-k4-gain-400-hi}\%
      & from $\kappa=1$ to $\kappa=4$ at 16 cases per step: \azn{sat-N6-400-k1-B16} to
        \azn{sat-N6-400-k4-B16} and \azn{sat-N7-400-k1-B16} to \azn{sat-N7-400-k4-B16}
      & \cref{fig:saturation}c \\
    Each doubling of cases lowers it by about \azn{sat-doubling-gain-200-400}\%
      & \azn{sat-doubling-gain-100-200}\% from 100 to 200 cases and
        \azn{sat-doubling-gain-200-400}\% from 200 to 400 (mean of the 20- and
        35-million-parameter models)
      & \cref{fig:saturation}a \\
    With unlimited cases the scaling floor is \azn{sat-within}\% below the best error
      & $L_\mathrm{irr}=\azn{sat-Lirr}$ (\azn{sat-Lirr-lo}--\azn{sat-Lirr-hi}); best error \azn{sat-best}
      & \cref{tab:saturation-fits}; \cref{fig:saturation}a \\
    Target averaging noise sets an analytic noise estimate of \azn{sat-noise-lo} to
      \azn{sat-noise-hi}
      & $n=\azn{sat-noise}$ at $\tau_{C_d}=\azn{sat-noise-tau}$\,ms (\azn{sat-noise-share}\% of $L_\mathrm{irr}^2$)
      & \cref{app:saturation} (Results) \\
    Single-case overfitting reaches \azn{sat-Lone-ratio} times $L_\mathrm{irr}$
      & $L_1=\azn{sat-Lone}$ ($10^6$ TF32 steps); \azn{sat-strict} in strict float32 ($10^5$ steps; \azn{sat-strict-ratio}$\times$)
      & \cref{fig:saturation}d \\
    \bottomrule
  \end{tabular}
\end{table}

\begin{figure}[!ht]
  \centering
  \includegraphics[width=\textwidth]{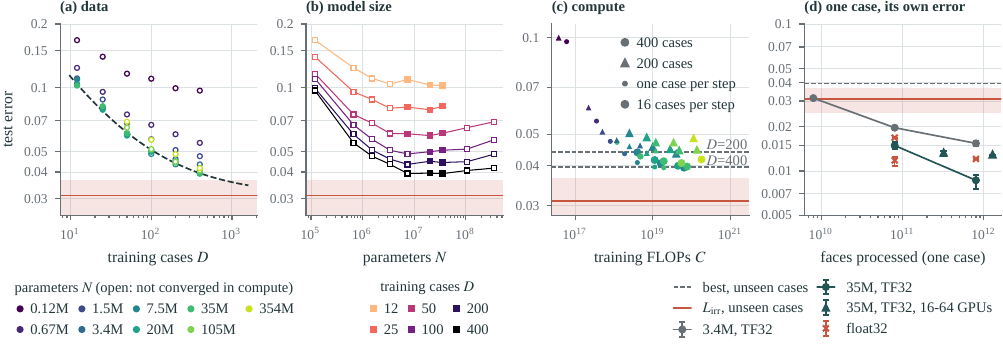}
  \caption{At 200 and 400 training cases, the test error of Transolver on DrivAerML changes by a
    few percent with more parameters (b) and more training compute (c), and each doubling of cases
    still lowers it by about \azn{sat-doubling-gain-200-400}\% (a); a
    single case is fitted to \azn{sat-Lone-ratio} times the floor $L_\mathrm{irr}$ (d). In every
    panel an open marker is a grid cell not converged in compute, and the red line and band are
    $L_\mathrm{irr}$ and its 95\% interval. (a, b) Test error of every grid cell against the number
    of training cases and the number of parameters. The dashed line in (a) is the fit
    $L_\mathrm{irr}+b\,(D/100)^{-\beta}$ to the converged cells of the 20- and 35-million-parameter
    models. (c) Test error against training FLOPs for every finished run at $D=400$ (circles) and
    $D=200$ (triangles), colored by model size as in (a). Larger markers are runs with more cases
    per step (1 or 16, and 4 in one run); the dashed lines are the fit's value at each $D$.
    (d) Error on every face of a single training case at the end of training, against the faces
    processed (steps times faces per step), for the 3.4- and 35-million-parameter models (mean over
    three cases, bars from the lowest to the highest); triangles are runs on 16 and 64 GPUs, crosses
    strict float32 training. This is the error on the case the model trained on, so it can fall
    below $L_\mathrm{irr}$, which bounds the error on cases the model has not seen; the dashed gray
    line is the best error on those cases.}
  \label{fig:saturation}
\end{figure}

\paragraph{Results.}
At $D=400$, the models from 7.5 to 35 million parameters end at the same error, and at equal budget
($\kappa=1$, $B=16$) the 105- and 354-million-parameter models give \azn{sat-N8-400-k1-B16} and
\azn{sat-N9-400-k1-B16}, compared with \azn{sat-N6-400-k1-B16} and \azn{sat-N7-400-k1-B16} at 20
and 35 million: a spread of \azn{sat-size-spread-400}\%, within the \azn{sat-seed-spread}\% spread
between seeds. Four times the budget lowers the error of the 20- and 35-million-parameter models
by \azn{sat-k4-gain-400-lo} to \azn{sat-k4-gain-400-hi}\% at 400 cases and by
\azn{sat-k4-gain-200-lo} to \azn{sat-k4-gain-200-hi}\% at 200. More cases still lower it, by
\azn{sat-doubling-gain-100-200}\% from 100 to 200 and \azn{sat-doubling-gain-200-400}\% from 200
to 400. The floor $L_\mathrm{irr}$ of the fit $L(D)=L_\mathrm{irr}+b\,(D/100)^{-\beta}$, by least
squares on $\log L$ over the \azn{sat-fit-cells} converged cells of the \azn{sat-fit-N1}- and
\azn{sat-fit-N2}-million-parameter models, is \azn{sat-Lirr} (95\% bootstrap interval
\azn{sat-Lirr-lo}--\azn{sat-Lirr-hi}) with $\beta=\azn{sat-beta}$, \azn{sat-within}\% below the
best error of \azn{sat-best} (\cref{tab:saturation-fits}).

\needspace{5\baselineskip}
Together with an analytic target-noise estimate and a single-case capacity check, $L_\mathrm{irr}$
brackets DrivAerML's dataset-intrinsic floor $\varepsilon_\star$ (here in unweighted relative $L^2$)
within roughly a factor of \azn{sat-bracket-ratio}:
\begin{enumerate}[leftmargin=*,itemsep=2pt,topsep=3pt]
  \item \emph{Analytic target-noise estimate ($n=\azn{sat-noise}$, range
    \azn{sat-noise-lo}--\azn{sat-noise-hi}).} Each stored target averages a hybrid RANS--LES run
    over a finite window $T$ (\azn{sat-noise-T-lo}--\azn{sat-noise-T-hi}\,s), and DrivAerML stores
    each face's pressure variance $\sigma^2$. Because zero-mean finite-window averaging noise is
    uncorrelated with any prediction made from geometry alone, its squared norm adds to every
    model's expected squared error. Using the drag coefficient's integral time scale
    $\tau_{C_d}=\azn{sat-noise-tau}$\,ms (read from the published convergence plots) as a uniform
    proxy across all faces gives
    $n=\langle r_\sigma(2\tau_\mathrm{int}/T)^{1/2}\rangle=\azn{sat-noise}$, where
    $r_\sigma=\lVert\sigma\rVert_2/\lVert p\rVert_2$ averages \azn{sat-noise-R} (for
    $\tau_\mathrm{int}$ from \azn{sat-noise-tau-lo} to \azn{sat-noise-tau-hi}\,ms, $n$ spans
    \azn{sat-noise-lo}--\azn{sat-noise-hi}, or \azn{sat-noise-share}\% of $L_\mathrm{irr}^2$ at the
    central value). Because \azn{sat-noise-wheel-sigma}\% of $\sum\sigma_i^2$ comes from the
    wheel/wheelhouse region (\azn{sat-noise-wheel-faces}\% of unweighted faces), where local time
    scales can differ from global $\tau_{C_d}$, and because Jensen's inequality relates expected
    ratios to root-mean-square expectations, $n$ is an analytic estimate rather than a sharp bound;
    independent mesh realizations and bistable wake switching across morphs add further variance.
  \item \emph{Single-case spatial capacity check ($L_1=\azn{sat-Lone}=\azn{sat-Lone-ratio}\,L_\mathrm{irr}$).}
    Trained on a single case (averaged over three cases) with the same \texttt{(centroid, normal)}
    inputs, 200{,}000-face subsampling, and optimizer, Transolver reaches $L_1=\azn{sat-Lone}$
    after $10^6$ TF32 steps and $\azn{sat-strict}$ after $10^5$ strict float32 steps
    (\cref{fig:saturation}d). Even by the triangle inequality alone, a residual of at least
    $L_\mathrm{irr}-L_1=\azn{sat-gap}$ cannot come from within-case spatial expressivity, face
    subsampling, or numerical precision. That leaves at least \azn{sat-triangle-sq-share}\% of
    $L_\mathrm{irr}^2$ unexplained (\azn{sat-nonspatial-sq-share}\% under an orthogonal
    decomposition, or \azn{sat-triangle-strict-sq-share}\% and
    \azn{sat-nonspatial-strict-sq-share}\% using the strict-float32 $L_1$). Single-case
    memorization tests within-case approximation ($x\mapsto p_S(x)$) rather than cross-case
    generalization ($S\mapsto p_S$) at fixed capacity $N$, so cross-case inductive bias can still
    keep $L_\mathrm{irr}$ above $\varepsilon_\star$.
  \item \emph{Empirical upper bound and cross-architecture check
    ($\varepsilon_\star\le L_\mathrm{irr}=\azn{sat-Lirr}$).} Because any remaining cross-case
    approximation residual is non-negative, $L_\mathrm{irr}$ upper-bounds $\varepsilon_\star$
    subject to the single-power-law extrapolation beyond $D=400$. As \cref{tab:saturation-fits}
    shows, identifying $L_\mathrm{irr}>0$ relies on curvature across $D\le50$: on the
    \azn{sat-power-cells} cells with $D\ge100$ alone, a floor-free power law ($\beta=\azn{sat-power-exponent}$)
    achieves a lower RMS log residual (\azn{sat-power-rms}\%) than the floor fit
    (\azn{sat-floor-rms-same-cells}\%), and the cell-bootstrap intervals treat the 12 nested
    $(N, D)$ cells as independent. Consistent with a dataset-level floor rather than a
    Transolver-specific ceiling, \citet{sharpe2026isla} finds that ISLA lowers DrivAerML
    surface-pressure error relative to Transolver and GeoTransolver on the smallest training split
    (\key{super_scarce}) yet converges with both architectures to nearly identical errors at
    \azn{sat-pool} training cases.
\end{enumerate}

\paragraph{Protocol.}
\label{app:saturation-protocol}
The data are the \azn{sat-runs} complete public runs of DrivAerML in the standard PhysicsNeMo
train/validation split (one of the 484 public runs lacks full surface data). The test set is that
split's \azn{sat-test-cars} validation runs, and a selection set of \azn{sat-val-cars} training
runs, on which the validation error is measured, selects each run's checkpoint, each cell's budget,
and whether a cell is converged in compute. The training sets are the first $D$ cases of a fixed
random order of the other \azn{sat-pool}, so they are nested. The model is the unmodified public
PhysicsNeMo Transolver~\citep{physicsnemo} (commit \texttt{\azn{sat-commit}}), given each face's
centroid and normal and trained on the mean squared error of standardized $p$ and $\vtau$, with
200{,}000 faces of one case per GPU and step, AdamW, and a cosine schedule. The metric is the
relative $L^2$ error of surface pressure over every face, averaged over cases. A cell's value is the
test error of its run with the lowest validation error, and a cell is converged in compute if its
last 4-fold increase in budget lowered the validation error by less than 1\%; \azn{sat-cells-conv}
of \azn{sat-cells} cells are. A run's training compute $C_\mathrm{FLOP}$ counts its case draws times
the FLOPs of one draw, and the study used \azn{sat-gpuh} GB300 GPU-hours. Every cell's error, every
run's budget, and the settings of each run \azcodepublic{are released}{will be released} in
\texttt{drivaerml-saturation.json}.

\begin{table}[!ht]
  \centering
  \caption{Estimates of the floor $L_\mathrm{irr}$. The text uses the first, fitted to the
    converged cells of two models whose validation errors agree within the spread between seeds.
    The joint law is
    $L(N,D)=L_\mathrm{irr}+a\,(N/10^6)^{-\alpha}+b\,(D/100)^{-\beta}$.}
  \label{tab:saturation-fits}
  \footnotesize
  \setlength{\tabcolsep}{4.5pt}
  \begin{tabular}{@{}l>{\raggedright\arraybackslash}p{0.27\textwidth}l>{\raggedright\arraybackslash}p{0.25\textwidth}@{}}
    \toprule
    Fit & Fitted to & $L_\mathrm{irr}$ (95\%) & Note \\
    \midrule
    $L(D)$ \textbf{(used)} & the \azn{sat-fit-cells} converged cells of the 20- and 35-million-parameter models
      & \azn{sat-Lirr} (\azn{sat-Lirr-lo}--\azn{sat-Lirr-hi}) & $\beta=\azn{sat-beta}$; residual \azn{sat-fit-rms}\% \\
    $L(D)$ & the same without $D=12$ & \azn{sat-noD12-E} & $\beta=\azn{sat-noD12-beta}$ \\
    Joint law & all \azn{sat-joint-cells} converged cells
      & \azn{sat-Lirr-joint} (\azn{sat-Lirr-joint-lo} to \azn{sat-Lirr-joint-hi})
      & $\alpha$ at its bound (\azn{sat-alpha-joint}): the size term is not identified \\
    Joint law & every cell, converged or not & \azn{sat-Lirr-all} & \\
    Law in $N$, $D$, and case draws & the end points of all \azn{sat-runs-C} runs at $B=16$
      & \azn{sat-Lirr-C} & floor at $D=400$ for unlimited compute and model size: \azn{sat-floor400-C} \\
    Floor-free $b\,(D/100)^{-\beta}$ & the \azn{sat-power-cells} converged cells with $D\ge100$
      & $0$ (fixed) & $\beta=\azn{sat-power-exponent}$; residual \azn{sat-power-rms}\% (floor fit: \azn{sat-floor-rms-same-cells}\%) \\
    \bottomrule
  \end{tabular}
\end{table}
\FloatBarrier

\section{Other datasets: sources and the shape-diversity score}
\label{app:datasets}

The values in \cref{tab:datasets,tab:datasets-2d} are quoted from each dataset's paper or card:
AirfRANS from its paper~\citep{bonnet2022airfrans}; UniFoil, CFDBench, and FlowBench from
theirs~\citep{kanchi2025unifoil,luo2023cfdbench,tali2024flowbench}; DrivAerML, WindsorML, AhmedML,
DrivAerNet++, HiLiftAeroML, BlendedNet, and DrivAerStar from their
papers~\citep{ashton2024drivaerml,ashton2024windsorml,ashton2024ahmedml,elrefaie2024drivaernetpp,ashton2026hiliftaeroml,sung2025blendednet,qiu2025drivaerstar}
and Hugging Face cards; SHIFT-SUV from its card and the publisher's
page~\citep{luminary2025shiftsuv}; ShapeNet-Car from its paper~\citep{umetani2018learning};
SuperWing and Emmi-Wing from their papers~\citep{yang2026superwing,paischer2025emmiwing}; and
WindTunnel-20k from its card and public simulation templates~\citep{inductiva2024windtunnel,valerio2024openfoam}.
Each cell was checked against these sources. DrivAerML's card lists 484 of its 500 runs as
public. SuperWing draws eight operating conditions per wing and keeps the converged ones, and
Emmi-Wing draws each wing with its own speed and incidence. \Cref{tab:datasets-2d} lists three
two-dimensional datasets in the columns of \cref{tab:datasets}.

\begin{table}[!ht]
  \centering
  \caption{Three two-dimensional CFD datasets for machine learning, in the columns of
    \cref{tab:datasets}.}
  \label{tab:datasets-2d}
  \scriptsize
  \setlength{\tabcolsep}{3pt}
  \begin{tabular}{@{}lcclllr@{}}
    \toprule
    Dataset & Families & Enclosures & Operating points & Solver & Samples & Cells per sample \\
    \midrule
    AirfRANS~\citep{bonnet2022airfrans} & 1 & 1 & $\Rey$ $2$--$6\times10^6$, incidence & RANS & \num{1000} & 0.25--0.3\,M \\
    UniFoil~\citep{kanchi2025unifoil} & 1 & 1 & wide $\Rey$ and Mach & RANS, $e^N$ transition & $>$\num{500000} & \textcolor{gray}{--} \\
    CFDBench~\citep{luo2023cfdbench} & 4 & 4 & varied & laminar or RANS & 739 & \textcolor{gray}{--} \\
    \bottomrule
  \end{tabular}
\end{table}

\paragraph{Shape diversity.}
For each reference dataset, the shape-diversity comparison (\cref{fig:vendi}) uses the training
surfaces of its design sweep. DrivAerML and SHIFT-SUV give 400 surfaces each, matched against 400
source geometries drawn from each \aerozoo{} geometry family's pool prior to solving. HiLiftAeroML
gives its 126 distinct training geometries (collapsed across angles of attack), and for that
comparison every \aerozoo{} geometry family is subsampled to 126. Each surface is described by 83
rotation- and scale-invariant numbers computed from 6{,}144 area-weighted surface points centered at
their mean and divided by the oriented bounding-box diagonal: 11 global geometric scalars and two
histograms: 48 bins of pairwise distances ($D_2$, 24{,}576 pairs) and 24 bins of angles formed by
triplets of sampled surface points ($A_3$, 12{,}288 triplets). Each of the three descriptor blocks
is z-scored and scaled by $1/\sqrt{d_b}$ so the three blocks contribute equal total variance. The
similarity of two surfaces is a Gaussian kernel of the Euclidean distance between descriptors, with
bandwidth set to the median distance to the 20th nearest neighbor.

\paragraph{How far the comparison depends on its settings.}
A check with the published descriptor tables and functions varied the bandwidth, the descriptors,
and the sample size.
\begin{itemize}[itemsep=1pt]
  \item \emph{Calibration.} Standardization and the bandwidth are fitted once, on the five
    non-Blender geometry families and the two automotive datasets pooled. SHIFT-SUV's surfaces are
    open, so \azn{vendi-shiftsuv-missing} of its volume-based scalars are missing and filled with
    pooled medians. Using the two histograms alone ($D_2+A_3$, dropping all scalars) still leaves
    the least varied family \azn{vendi-hist-ratio-min-lo} to \azn{vendi-hist-ratio-min-hi} times
    the more varied automotive dataset.
  \item \emph{The ceiling.} A score cannot exceed the number of shapes, and Zero-to-CAD is at that
    ceiling (\azn{vendi-ceiling-pct}\% of its shapes count as distinct). Its ratio to the automotive
    datasets therefore grows with the sample size, from \azn{vendi-ratio-max-n-lo} at
    \azn{vendi-n-lo} shapes to \azn{vendi-ratio-max} at \azn{vendi-n}, and with the bandwidth, from
    \azn{vendi-ratio-max-k-lo} to \azn{vendi-ratio-max-k-hi}.
  \item \emph{The least varied family.} With the bandwidth set by the \azn{vendi-k-lo}th to the
    \azn{vendi-k-hi}th nearest neighbor, the least varied family stays at
    \azn{vendi-ratio-min-k-lo} times the more varied automotive dataset.
  \item \emph{The order.} Each of those five families scores above both automotive datasets in
    \azn{vendi-settings-ordered} of \azn{vendi-settings} settings (five bandwidths by seven
    subsets of the three descriptor blocks). In the others, the distance histogram alone, the
    Streamlined family scores \azn{vendi-d2-ratio-min-lo} to \azn{vendi-d2-ratio-min-hi} times
    DrivAerML.
  \item \emph{Blender families.} Scored at the primary setting under the same calibration, the two
    Blender families lie between the lowest and the highest.
\end{itemize}

\section{Datasheet}
\label{app:datasheet}

We answer the questions of \citet{gebru2021datasheets} in brief.

\paragraph{Motivation.}
\aerozoo{} was created to measure whether neural surrogates of incompressible RANS flow generalize
across shapes, enclosures, boundary conditions, and flow regimes (which datasets varying one design
family in one flow domain cannot measure) and to show what a well-posed CFD sample must contain. It
is a benchmark that can also serve as a training corpus. It was created by the
authors\reviewpreprint{ \rvFunding}{ at NVIDIA, which funded the work}.

\paragraph{Composition.}
Each instance is a \emph{sample}: one solved flow problem stored as a PhysicsNeMo
\key{DomainMesh}~\citep{physicsnemo}. \Cref{app:schema} lists its contents. Release v1.3 holds
\azn{complete} samples of a designed population of \azn{planned} cases; the release manifest gives
every planned case a status, so that the \azn{failed} failed and \azn{timed-out} timed-out cases
are listed rather than missing. The release holds one completed mascot sample (of two unplanned
mascot cases), which belongs to no split and is not among the \azn{complete} counted in this paper.
No two cases share a primary body; the second body that a case adds with probability 0.03 is a
scaled copy of its own primary body. The \azn{splits} frozen splits, disjoint in geometry, are lists
of case identities. Of the samples, \azn{stable} have a stable time mean and form the canonical
benchmark. Numerical verification is in \cref{sec:trust,app:verification}. Six of the seven
geometry families were generated for this dataset. The Zero-to-CAD family comprises 400 planned
cases drawn from a farthest-point sample of 500 parts, within three strata of bluffness, from the
52{,}408 watertight single-body parts of the public Zero-to-CAD 100K corpus that pass simple shape
gates~\citep{adsk2026zerotocad100k}; that corpus is released under the Apache License 2.0, so every
geometry in the release is permissively licensed. The release contains no personal or confidential
data.

\paragraph{Collection.}
Every sample was generated by simulation, from a seed and a frozen recipe (\cref{app:recipe}), with
OpenFOAM~14 and \texttt{snappyHexMesh}, between 2026-09-13 and \azn{frozen-date}. The released
samples took \azn{core-hours-total-m}\,M CPU core-hours of solver time on preemptible nodes
(excluding meshing and the failed and timed-out cases). Every case's problem was posed before any mesh
existed, including the cases that later failed or timed out, and failed cases were never replaced.

\paragraph{Preprocessing.}
The exporter merges the solver's patches by boundary-condition type, triangulates them, and converts
kinematic pressure to pascals. Interior points are OpenFOAM's own cell centroids, checked against
centroids recomputed from the face geometry. Release v1.3 closes boundary meshes in 3{,}396 samples
by replacing affected boundary polygons with positive-area triangle fans and keeps interior
fields, cell volumes, and splits byte-identical to v1.2 (\cref{app:schema}). A sample becomes
visible only after it has been written, reloaded, and compared tensor by tensor. The native solver
cases are retained, so samples can be re-exported without solving again.

\paragraph{Uses.}
Intended uses are training and evaluating surrogates of steady or time-mean incompressible flow,
pretraining, and studying generalization across geometry, enclosure, boundary conditions, and
Reynolds number. Samples store time-mean solutions of incompressible, isothermal RANS equations
(\cref{sec:trust}) rather than time series or physical wind-tunnel measurements.

\paragraph{Distribution.}
The release \azpublic{is}{will be} hosted at \datasetlink{} as one archive per sample
(\azn{bytes-total-tb}\,TB in total), with the split files, the release manifest, and a catalog
(\key{index.json}) that answers subset questions without opening a sample\reviewonly{, and \rvCroissant}.
The dataset is released under the Creative Commons Attribution 4.0 International (CC BY 4.0)
license; the Zero-to-CAD parts keep the Apache License 2.0 and its notice.
The scoring code \azcodepublic{is at}{will be added to} \codelink.

\paragraph{Maintenance.}
The authors maintain the release. A release that only adds samples, such as one that finishes the
timed-out cases, is a point release and moves no case between splits; a change to the artifacts,
the recipe, the schema, or the planned population takes a new minor version. A changelog records
every version and its updates.

\end{document}